\documentclass[11pt]{article}

\usepackage[preprint]{acl}

\usepackage{times}
\usepackage{latexsym}

\usepackage[T1]{fontenc}

\usepackage[utf8]{inputenc}

\usepackage{microtype}

\usepackage{inconsolata}

\usepackage{graphicx}

\usepackage{booktabs}
\usepackage{multicol}
\usepackage{amsmath}
\usepackage{bm}
\usepackage{subcaption}

\makeatletter
\def\new@fontshape{}
\makeatother
\usepackage{gb4e}
\noautomath

\title{Fine-Grained Analysis of Shared Syntactic Mechanisms \\ in Language Models}

\author{
 \textbf{Ryoma Kumon\textsuperscript{1,2}} \quad
 \textbf{Hitomi Yanaka\textsuperscript{1,2,3}}
\\
 \textsuperscript{1}The University of Tokyo\quad
 \textsuperscript{2}RIKEN\quad
 \textsuperscript{3}Tohoku University
\\
   \texttt{\{kumoryo9, hyanaka\}@is.s.u-tokyo.ac.jp}
}

\begin{document}
\maketitle
\begin{abstract}
While language models demonstrate sophisticated syntactic capabilities, the extent to which their internal mechanisms align with cross-constructional principles studied in linguistics remains poorly understood.
This study investigates whether models employ shared neural mechanisms across different syntactic constructions by applying causal interpretability methods at a granular level.
Focusing on filler-gap dependencies and negative polarity item (NPI) licensing, we utilize activation patching to identify the functional roles of specific attention heads and MLP blocks.
Our results reveal a highly localized and shared mechanism for filler-gap dependencies located in the early to middle layers, whereas NPI processing exhibits no such unified mechanism.
Furthermore, we find that these mechanisms identified by activation patching generalize to out-of-distribution, while distributed alignment search, a supervised interpretability method, is susceptible to overfitting on narrow linguistic distributions.
Finally, we validate our findings by demonstrating that the manipulation of the identified components improves model performance on acceptability judgment benchmarks.\footnote{All code and datasets are avilable at \url{https://github.com/ynklab/shared_syntactic_mechanism}.}
\end{abstract}

\section{Introduction}

Recent advancements in language models (LMs) have demonstrated their ability to process linguistic expressions with complex syntactic structures.
However, it remains unclear to what extent their internal mechanisms align with linguistic theory or resemble human processing mechanisms.
While linguistics has long studied underlying elements shared across constructions, clarifying whether LMs employ shared syntactic mechanisms across different constructions will lead to a deeper understanding of both linguistic structure and language modeling~\citep{Futrell_Mahowald_2025}.

To investigate the internal processes, causal abstraction methods \cite{NEURIPS2020_92650b2e,geiger2021causal,pmlr-v236-geiger24a} have emerged for identifying the causal impact of internal components on model behavior.
Previous studies have utilized these methods to analyze syntactic mechanisms~\citep{arora-etal-2024-causalgym,elazar-etal-2021-amnesic}, specifically investigating the commonality of mechanisms in subject-verb agreement~\citep{finlayson-etal-2021-causal} and filler-gap dependencies~\citep{boguraev-etal-2025-causal}.

However, several challenges remain.
First, existing analyses primarily focus on examining representations in the residual stream and comparing their scores across tokens, without identifying the components that impact the residual stream, leaving the underlying mechanisms at the level of layers and attention heads largely unexplored.
As a result, it remains unclear at what level of granularity shared mechanisms operate.
For example, \citet{boguraev-etal-2025-causal} analyzes representations in the residual stream, but this approach may fail to distinguish between mechanisms that rely on different sets of attention heads or MLP blocks while producing similar residual representations.

Second, causal analysis methods that require training are susceptible to overfitting, which makes verifying out-of-distribution (OOD) generalization essential~\citep{wu2024replymakelovetal}.
Nevertheless, prior research applying these methods to the analysis of syntactic mechanisms~\citep{arora-etal-2024-causalgym,boguraev-etal-2025-causal} has not sufficiently validated the OOD generalization of the identified mechanisms.
Thus, their faithfulness, which concerns whether the identified components truly reflect the model’s internal mechanisms rather than artifacts of specific datasets or lexical distributions, remains unestablished.

To address these issues, this study employs activation patching~\citep{NEURIPS2020_92650b2e, geiger2021causal}, which is a causal analysis method without any additional training, and investigates the commonality of mechanisms across different constructions at the granularity of layers and attention heads.
We focus on two well-studied phenomena, filler-gap dependencies (FGDs) and negative polarity item (NPI) licensing.
Both phenomena appear across diverse surface constructions while posing distinct linguistic challenges.
FGDs require long-range structural processing, whereas NPI licensing involves integrating syntactic structure with semantic properties of licensors, which may give rise to different or construction-specific mechanisms.
We analyze whether LMs employ any shared mechanisms across constructions for each of these phenomena and, if so, how the mechanisms work.
Furthermore, to validate the identified mechanisms, we evaluate changes in model accuracy on acceptability judgment benchmarks by manipulating the identified components.
Additionally, we assess the OOD generalization of these mechanisms by comparing our results with distributed alignment search (DAS; \citealp{pmlr-v236-geiger24a}), a training-based causal interpretability method.

Our analysis reveals that, while LMs have a shared mechanism in FGDs, such commonality was not observed in NPI licensing.
Specifically, we found that the mechanisms contributing to FGDs are localized within a small number of attention heads situated in the early to middle layers. 
Furthermore, amplifying the activation values of these identified components led to improved accuracy on acceptability judgment tasks.
Additionally, while DAS results fluctuated significantly in OOD settings, suggesting a potential overfitting to specific lexical or syntactic distributions, activation patching yielded consistent results across different distributions.
These findings demonstrate the validity and faithfulness of our analytical approach.

\section{Related Work}
\subsection{Analysis of Syntactic Mechanisms}
Analysis of syntactic mechanisms in LMs is broadly categorized into probing, which examines the extent to which models encode syntactic knowledge~\citep{hewitt-manning-2019-structural, clark-etal-2019-bert}, and causal analysis, which investigates the mechanisms that causally influence model behavior during syntactic processing~\citep{elazar-etal-2021-amnesic, lasri-etal-2022-probing, arora-etal-2024-causalgym}.
To elucidate the mechanisms directly governing model behavior, this study adopts the latter causal approach.

Shared mechanisms across different constructions have been explored through both causal methods (e.g., subject-verb agreement~\citep{finlayson-etal-2021-causal} and FGDs~\citep{boguraev-etal-2025-causal}) and probing~\citep{kryvosheieva2025differenttypessyntacticagreement}.
However, studies using causal methods mainly focus on token-level analysis, leaving the degree of commonality at finer granularities, such as attention heads, largely unexplored.
Moreover, learning-based causal methods face the risks of overfitting, which requires verification of OOD generalization~\citep{wu2024replymakelovetal}, and yet existing studies \cite{arora-etal-2024-causalgym, boguraev-etal-2025-causal} have not sufficiently validated this aspect.
Conversely, probing-based analyses~\citep{kryvosheieva2025differenttypessyntacticagreement} fail to verify the direct causal impact of identified components on actual behavior.
In this work, we primarily utilize training-free causal analysis to investigate both OOD generalization and the functional impact of internal components on model behavior.

\subsection{Filler-Gap Dependencies and Negative Polarity Items}
\begin{table*}[ht]
\centering
\small
\setlength{\tabcolsep}{2pt}
\begin{tabular}{lllllllll}
\toprule
\multicolumn{9}{c}{\emph{Filler-Gap Dependencies}} \\
\midrule
Construction & Abbr. & Prefix & Filler & NC & Article & Noun & Verb & Output\\
\midrule
Emb. Wh-q. (know)& \textsc{EWhK} & The man knows & [\textbf{who}/\textbf{that}] && the & teacher & liked & [\textbf{.}/\textbf{her}]\\
Emb. Wh-q. (wonder)& \textsc{EWhW} & The boy wondered & [\textbf{who}/\textbf{if}] && the & doctor & admired & [\textbf{.}/\textbf{him}]\\
Matrix wh-q& \textsc{MWh} & Then, & [\textbf{who}/$\mathbf{\phi}$] & did & the & girl & choose &[\textbf{?}/\textbf{them}]\\
Restr. Rel. Clause& \textsc{RelCl} & The customer & [\textbf{who}/\textbf{and}] && the &lady &sounded like &[\textbf{.}/\textbf{me}]\\
Cleft & \textsc{Cleft} & It was & [\textbf{the man}/\textbf{clear}] & that&  the &boss &scared &[\textbf{.}/\textbf{him}]\\
Pseudo-cleft & \textsc{PCleft} & & [\textbf{Who}/\textbf{That}] && the & dancer & is listening to &[\textbf{is}/\textbf{you}]\\
Topicalization & \textsc{Topic} & Actually, & [\textbf{the kid}/$\mathbf{\phi}$] & & the& guest & hated & [\textbf{.}/\textbf{them}]\\
\midrule
\multicolumn{9}{c}{\emph{Negative Polarity Item Licensing}} \\
\midrule
Construction & Abbr. & Prefix & Licensor &\multicolumn{3}{l}{NC} & Last & Output\\
\midrule
Conditionals & \textsc{Cond} & The host will sleep & [\textbf{if}/\textbf{while}] & \multicolumn{3}{l}{the guest} & eats & [\textbf{any}/\textbf{some}]\\
Determiner Negation & \textsc{DNeg} & & [\textbf{No}/\textbf{The}] &\multicolumn{3}{l}{patient have} & liked & [\textbf{any}/\textbf{some}]\\
Scope of Only & \textsc{SOnly} & & [\textbf{Only}/\textbf{Even}] & \multicolumn{3}{l}{the boys} & have & [\textbf{any}/\textbf{some}]\\
Quantifers & \textsc{Qnt} & These are & [\textbf{all}/\textbf{some}]  &\multicolumn{3}{l}{of the students who} & showed & [\textbf{any}/\textbf{some}]\\
Embedded Questions & \textsc{EmbQ} & The senators & [\textbf{wonder whether} &\multicolumn{3}{l}{the man has} &  found& [\textbf{any}/\textbf{some}]\\
& & &  /\textbf{know that}]&&&\\
Simple Questions & \textsc{SmpQ} & & [\textbf{Has}/$\mathbf{\phi}$] &\multicolumn{3}{l}{the actor} & sold & [\textbf{any}/\textbf{some}]\\
Superlatives & \textsc{Sup} & This is the & [\textbf{fastest}/\textbf{fast}] &\multicolumn{3}{l}{kid that had} & liked & [\textbf{any}/\textbf{some}] \\
Only & \textsc{Only} & They are the& [\textbf{only}/\textbf{upset}] &\multicolumn{3}{l}{teachers that} & makes & [\textbf{any}/\textbf{some}]\\
\midrule
\multicolumn{9}{c}{\emph{Control}} \\
\midrule
Construction & Abbr. & Prefix & Filler/Licensor & NC & Article & Noun & Verb/Last & Output\\
\midrule
Capital Knowledge & \textsc{Ctrl} &  & [\textbf{Paris}/\textbf{Rome}] & is & the  & captital & of & [\textbf{France}/\textbf{Italy}]\\
\bottomrule
\end{tabular}

\caption{Examples of minimal pairs of the constructions in the dataset.
NC stands for ``No comparison", and is not used for analysis.
}
\label{tab:patterns}
\end{table*}

FGDs refer to relationships between a filler and a gap, where the filler is typically a wh-word or phrase, and the gap is an empty position corresponding to the filler.
These dependencies underlie various constructions.
For example, in the \textsc{EWhK} construction in Table~\ref{tab:patterns}, the filler ``who'' is interpreted as the direct object of ``liked'' and creates the gap before ``.'', making the pronoun ``her'' ungrammatical.
Several studies~\citep{wilcox-etal-2018-rnn,ozaki-etal-2022-well,wilcox2024learnability,lan2024} have conducted surprisal-based analyses using FGDs to investigate the syntactic capabilities of LMs and have shown the sensitivity of LMs to structural constraints.
Generalizations across constructions of FGDs have also been analyzed.
It has been reported that long short-term memory (LSTM) networks struggle to generalize to filler-gap constructions beyond those seen in training~\citep{howitt-etal-2024-generalizations} and that LMs trained on child-language data struggle to generalize across constructions~\citep{chang-etal-2025-mind}.

NPIs are expressions that must be licensed by certain contexts, such as negation and questions.
For instance, in the \textsc{DNeg} construction in Table~\ref{tab:patterns}, the NPI ``any'' is licensed by the licensor ``No''.
Early formal accounts attempted to uniformly explain that NPIs are licensed by downward entailing environments~\citep{ladusaw1979polarity}.
However, \citet{zwarts1998three} and \citet{giannakidou1998polarity} proposed a more nuanced hierarchy of NPIs based on the strength of their licensing requirements.
This was motivated by the observation that different NPIs require different licensing contexts and many are licensed by non-downward entailing environments, such as questions.
Regarding LMs, \citet{warstadt-etal-2019-investigating} showed that masked LMs can handle NPIs.
\citet{jumelet-etal-2021-language} argued that a two-layer LSTM language model processes NPI constructions uniformly through a monotonicity-based mechanism based on probing the internal representation.
Under this mechanism, the model evaluates NPI licensing by determining whether the items occur within a downward entailing environment.
\citet{decarlo-etal-2023-npis} showed that LMs handle NPI licensing in a more complex and non-uniform way, based on analysis of output probabilities.
In contrast, we study the internal workings of LMs in processing FGDs and NPI licensing by employing causal interpretability methods rather than probing or behavior analysis focused on model output.

\section{Experimental Settings}
\label{sec:settings}

\subsection{Dataset}
\label{subsec:dataset}
We construct the dataset for analysis, which consists of minimal pairs of constructions for FGDs and NPI licensing.
Table~\ref{tab:patterns} shows examples of minimal pairs of each construction in FGDs and NPI licensing, seven and eight each.
We determine the construction coverage based on the previous studies for FGDs~\citep{boguraev-etal-2025-causal} and for NPI licensing~\citep{warstadt-etal-2019-investigating,decarlo-etal-2023-npis}.
These constructions have lexical variations, which makes it hard for interpretability methods to capture shallow lexical patterns.
For example, the constructions in FGDs have several words for filler, which prevents us from identifying a mechanism sensitive to a single lexical item such as ``who''.

To distinguish mechanisms associated with syntactic structures from those impacted by unrelated factors such as linear distance, we incorporate a control construction that involves the retrieval of world knowledge.
Specifically, we use sentences that require the model to associate a capital city with its country name.
The sentences are designed so that the linear distance between the dependent tokens is similar to that in FGDs and NPI constructions.

We generate the dataset by extending the data generation scripts\footnote{\url{https://github.com/alexwarstadt/data_generation}} by \citet{warstadt-etal-2020-blimp-benchmark}.
The generation scripts use pre-defined structural templates, with lexical items sampled from an extensive vocabulary.
This approach enables the creation of sentences that span a wider distribution of vocabulary and constructions than previous studies~\citep{arora-etal-2024-causalgym,boguraev-etal-2025-causal}.
Notably, using only one output word per construction pattern may lead to the identification of a mechanism specific to that lexical item and cause supervised interpretability methods to overfit to the output token.
To mitigate these issues, we use multiple words for output tokens.
Note that, for the NPI constructions, we only use ``ever'' and ``any'' for output tokens because these are the only items that are licensed across all eight selected licensing contexts.

For the training and evaluation of DAS, we also split the dataset into the training, in-distribution (ID), and OOD test sets.
The training and ID test sets are generated from the same vocabulary set, and the OOD test set is generated from an entirely disjoint vocabulary to evaluate the robustness\footnote{\citet{boguraev-etal-2025-causal} examine the impact of animacy differences in the inputs.
However, in their setting, some of the output tokens overlap, and therefore their evaluation is not strictly out-of-distribution.}.
We ensure that no sentences overlap between the training and ID test sets.
Unless otherwise specified, we report the results averaged across the ID and OOD test sets.
The examples in Table~\ref{tab:patterns} are all drawn from the ID test set.

\subsection{Models}
We use Pythia 1B, 2.8B, 6.9B~\citep{biderman2023pythia} and Gemma 3 1B, 12B~\citep{gemma_2025}.
Using different model sizes and model families allows for analyses of their effect on the development of syntactic mechanisms.
Furthermore, we use multiple checkpoints at different training steps for Pythia models at 1k, 2k, 3k, 4k, 10k, and 143k (final checkpoint) to compare how the mechanism is acquired throughout the pre-training.
The Pythia family is particularly suited for this comparison because all models are trained on the same training data in the same order.
Unless otherwise specified, we report the results of Pythia 1B.

\section{Analysis of Shared Mechanisms}
\begin{figure*}[ht]
\centering

  \begin{subfigure}{\linewidth}
    \centering
    \includegraphics[width=\linewidth]{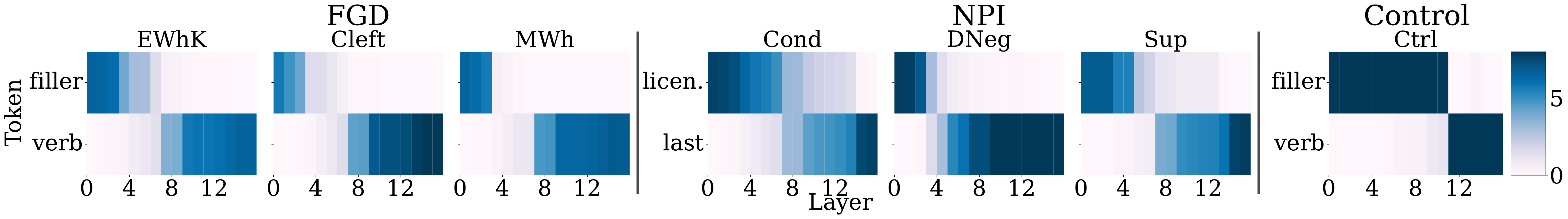}
    \caption{Residual stream}
  \end{subfigure}
  \begin{subfigure}{\linewidth}
    \centering
    \includegraphics[width=\linewidth]{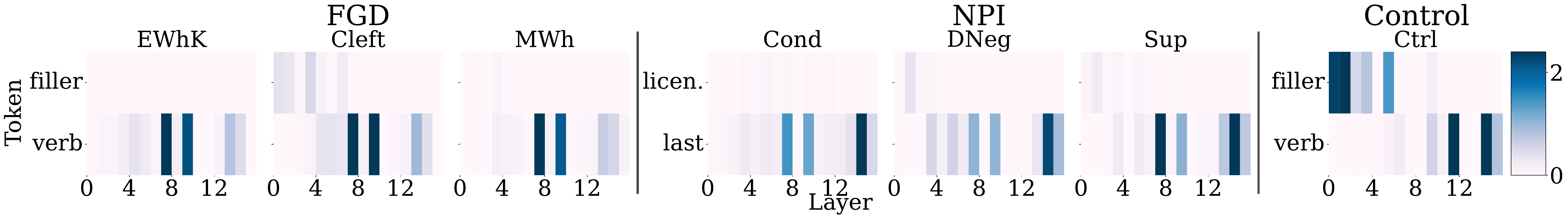}
    \caption{Attention}
  \end{subfigure}
  \begin{subfigure}{\linewidth}
    \centering
    \includegraphics[width=\linewidth]{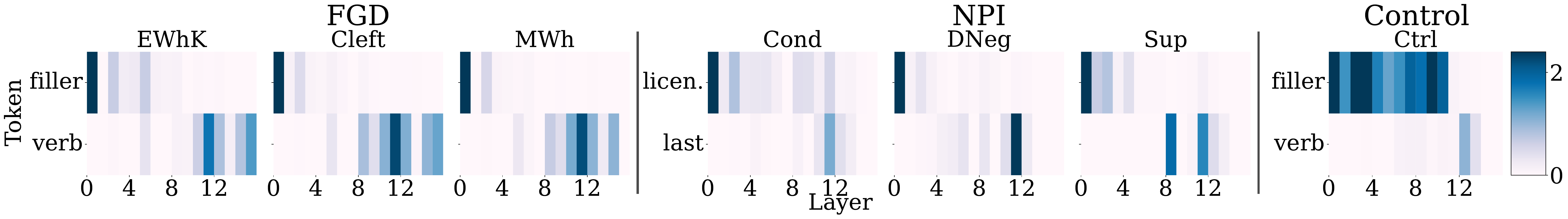}
    \caption{MLP}
  \end{subfigure}
  \begin{subfigure}{\linewidth}
    \centering
    \includegraphics[width=\linewidth]{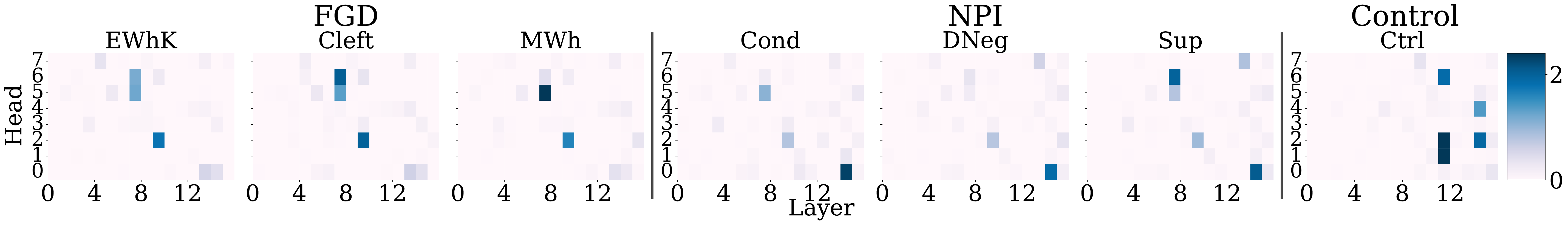}
    \caption{Attention head}
  \end{subfigure}

  \caption{\textsc{Odds} scores with activation patching in Pythia 1B.
  Note that layer numbers are zero-indexed and token names correspond to those in Table~\ref{tab:patterns}.
  The full results including other constructions and tokens are presented in Appendix~\ref{sec:full_ap}.}
  \label{fig:act_patch_fg}
\end{figure*}

\label{sec:methods}
\subsection{Activation Patching}
To analyze mechanisms employed in the task described in the previous section, we use activation patching~\citep{NEURIPS2020_92650b2e, geiger2021causal}.
Activation patching is a method to identify which component is responsible for the model behavior and is defined by the following:
\begin{equation*}
f_{\mathrm{interv}}(\bm{b},\bm{s})=f(\bm{s})
\end{equation*}
It replaces the activation $f(\bm{b})$ of a component $f$ during an inference of a base input $\bm{b}$ with cached activations $f(\bm{s})$ from another inference with a source input $\bm{s}$, and observes how this affects the model’s output.
A source input is designed to be minimally different from a base input, the only difference being the concept of interest, to isolate the causal role regarding the concept.

\subsection{Evaluation Metric}
To quantify causal effect, we employ a modified version of log-odds ratio (\textsc{Odds}), originally introduced by \citet{arora-etal-2024-causalgym}.
It is defined as:
\begin{align*}
&\mathrm{\textsc{Odds}}(p,p_{f_{\mathrm{interv}}},T)=\\&\frac{1}{|T|}\sum_{(\bm{b},\bm{s},y_b,y_s)\in T}\log\left(\frac{p(y_b|\bm{b})}{p(y_b|\bm{s})}\frac{p_{f_{\mathrm{interv}}}(y_b|\bm{s},\bm{b})}{p_{f_{\mathrm{interv}}}(y_b|\bm{b},\bm{s})}\right),
\end{align*}
where $y_b$ and $y_s$ are the output tokens of the inputs $\bm{b}$ and $\bm{s}$, respectively, and $T$ denotes the test set.
While the original metric compares the relative probabilities of two tokens $y_b$ and $y_s$ given a single input, we modify it to measure the shift in probability for a specific token $y_b$ for two inputs $\bm{b}$ and $\bm{s}$ before and after the targeted intervention.
This adjustment is critical because the grammatical licensing of an NPI is largely independent of the choice of alternative, non-NPI lexical items.
We use only minimal pairs that have NPIs as $y_b$ and non-NPIs as $y_s$ for analysis, which leads to measuring the shift in probability of NPIs across two inputs.
Note that these restrictions do not apply to the minimal pairs in FGDs, and the \textsc{Odds} score in FGDs is equivalent to the one originally defined by \citet{arora-etal-2024-causalgym} (see details in Appendix~\ref{sec:a_metric}).

This metric captures the extent to which a specific model component contributes to the next token prediction.
A higher \textsc{Odds} score indicates that the intervened component is more causally efficacious.
Thus, we expect higher scores in components that propagate information from critical tokens, such as fillers or licensors, to the final layers at the last token.
If mechanisms are shared across constructions, score distributions will be similar across constructions at each component level.
In contrast, distinct mechanisms should yield different score distributions.

\subsection{Results}
\begin{figure*}[ht]
\centering

  \begin{subfigure}{0.8\linewidth}
    \centering
    \includegraphics[width=\linewidth]{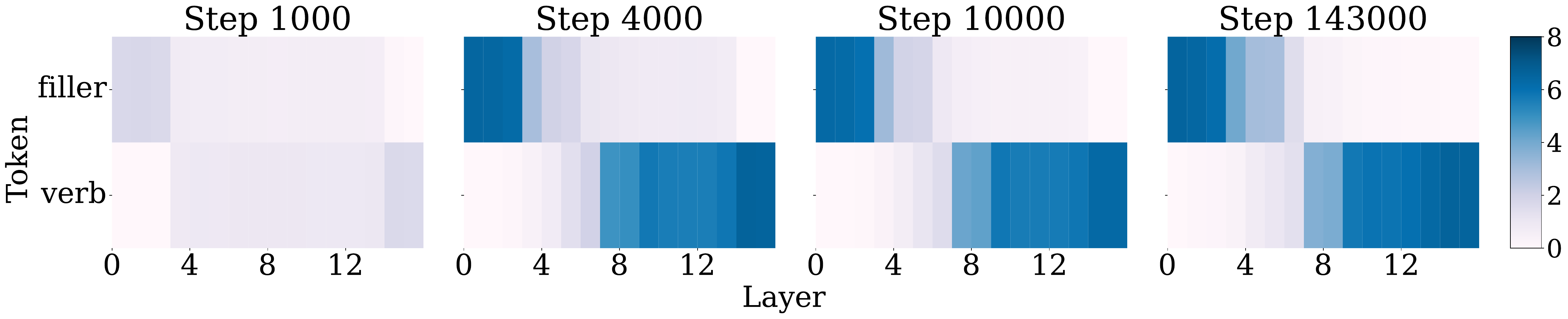}
    \caption{\textsc{EWhK}, Residual stream}
  \end{subfigure}
  \begin{subfigure}{0.8\linewidth}
    \centering
    \includegraphics[width=\linewidth]{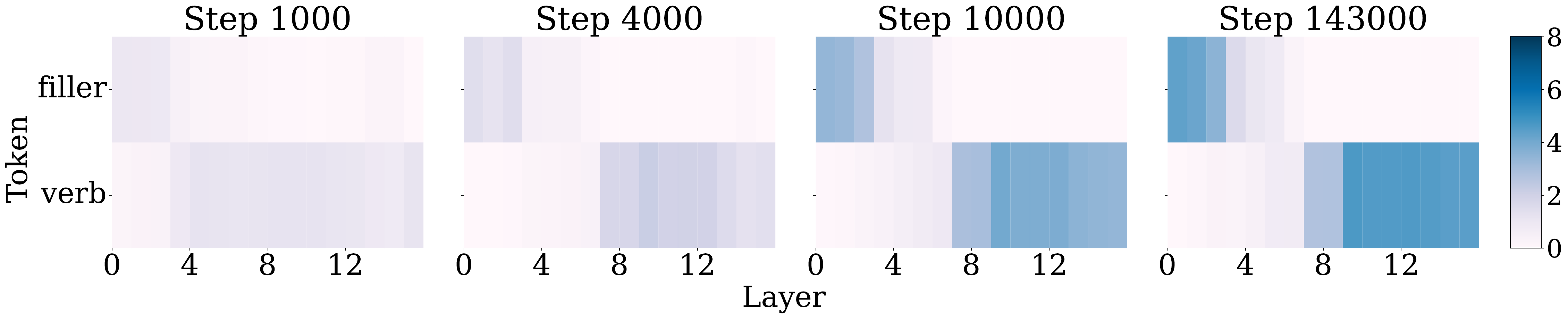}
    \caption{\textsc{PCleft}, Residual stream}
  \end{subfigure}
  \begin{subfigure}{0.8\linewidth}
    \centering
    \includegraphics[width=\linewidth]{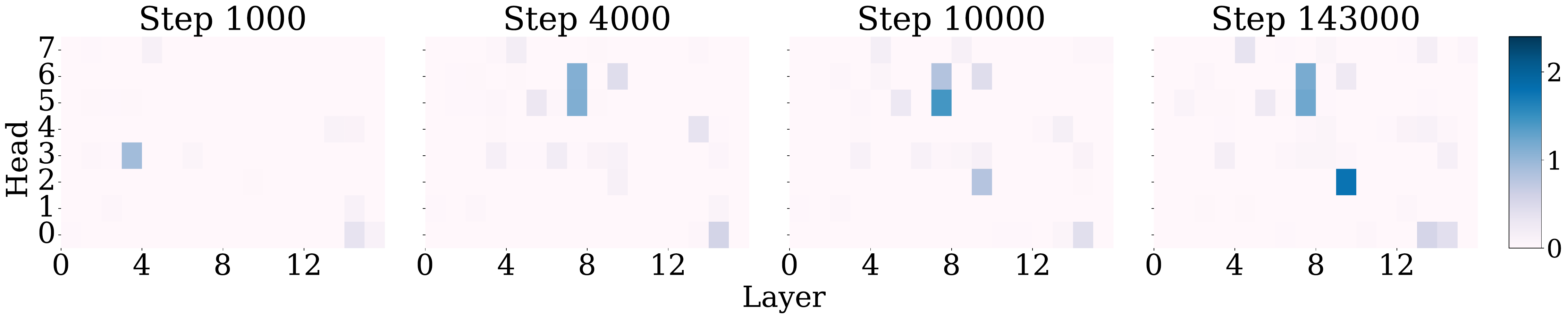}
    \caption{\textsc{EWhK}, Attention head}
  \end{subfigure}
  \begin{subfigure}{0.8\linewidth}
    \centering
    \includegraphics[width=\linewidth]{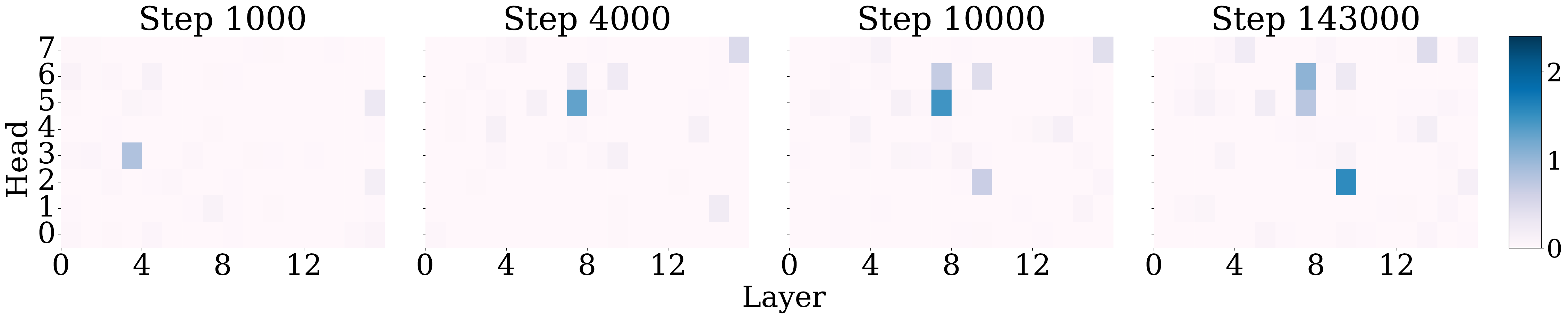}
    \caption{\textsc{PCleft}, Attention head}
  \end{subfigure}

  \caption{\textsc{Odds} scores with activation patching of the residual stream in Pythia 1B with various training steps in filler-gap dependencies.}
  \label{fig:act_patch_pattern_dynamics}
\end{figure*}

Figure~\ref{fig:act_patch_fg} illustrates the results of the activation patching for each component.
The distribution of \textsc{Odds} scores across components is largely consistent across different FGD constructions, whereas the control pattern exhibits a distinct trend.
Specifically, we observe that attention heads contributing to the prediction at the final token are sparsely located in the middle layers, particularly heads 7.5, 7.6, and 9.2, and this localization is shared across constructions. 
Furthermore, the \textsc{Odds} scores of the residual stream at the final token show a sharp increase around the seventh layer, aligning with the depth of these heads.
We also observe the same trend in naturally occurring sentences (see details in Appendix~\ref{subsec:natural}).

This pattern reflects a consistent information flow across FGD constructions. The filler token is processed in the early layers, with MLP blocks encoding its syntactic role, and this information is then transferred to the final (verb) token via attention in the middle layers, where the residual stream scores increase. In the later layers, MLP blocks further process this information to produce the final output. The sparsity and localization of contributing attention heads, together with this flow from filler to final token, are consistently observed across constructions.

In contrast, a slightly different trend was observed in the constructions regarding NPI.
The distribution of \textsc{Odds} scores varies across constructions, indicating differences in the underlying mechanisms.
For example, in \textsc{DNeg}, some contributing attention output are located in earlier layers, leading to earlier propagation of information and a corresponding increase in residual stream scores at earlier layers.
In \textsc{Cond} and \textsc{Sup}, the highest-scoring attention and MLP blocks at the final token differ across constructions.
These variations suggest that, although similar types of components are involved, their functional roles differ across NPI constructions.
A possible explanation is that NPI licensing requires integrating both syntactic and semantic information, which varies across constructions, and thus does not exhibit a single shared mechanism such as a monotonicity-based account~\citep{jumelet-etal-2021-language} in decoder-based LMs.

\subsection{Comparison with Different Training Steps}

\begin{figure*}[ht]
\centering

  \begin{subfigure}{0.48\linewidth}
    \centering
    \includegraphics[width=\linewidth]{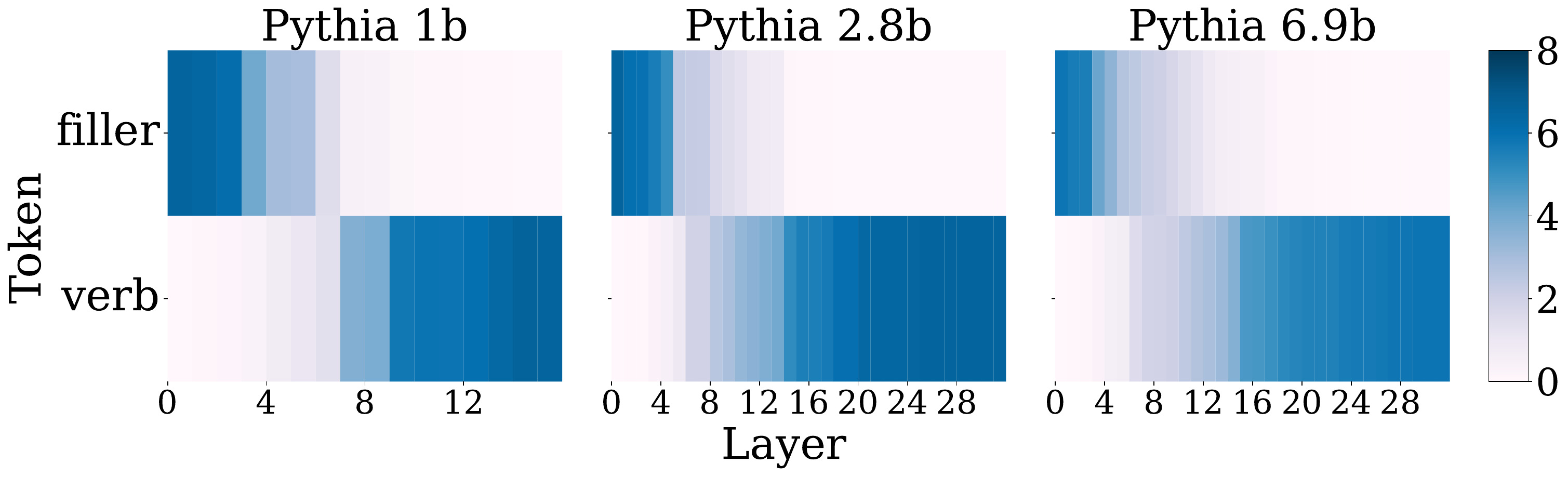}
    \caption{\textsc{EWhK}, Residual stream}
  \end{subfigure}
  \begin{subfigure}{0.48\linewidth}
    \centering
    \includegraphics[width=\linewidth]{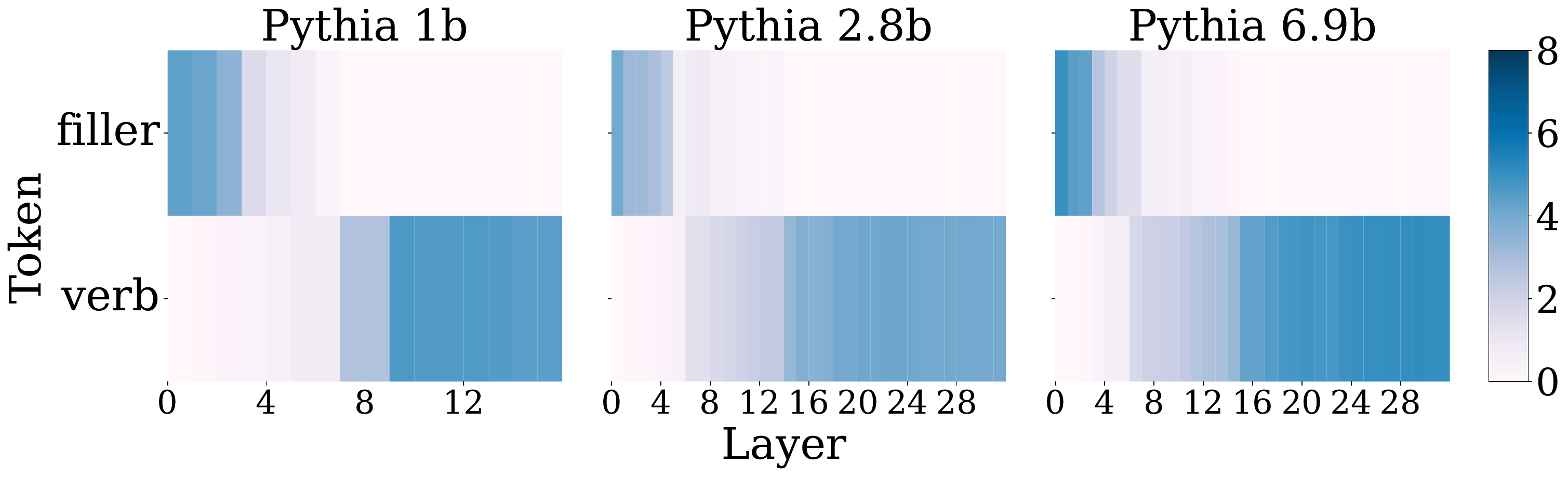}
    \caption{\textsc{PCleft}, Residual stream}
  \end{subfigure}
  \begin{subfigure}{0.48\linewidth}
    \centering
    \includegraphics[width=\linewidth]{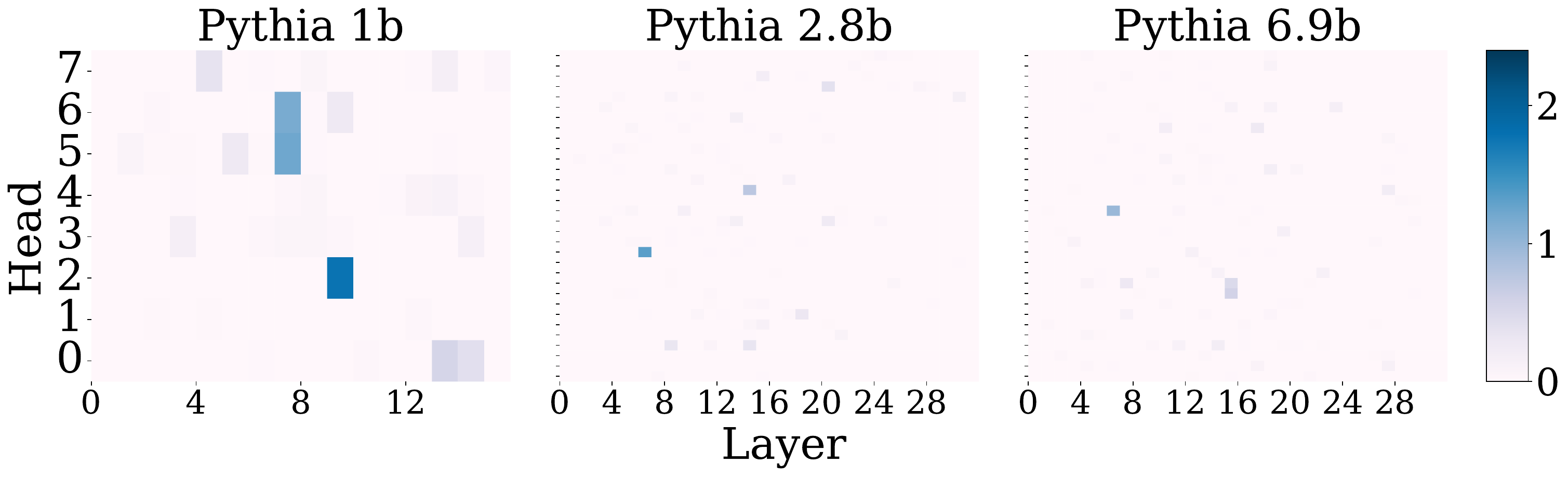}
    \caption{\textsc{EWhK}, Attention head}
  \end{subfigure}
  \begin{subfigure}{0.48\linewidth}
    \centering
    \includegraphics[width=\linewidth]{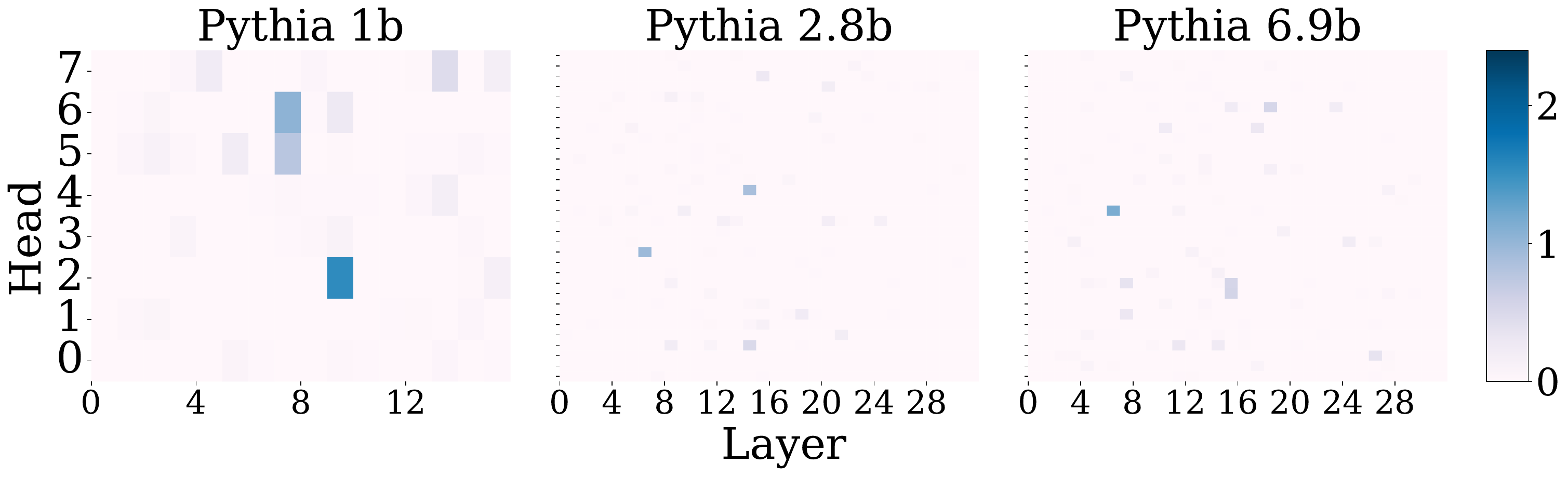}
    \caption{\textsc{PCleft}, Attention head}
  \end{subfigure}

  \caption{\textsc{Odds} scores with activation patching of residual stream in Pythia models with various sizes in filler-gap dependencies.}
  \label{fig:act_patch_pattern_sizes}
\end{figure*}

We next analyze the emergence of the mechanisms in FGDs during training, specifically focusing on the activation patching of the residual stream and attention heads in the context of embedded wh-questions \textsc{EWhK} and pseudo-clefts \textsc{PCleft}.
According to \citet{boguraev-etal-2025-causal}, embedded wh-questions occur approximately 15 times more frequently than pseudo-clefts in the English-EWT Universal Dependencies dataset~\citep{de-marneffe-etal-2021-universal, nivre-etal-2020-universal, silveira-etal-2014-gold}.
We investigate whether this discrepancy in frequency impacts the training dynamics.

The results are illustrated in Figure~\ref{fig:act_patch_pattern_dynamics}.
The \textsc{Odds} scores in both constructions increase monotonically with the training steps.
In particular, scores in the earlier layers improve more rapidly than those in the later layers, suggesting that the mechanism is established hierarchically, beginning with lower-level representations.
However, the two constructions differ in their convergence.
In the residual stream, the scores in \textsc{EWhK} reach levels comparable to the final training step by step 4,000, while the scores in \textsc{PCleft} at step 10,000 remain lower than their final values.
A similar trend was observed in attention heads, although the improvement of the scores was quicker in both constructions compared to that in the residual stream.

These findings suggest that high-frequency constructions are learned during the initial stages of training, whereas less frequent constructions are acquired more gradually.
Although both constructions eventually converge upon a shared mechanism, the development toward that mechanism depends on their frequency.
This trend was consistently observed across other constructions and MLP components (see Appendix~\ref{sec:full_dynamics} for details).

\subsection{Comparison with Different Parameter Sizes and Model Family}

Next, we analyze how the mechanism varies with respect to the parameter size of the models.
Figure~\ref{fig:act_patch_pattern_sizes} shows the results.
It can be seen that the models with more layers process FGDs in earlier layers, while the mechanism is still shared and localized in each parameter size.
Also, the distribution of the scores was similar between Pythia 2.8B and 6.9B.
It indicates that the hidden dimension of the model, which is the only architectural difference between 2.8B and 6.9B, does not impact the processing mechanism of FGDs.

We also saw similar trends in Gemma 3 1B and 12B, which have 10 and 32 more layers than Pythia 1B respectively, where contributing attention heads are located in earlier layers, and the \textsc{Odds} scores increase in earlier layers than Pythia 1B, while there is a shared mechanism across FGD constructions (see full results in Appendix~\ref{sec:gemma}).

\section{Steering Based on the Mechanism}
\label{sec:steering}
\subsection{Method}
The previous sections revealed a shared mechanism underlying FGDs.
However, it is possible that the analysis failed to distinguish heuristic patterns from the core syntactic phenomena if the dataset or methods were insufficiently robust.
To address this, we investigate whether model behavior changes in an existing targeted syntactic evaluation benchmark when we manipulate only the specific components identified as contributors to the found mechanism.
Specifically, we scale the activation values of the attention heads 7.5, 7.6, and 9.2 by a factor $\alpha$ and evaluate the resulting performance in BLiMP~\citep{warstadt-etal-2020-blimp-benchmark}.

BLiMP assesses the extent to which an LM assigns a higher probability to a grammatical sentence compared to its ungrammatical counterpart within a minimal pair.
The BLiMP task setting, which calculates probability over an entire sentence, allows us to verify that the identified mechanisms are not merely localized heuristics specific to individual tokens.
It should be noted that we restrict our analysis to minimal pairs with identical tokenized lengths.
This is necessary because different token lengths have been shown to skew the probability assignments, thereby hindering an accurate evaluation of the model's capability~\citep{ueda-etal-2024-token}.

\subsection{Results}
\begin{figure*}[ht]
\centering

  \begin{subfigure}{0.49\linewidth}
    \centering
    \includegraphics[width=\linewidth]{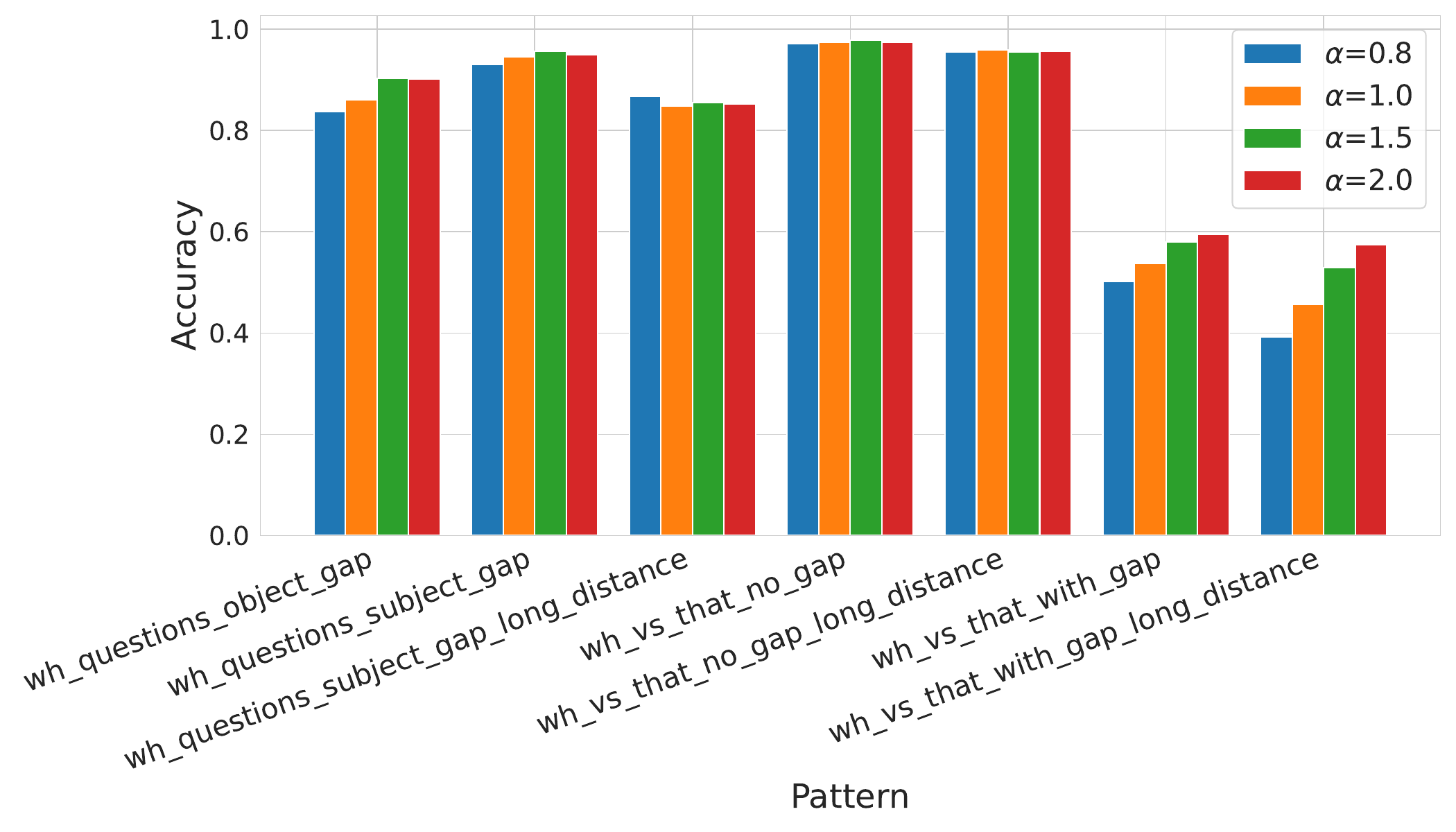}
    \caption{\textsc{Filler gap} category}
     \label{fig:blimp_fg}
  \end{subfigure}
  \begin{subfigure}{0.49\linewidth}
    \centering
    \includegraphics[width=\linewidth]{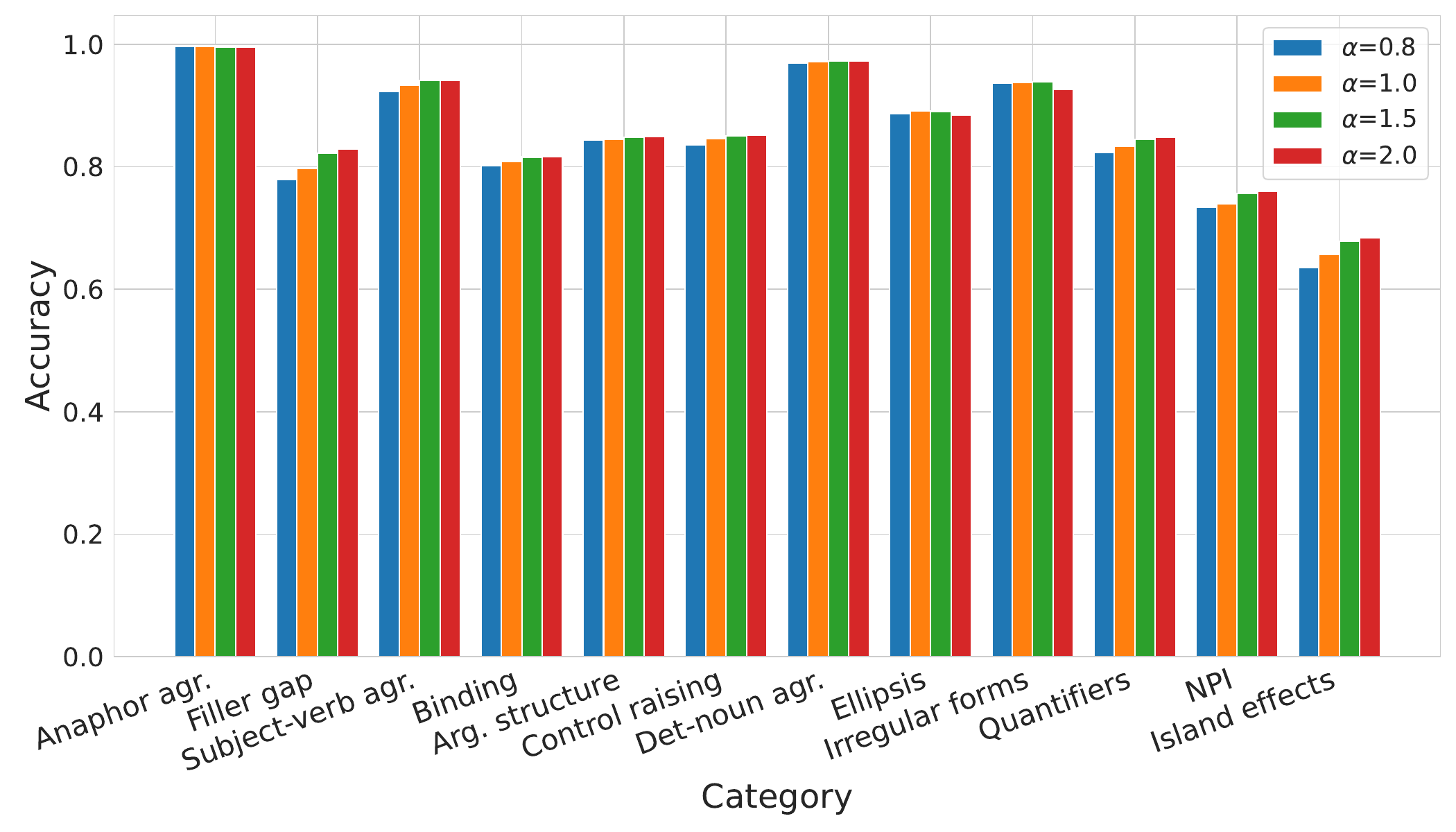}
    \caption{All categories}
    \label{fig:blimp_all}
  \end{subfigure}

  \caption{Accuracy in the category involving filler-gap dependencies (left) and in all categories (right) in BLiMP after multiplying activations of several attention heads in Pythia 1B by $\alpha$.}
  \label{fig:steer_blimp}
\end{figure*}

First, we focus on how the model performance changed in the categories involving FGDs.
As illustrated in Figure~\ref{fig:blimp_fg}, amplifying the activation of the three attention heads with $\alpha>1$ improved or maintained performance, the latter occurring in cases of score saturation.
Moreover, for most unsaturated patterns, performance improved monotonically with increasing $\alpha$ from $0.8$ to $1.5$, although there was one exception pattern, where the score was higher with $\alpha=0.8$ than with $\alpha=1.0$.
This result shows that the manipulated attention heads indeed contribute to assigning higher probabilities to grammatical sentences involving FGDs, not only in those with an object gap that we use for analysis, but also those with a subject gap, a prepositional object gap, or long-distance dependencies.

We next examine the model performance in other categories, as shown in Figure~\ref{fig:blimp_all}.
There were noticeable improvements in categories such as island effects, binding, NPI, quantifiers, and center embedding.
Also, the improvements in subject-verb agreement in BLiMP all came from the patterns involving agreement beyond the distractors with prepositional phrases or relative clauses.
The results thus indicate that the attention heads manipulated here might serve a more general function in capturing hierarchical structural dependency rather than being specialized solely for FGDs.

We also tested with SyntaxGym~\citep{hu-etal-2020-systematic}, and the overall trends were similar (see the full results in Appendix~\ref{sec:steering_syntaxgym}).
In addition, to investigate performance gains in downstream tasks that require syntactic dependency processing, we conducted a steering experiment on HANS~\citep{mccoy-etal-2019-right}, an NLI benchmark designed for analyzing syntactic heuristics.
The results show that amplifying the activation values of the same attention heads leads to improved model performance (see details in Appendix~\ref{sec:hans}).

\section{Comparison Between DAS and Activation Patching}
\label{sec:results}
\begin{figure*}
\centering

  \begin{subfigure}{0.32\linewidth}
    \centering
    \includegraphics[width=\linewidth]{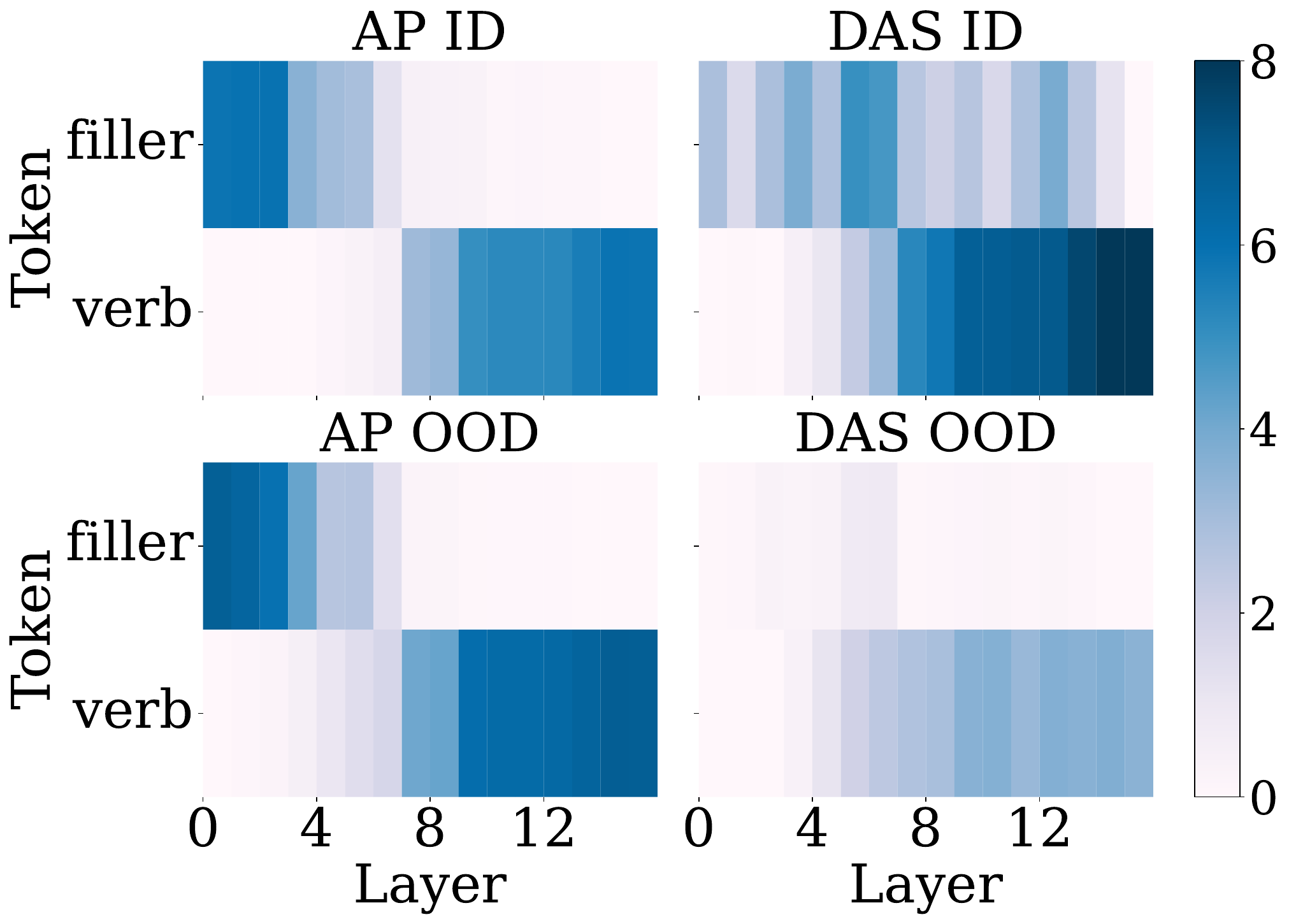}
    \caption{Residual stream}
  \end{subfigure}
  \begin{subfigure}{0.335\linewidth}
    \centering
    \includegraphics[width=\linewidth]{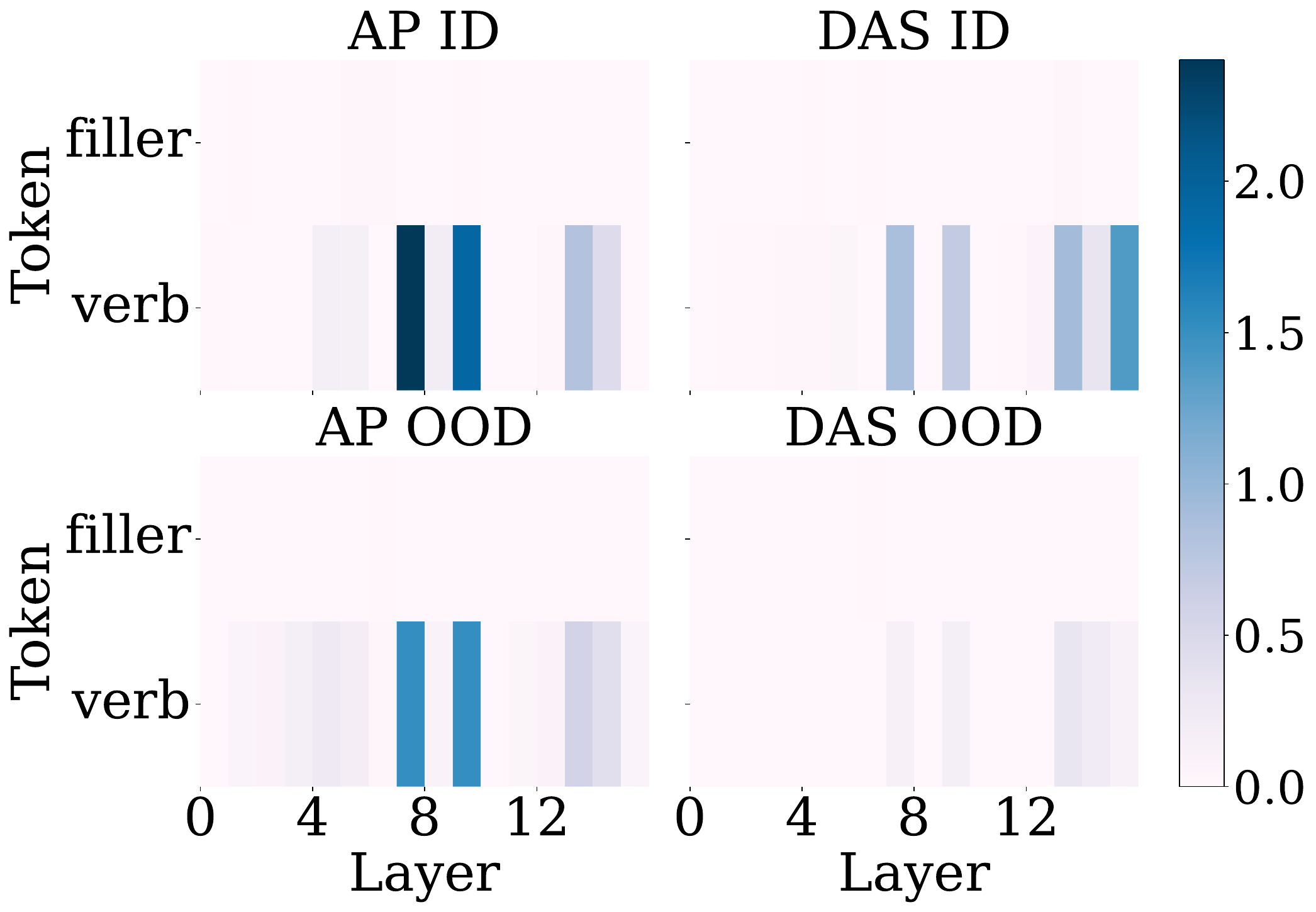}
    \caption{Attention}
  \end{subfigure}
  \begin{subfigure}{0.32\linewidth}
    \centering
    \includegraphics[width=\linewidth]{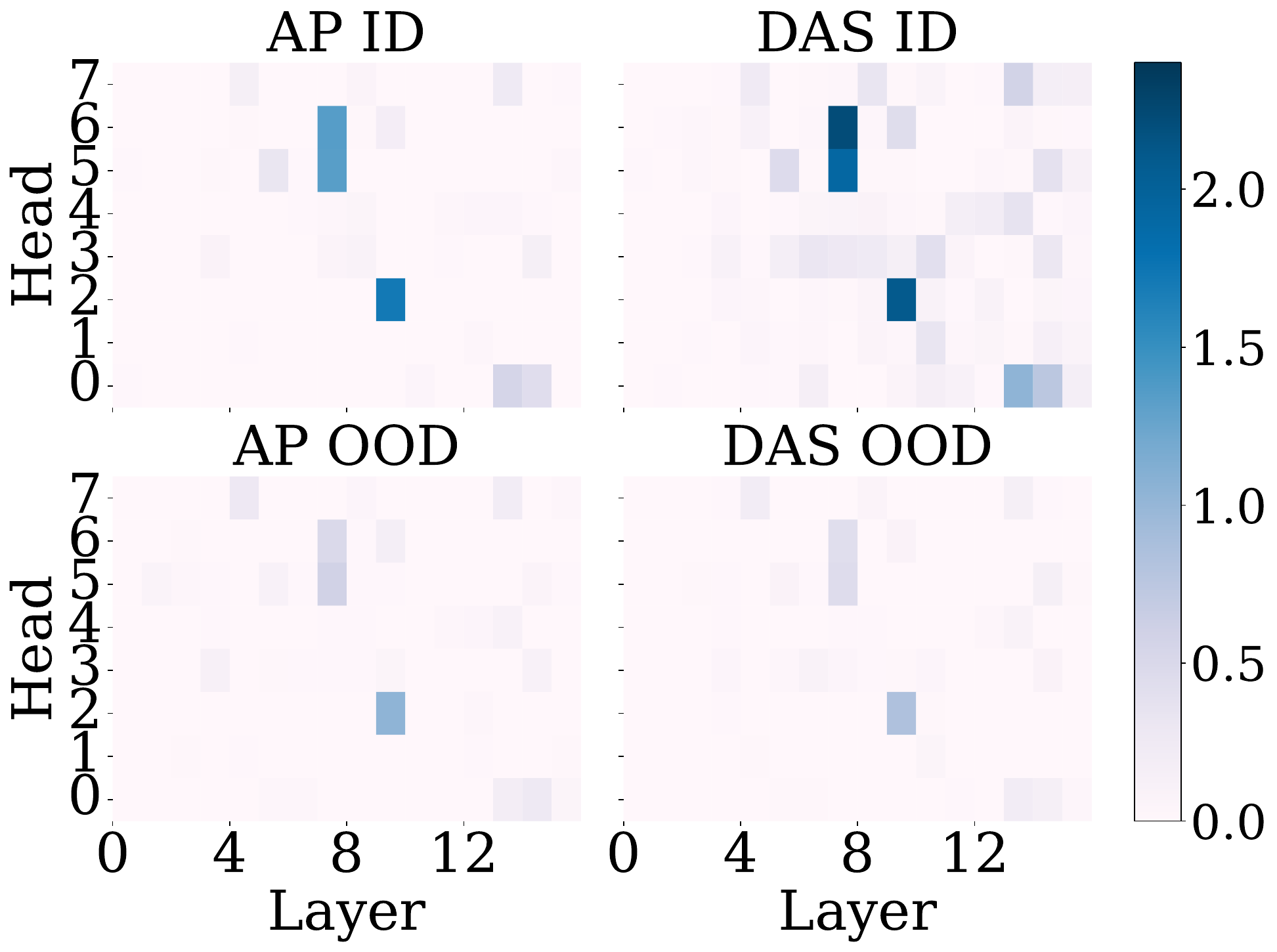}
    \caption{Attention head}
  \end{subfigure}

  \caption{Comparison of the \textsc{Odds} scores of Pythia 1B between activation patching (AP) and DAS, evaluated on the \textsc{EWhK} ID and OOD test sets.}
  \label{fig:act_patch_fg_generalization_pattern}
\end{figure*}

\subsection{Distributed Alignment Search}
To assess the robustness of our findings, we compare activation patching with DAS, which is used by previous studies~\citep{arora-etal-2024-causalgym,boguraev-etal-2025-causal} for identifying causally efficacious components in syntactic tasks.
DAS~\citep{pmlr-v236-geiger24a} is a supervised causal localization method designed to identify subspaces that mediate specific task behavior.
Following \citet{arora-etal-2024-causalgym} and \citet{boguraev-etal-2025-causal}, we employ the one-dimensional variant of DAS, which learns a direction that causally impacts the output distribution.
The intervention with DAS is defined by the following transformation: 
\begin{equation*}
f_{\mathrm{interv}}(\bm{b},\bm{s}) = f(\bm{b}) + (f(\bm{s})\bm{a}-f(\bm{b})\bm{a})\bm{a}^T,
\end{equation*}
where $\bm{a}$ is the vector to be learned.

During the training, the objective is to optimize $\bm{a}$ such that the intervention shifts the model's output toward the correct token associated with the source input.
The loss function is defined as:
\begin{equation*}
\min_{\bm{a}}\left\{-\sum_{(\bm{b},\bm{s},y_b,y_s)\in D} \log p_{f_{\mathrm{interv}}}(y_s|\bm{b}, \bm{s})\right\},
\end{equation*}
where $D$ denotes the training set.

Following \citet{boguraev-etal-2025-causal}, we apply DAS to analyze shared mechanisms through a leave-one-out training paradigm.
In this setup, the direction is trained using a subset of the constructions $\{c_j\mid j\neq i, 0\leq i,j\leq n\}$ and then evaluated on the remaining held-out construction $c_i$.
This approach allows us to assess whether the learned direction for one set of constructions generalizes to another, thereby measuring the similarity of the mechanism across constructions.

\subsection{Results}

Figure~\ref{fig:act_patch_fg_generalization_pattern} presents the comparative results between DAS and activation patching.
First, we observe that DAS fails to generalize to the OOD test set, whereas training-free activation patching revealed a consistent distribution of scores between the ID and OOD test sets.
This discrepancy suggests that the direction learned in DAS is likely overfit to the training set, capturing features specific to the dataset rather than the structural dependencies.
Second, the scores obtained when learning a direction in attention output with DAS were lower than those with activation patching, even in the ID test set.
This trend was more salient in the early to middle layers compared to the late ones.
It significantly changes the interpretation of the mechanism, as otherwise, we would see more active contribution in the components of later layers in DAS.
Interestingly, applying DAS to attention heads yields results more closely aligned with activation patching, despite lower \textsc{Odds} scores in the OOD test set and a slight overestimation of causal efficacy in the later layers.
This indicates that DAS may be more suitable for learning a direction in the representation space of individual attention heads.

These results suggest that interpretability methods requiring supervision necessitate rigorous OOD evaluation to ensure that the identified mechanism is functionally relevant to the task.
Moreover, using DAS in datasets with relatively limited diversity may not always detect the most relevant mechanisms.
The full comparison with DAS is provided in Appendix~\ref{sec:full_das}, and the above trend was consistent in other constructions.

\section{Conclusion}
\label{sec:conclusion}

This paper investigated the internal mechanisms of LMs to determine whether they utilize shared mechanisms across different syntactic constructions.
We identified the existence of a shared mechanism for FGDs, primarily mediated by a small number of attention heads located in the early to middle layers.
In contrast, NPI licensing appears to rely on construction-specific mechanisms, reflecting its complex syntactic and semantic requirements.
We also validated the causal efficacy of the identified components through steering experiments, where activation scaling consistently improved model performance on BLiMP.
Furthermore, we compared activation patching and DAS, which revealed poor OOD generalization with DAS.
These insights not only advance our understanding of how LMs process language but also can provide a basis for developing methods to improve the syntactic capability of LMs.

\section*{Limitations}
\label{sec:limitation}
First, our study is restricted to English, a limitation shared by much of the current work on mechanistic interpretability.
It remains unclear whether our findings generalize to other languages with different linguistic features.
For example, factors such as word order may require different processing strategies.
Moreover, as the frequency of each language in the pretraining corpus likely impacts the internal mechanisms of language models, models may employ different mechanisms for low-resource languages compared to English.
We leave these questions to future work.

Second, our dataset used for our main analysis is synthetically generated rather than sampled from natural corpora, following previous studies~\citep{arora-etal-2024-causalgym,boguraev-etal-2025-causal}.
Although this approach allows for the precise control of minimal pairs, it limits the diversity of language expressions observed in the wild.
However, we made efforts to diversify the dataset compared to the previous studies~\citep{arora-etal-2024-causalgym,boguraev-etal-2025-causal}, and conducted experiments by using a small set of naturally occurring sentences.

\section*{Ethical Considerations}
Our study investigates the internal workings of language models for a deeper understanding of their language processing and ultimately utilizes these insights to improve model capabilities and reliability.
However, we acknowledge that such insights could potentially be misused to modify the behavior of models in unsafe or harmful ways.
We warn model developers and users against the malicious application of these techniques.

The dataset does not contain any information that names or uniquely identifies individual people or offensive content, as it is generated with our templates.

\section*{Acknowledgments}
We appreciate the anonymous reviewers on ACL Rolling Review, who helped refine this paper from various perspectives.
We also thank Adam Nohejl, Taisei Yamamoto, Amane Watahiki, Tomoki Doi, Daiki Matsuoka, Gaëtan Margueritte, and Koki Ryu for their helpful comments on this paper.
This work was supported by JST CREST Grant Number JPMJCR2565 and JST BOOST Program Grant Number JPMJBY24H5, Japan.

\bibliography{custom,anthology-1,anthology-2}

\appendix

\section{Details of the Dataset}
\label{sec:stats_dataset}

We adopt the dataset sizes from the setting used by \citet{arora-etal-2024-causalgym}.
For each construction pattern, the training set consists of 200 examples, and each of the ID and OOD tests contains 50 examples.
To maintain a strict balance between labels, we swap the base and source inputs and labels to double the size.
Thus, our combined analysis across the ID and OOD test sets utilizes 200 examples in total, which is a larger sample size than the test sets used in related work~\citep{arora-etal-2024-causalgym,boguraev-etal-2025-causal}.

The sentences are generated using a vocabulary set of 4,275 words, which was originally created by \citet{warstadt-etal-2020-blimp-benchmark}.
To properly test generalization, we create two templates for each construction, one for the training and ID test sets and another for the OOD test set.
This ensures that the model is evaluated on a wide range of lexical and structural variations.

\section{Hyperparameters}
\label{sec:hyper}

In the training of DAS, we use the hyperparameters that \citet{arora-etal-2024-causalgym} used based on tuning.
We set learning rate as $5\times 10^{-3}$ with linear warmup until 10\% of the training steps.
We set the batch size to 4 and train for 100 steps.
We also experimented with other learning rates ($5\times 10^{-2}, 5\times 10^{-4}$) and training steps ($50, 200$) and observed largely consistent trends across settings.

\section{Details of the \textsc{Odds} score}
\label{sec:a_metric}

In this section, we demonstrate that for FGDs, our \textsc{Odds} score is mathematically equivalent to the one used by related work~\citep{arora-etal-2024-causalgym,boguraev-etal-2025-causal}, which is defined as follows:
\begin{align*}
&\mathrm{\textsc{Odds}^\ast}(p,p_{f_{\mathrm{interv}}},\langle \bm{b},\bm{s},y_b,y_s\rangle)\\&=\log\left(\frac{p(y_b|\bm{b})}{p(y_s|\bm{b})}\frac{p_{f_{\mathrm{interv}}}(y_s|\bm{b},\bm{s})}{p_{f_{\mathrm{interv}}}(y_b|\bm{b},\bm{s})}\right)
\end{align*}

As described in Appendix~\ref{sec:stats_dataset}, our dataset includes both the original minimal pair $(\bm{b},\bm{s}, y_b, y_s)$ and its flipped counterpart $(\bm{b}, \bm{s}, y_b, y_s)$ and ($\bm{s}, \bm{b}, y_s, y_b)$.
Summing the \textsc{Odds} scores for these two samples yields the following derivation:
\begin{align*}
&\mathrm{\textsc{Odds}}(p,p_{f_{\mathrm{interv}}},\langle \bm{b},\bm{s},y_b,y_s\rangle)\\
&\quad+\mathrm{\textsc{Odds}}(p,p_{f_{\mathrm{interv}}},\langle \bm{s},\bm{b},y_s,y_b\rangle)\\&=\log\left(\frac{p(y_b|\bm{b})}{p(y_b|\bm{s})}\frac{p_{f_{\mathrm{interv}}}(y_b|\bm{s},\bm{b})}{p_{f_{\mathrm{interv}}}(y_b|\bm{b},\bm{s})}\right) \\
&\quad+ \log\left(\frac{p(y_s|\bm{s})}{p(y_s|\bm{b})}\frac{p_{f_{\mathrm{interv}}}(y_s|\bm{b},\bm{s})}{p_{f_{\mathrm{interv}}}(y_s|\bm{s},\bm{b})}\right)\\
&=\log\left(\frac{p(y_b|\bm{b})}{p(y_s|\bm{b})}\frac{p_{f_{\mathrm{interv}}}(y_s|\bm{b},\bm{s})}{p_{f_{\mathrm{interv}}}(y_b|\bm{b},\bm{s})}\right) \\
&\quad+ \log\left(\frac{p(y_s|\bm{s})}{p(y_b|\bm{s})}\frac{p_{f_{\mathrm{interv}}}(y_b|\bm{s},\bm{b})}{p_{f_{\mathrm{interv}}}(y_s|\bm{s},\bm{b})}\right) \\
&=\mathrm{\textsc{Odds}^\ast}(p,p_{f_{\mathrm{interv}}},\langle \bm{b},\bm{s},y_b,y_s\rangle)\\
&\quad+\mathrm{\textsc{Odds}^\ast}(p,p_{f_{\mathrm{interv}}},\langle \bm{s},\bm{b},y_s,y_b\rangle).
\end{align*}

Therefore, our \textsc{Odds} score is equivalent to the one defined in previous studies when we use symmetric pairs.
It should be noted that because we use only one of the two for NPI licensing, as mentioned in Section~\ref{subsec:dataset}, the \textsc{Odds} score in NPI licensing differs from the previous one.

\section{Detailed Activation Patching Results}
\label{sec:full_ap}
\subsection{All Constructions in FGDs}
\begin{figure*}
\centering

  \begin{subfigure}{\linewidth}
    \centering
    \includegraphics[width=\linewidth]{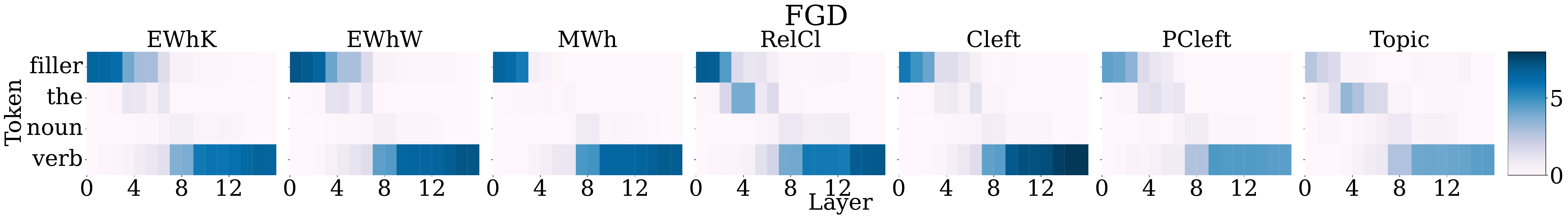}
    \caption{Residual stream}
  \end{subfigure}
  \begin{subfigure}{\linewidth}
    \centering
    \includegraphics[width=\linewidth]{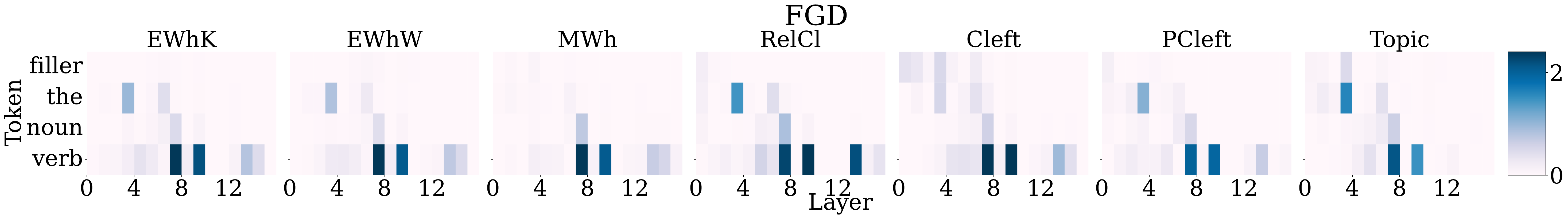}
    \caption{Attention}
  \end{subfigure}
  \begin{subfigure}{\linewidth}
    \centering
    \includegraphics[width=\linewidth]{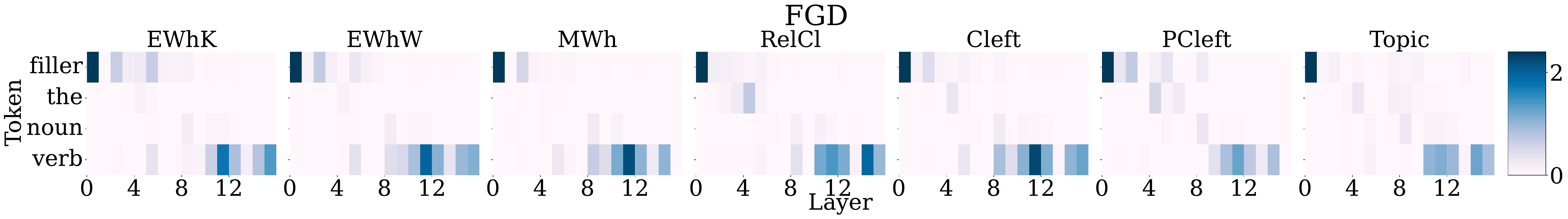}
    \caption{MLP}
  \end{subfigure}
  \begin{subfigure}{\linewidth}
    \centering
    \includegraphics[width=\linewidth]{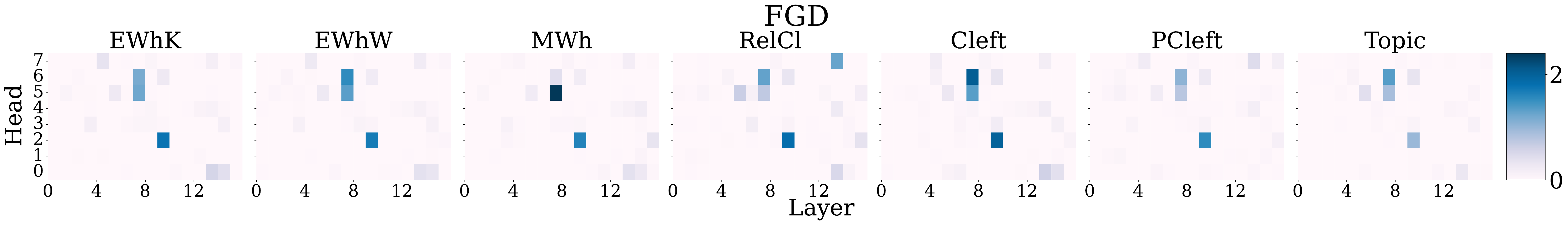}
    \caption{Attention head}
  \end{subfigure}

  \caption{\textsc{Odds} scores with activation patching of Pythia 1B in all the constructions in FGDs.}
  \label{fig:act_patch_all_fgd}
\end{figure*}

The results are shown in Figure~\ref{fig:act_patch_all_fgd}.
\subsection{All Constructions in NPI Licensing}
\begin{figure*}
\centering

  \begin{subfigure}{\linewidth}
    \centering
    \includegraphics[width=\linewidth]{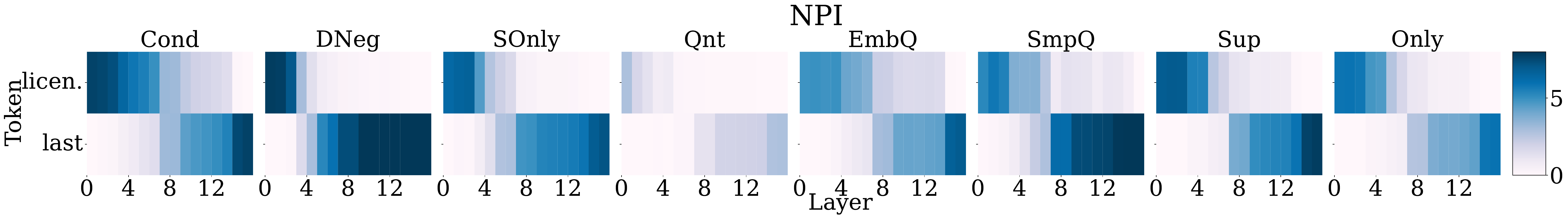}
    \caption{Residual Stream}
  \end{subfigure}
  \begin{subfigure}{\linewidth}
    \centering
    \includegraphics[width=\linewidth]{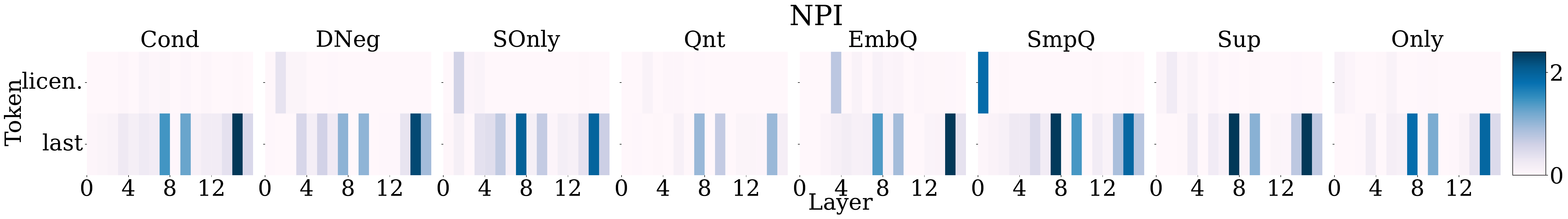}
    \caption{Attention}
  \end{subfigure}
  \begin{subfigure}{\linewidth}
    \centering
    \includegraphics[width=\linewidth]{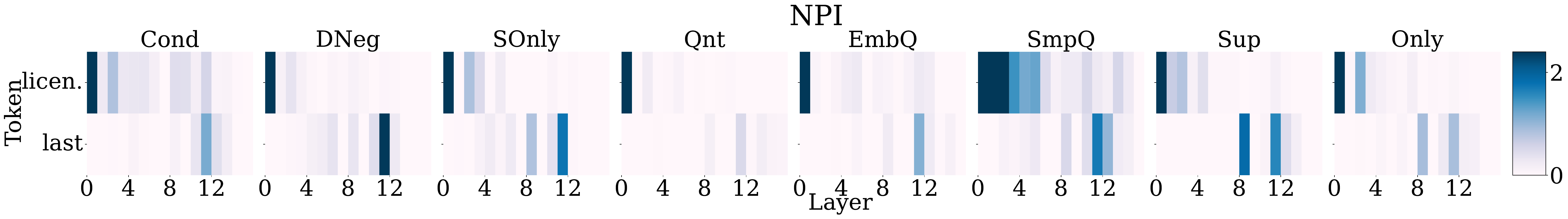}
    \caption{MLP}
  \end{subfigure}
  \begin{subfigure}{\linewidth}
    \centering
    \includegraphics[width=\linewidth]{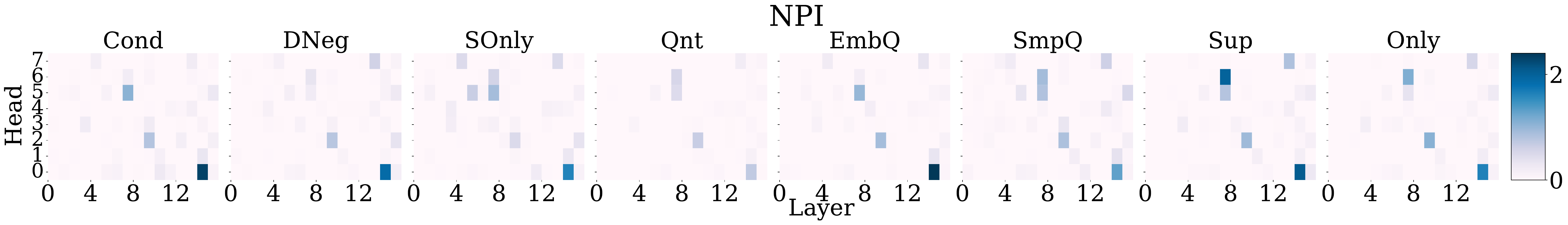}
    \caption{Attention Heads}
  \end{subfigure}

  \caption{\textsc{Odds} scores with activation patching of Pythia 1B in all the constructions of NPI licensing.}
  \label{fig:act_patch_npi}
\end{figure*}

The results are shown in Figure~\ref{fig:act_patch_npi}.

\subsection{Training Dynamics of All Constructions in FGDs}
\label{sec:full_dynamics}
\begin{figure*}
\centering

  \begin{subfigure}{\linewidth}
    \centering
    \includegraphics[width=\linewidth]{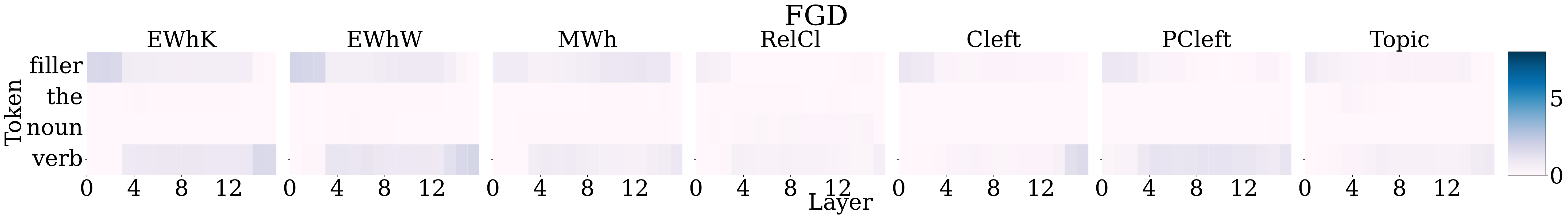}
    \caption{Training Steps: 1000}
  \end{subfigure}
  \begin{subfigure}{\linewidth}
    \centering
    \includegraphics[width=\linewidth]{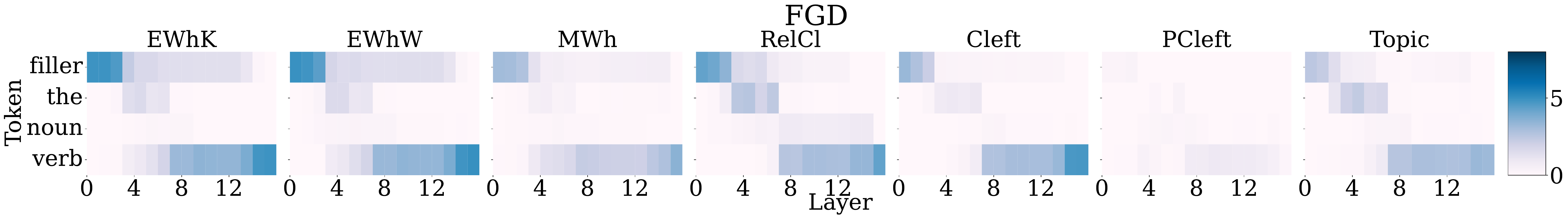}
    \caption{Training Steps: 2000}
  \end{subfigure}
  \begin{subfigure}{\linewidth}
    \centering
    \includegraphics[width=\linewidth]{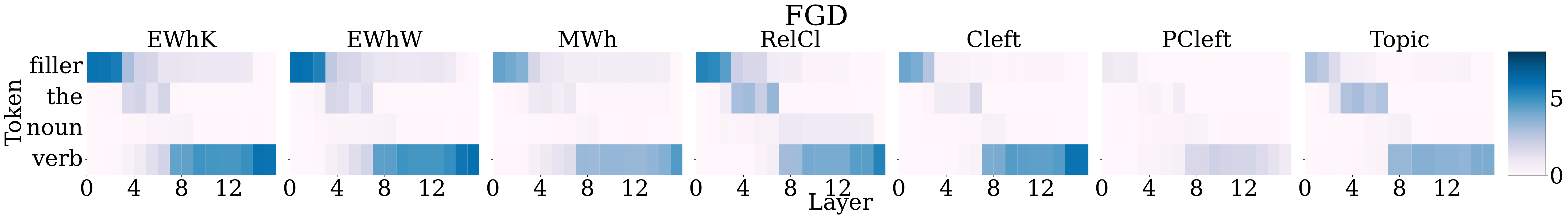}
    \caption{Training Steps: 3000}
  \end{subfigure}
  \begin{subfigure}{\linewidth}
    \centering
    \includegraphics[width=\linewidth]{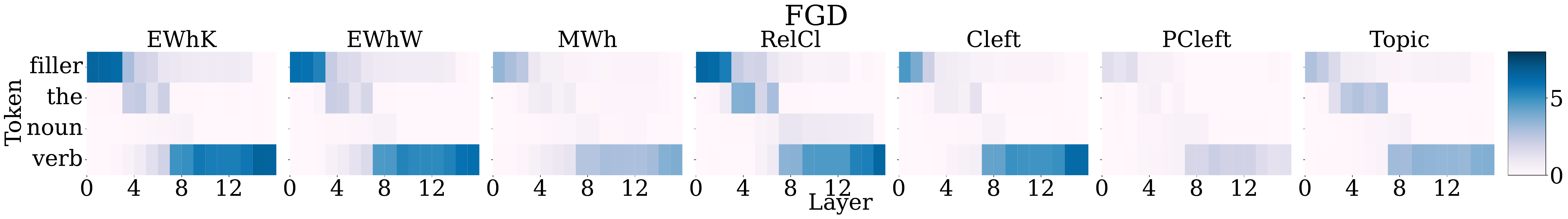}
    \caption{Training Steps: 4000}
  \end{subfigure}
  \begin{subfigure}{\linewidth}
    \centering
    \includegraphics[width=\linewidth]{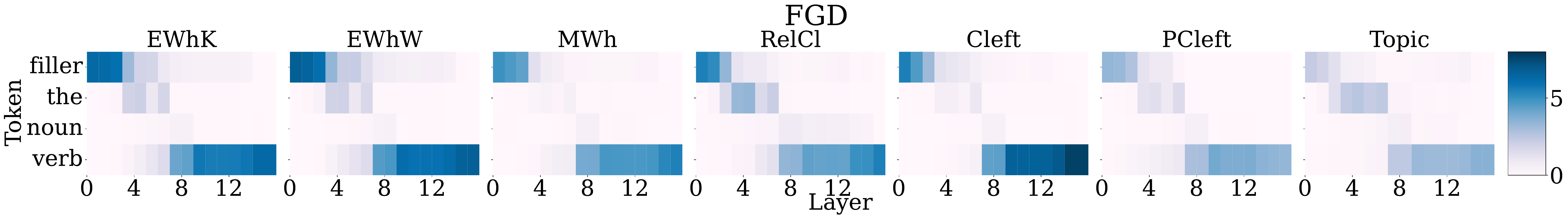}
    \caption{Training Steps: 10000}
  \end{subfigure}
  \begin{subfigure}{\linewidth}
    \centering
    \includegraphics[width=\linewidth]{sections/figures/results/odds_ratio_pythia_1b_143000_resid_vanilla_repr_fg.pdf}
    \caption{Training Steps: 143000 (Final)}
  \end{subfigure}

  \caption{\textsc{Odds} scores with activation patching of residual stream of models with various training steps in filler-gap dependencies.}
  \label{fig:act_patch_fg_resid_dynamics}
\end{figure*}

\begin{figure*}
\centering

  \begin{subfigure}{\linewidth}
    \centering
    \includegraphics[width=\linewidth]{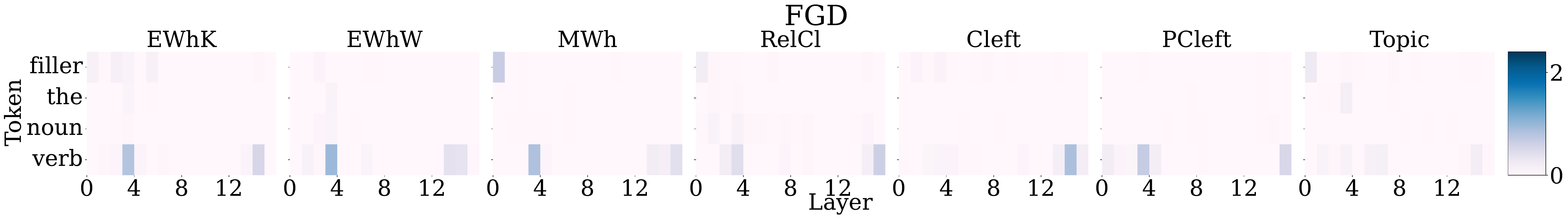}
    \caption{Training Steps: 1000}
  \end{subfigure}
  \begin{subfigure}{\linewidth}
    \centering
    \includegraphics[width=\linewidth]{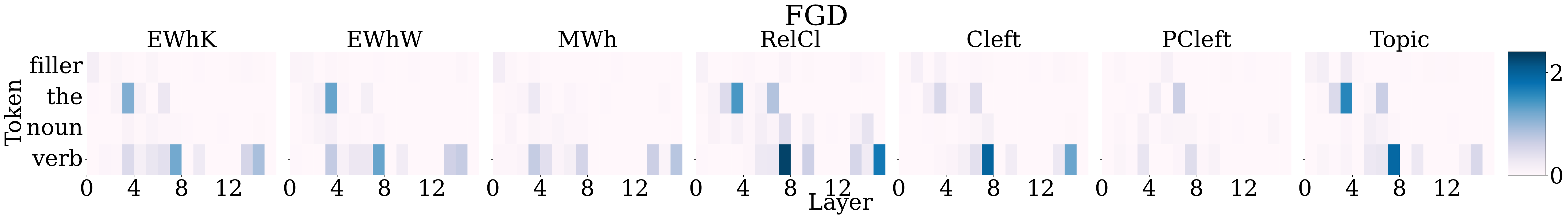}
    \caption{Training Steps: 2000}
  \end{subfigure}
  \begin{subfigure}{\linewidth}
    \centering
    \includegraphics[width=\linewidth]{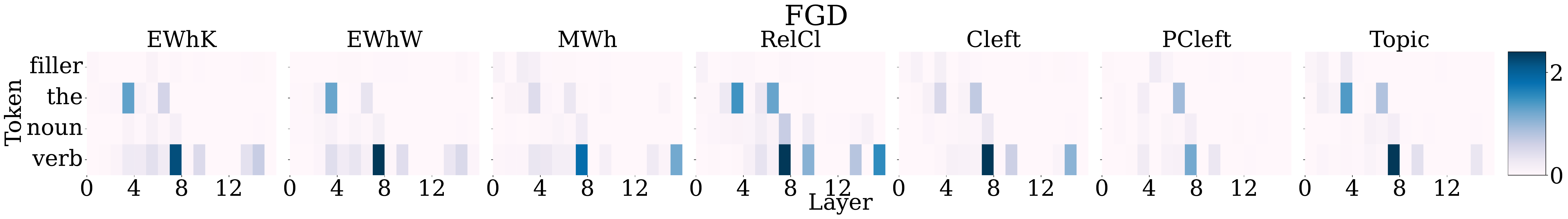}
    \caption{Training Steps: 3000}
  \end{subfigure}
  \begin{subfigure}{\linewidth}
    \centering
    \includegraphics[width=\linewidth]{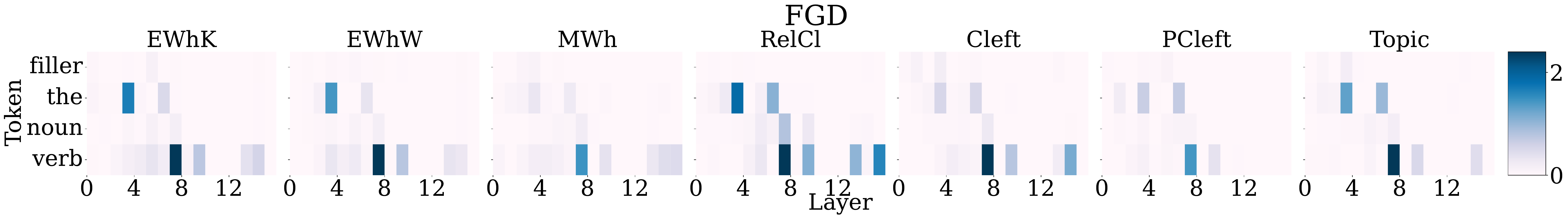}
    \caption{Training Steps: 4000}
  \end{subfigure}
  \begin{subfigure}{\linewidth}
    \centering
    \includegraphics[width=\linewidth]{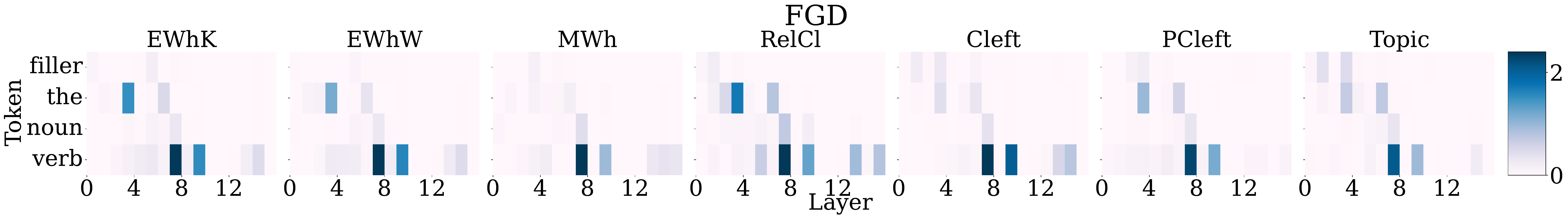}
    \caption{Training Steps: 10000}
  \end{subfigure}
  \begin{subfigure}{\linewidth}
    \centering
    \includegraphics[width=\linewidth]{sections/figures/results/odds_ratio_pythia_1b_143000_attn_vanilla_repr_fg.pdf}
    \caption{Training Steps: 143000 (Final)}
  \end{subfigure}

  \caption{\textsc{Odds} scores with activation patching of attention output of models with various training steps in filler-gap dependencies.}
  \label{fig:act_patch_fg_attn_dynamics}
\end{figure*}

\begin{figure*}
\centering

  \begin{subfigure}{\linewidth}
    \centering
    \includegraphics[width=\linewidth]{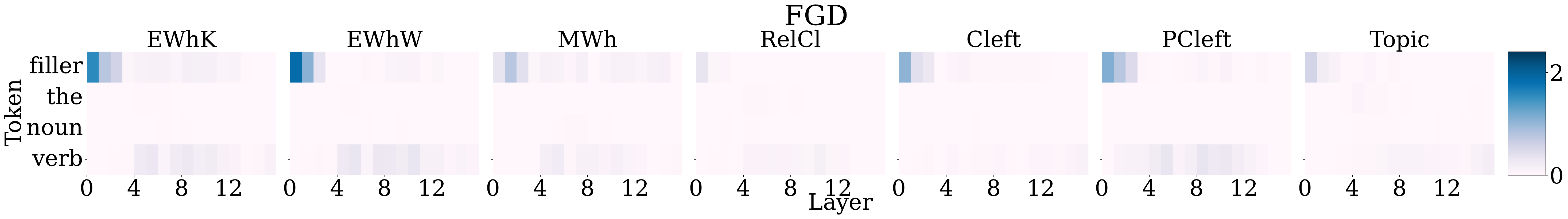}
    \caption{Training Steps: 1000}
  \end{subfigure}
  \begin{subfigure}{\linewidth}
    \centering
    \includegraphics[width=\linewidth]{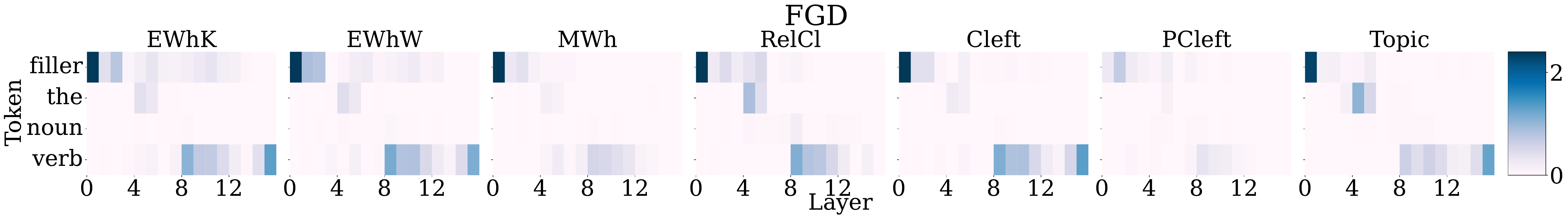}
    \caption{Training Steps: 2000}
  \end{subfigure}
  \begin{subfigure}{\linewidth}
    \centering
    \includegraphics[width=\linewidth]{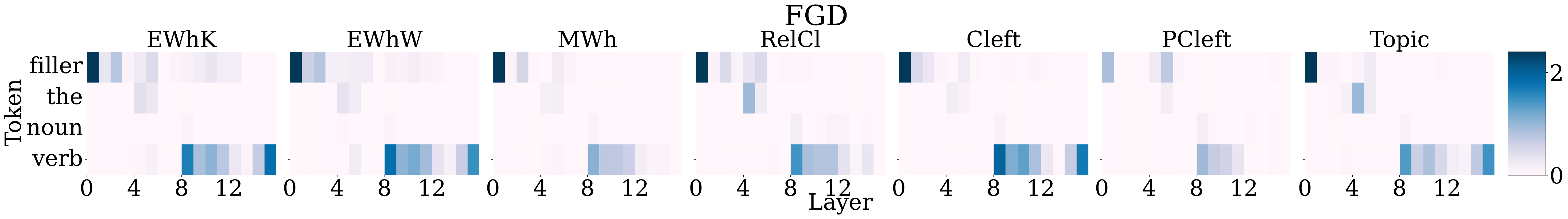}
    \caption{Training Steps: 3000}
  \end{subfigure}
  \begin{subfigure}{\linewidth}
    \centering
    \includegraphics[width=\linewidth]{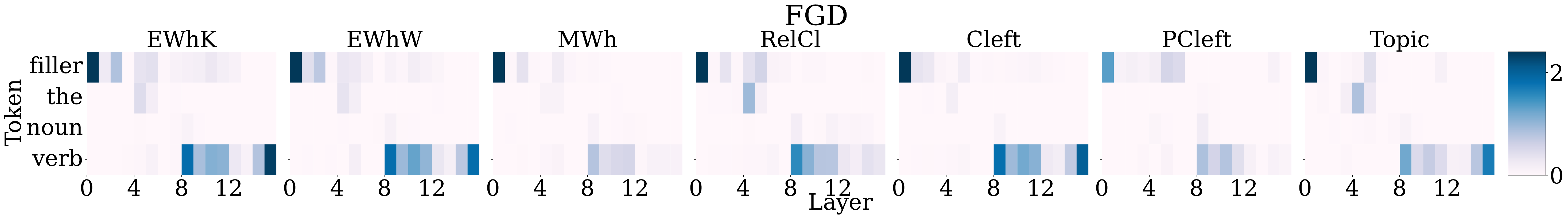}
    \caption{Training Steps: 4000}
  \end{subfigure}
  \begin{subfigure}{\linewidth}
    \centering
    \includegraphics[width=\linewidth]{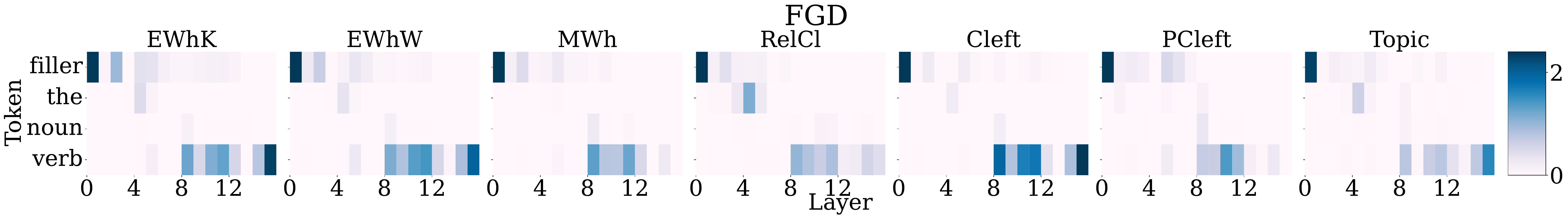}
    \caption{Training Steps: 10000}
  \end{subfigure}
  \begin{subfigure}{\linewidth}
    \centering
    \includegraphics[width=\linewidth]{sections/figures/results/odds_ratio_pythia_1b_143000_mlp_vanilla_repr_fg.pdf}
    \caption{Training Steps: 143000 (Final)}
  \end{subfigure}

  \caption{\textsc{Odds} scores with activation patching of MLP output of models with various training steps in filler-gap dependencies.}
  \label{fig:act_patch_fg_mlp_dynamics}
\end{figure*}

\begin{figure*}
\centering

  \begin{subfigure}{\linewidth}
    \centering
    \includegraphics[width=\linewidth]{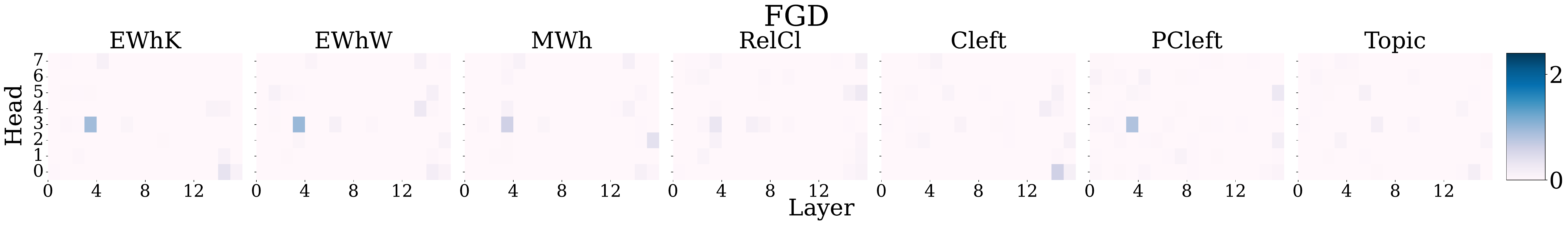}
    \caption{Training Steps: 1000}
  \end{subfigure}
  \begin{subfigure}{\linewidth}
    \centering
    \includegraphics[width=\linewidth]{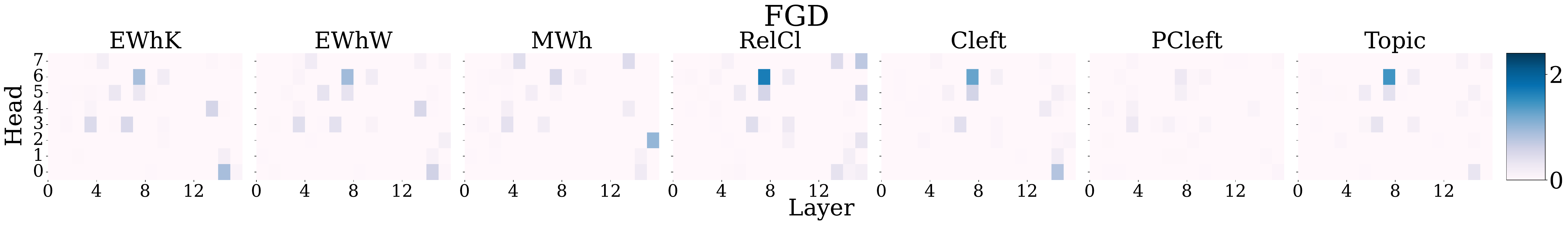}
    \caption{Training Steps: 2000}
  \end{subfigure}
  \begin{subfigure}{\linewidth}
    \centering
    \includegraphics[width=\linewidth]{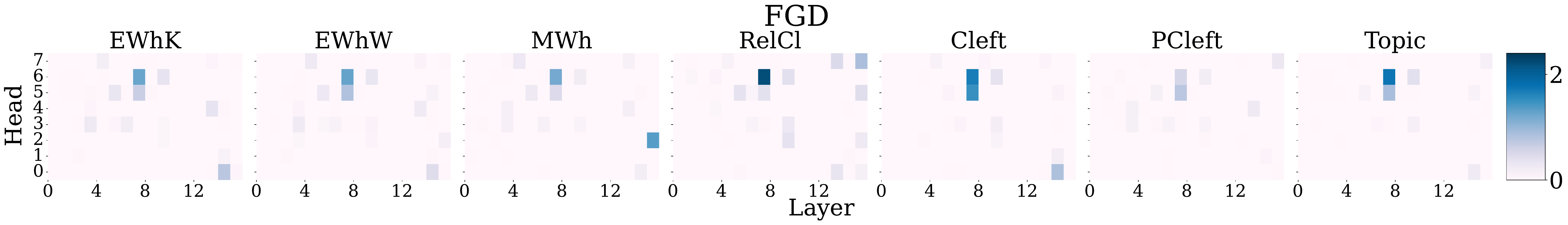}
    \caption{Training Steps: 3000}
  \end{subfigure}
  \begin{subfigure}{\linewidth}
    \centering
    \includegraphics[width=\linewidth]{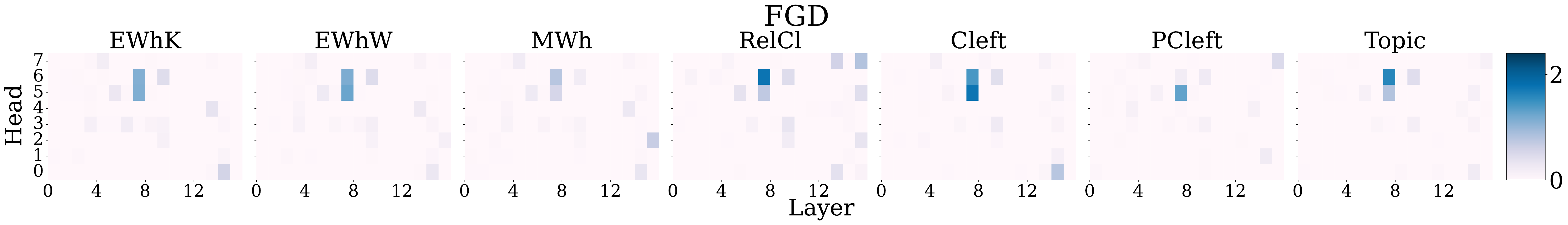}
    \caption{Training Steps: 4000}
  \end{subfigure}
  \begin{subfigure}{\linewidth}
    \centering
    \includegraphics[width=\linewidth]{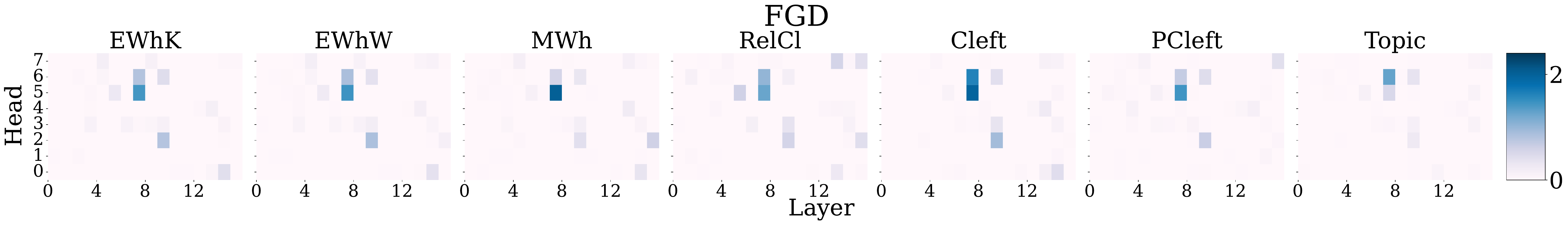}
    \caption{Training Steps: 10000}
  \end{subfigure}
  \begin{subfigure}{\linewidth}
    \centering
    \includegraphics[width=\linewidth]{sections/figures/results/odds_ratio_pythia_1b_143000_attn-head_vanilla_repr_fg.pdf}
    \caption{Training Steps: 143000 (Final)}
  \end{subfigure}

  \caption{\textsc{Odds} scores with activation patching of attention output of models with various training steps in filler-gap dependencies.}
  \label{fig:act_patch_fg_attn-head_dynamics}
\end{figure*}

We present the results of activation patching applied to each component of Pythia 1B across several training steps.
Figure~\ref{fig:act_patch_fg_resid_dynamics} through \ref{fig:act_patch_fg_attn-head_dynamics} show the results of residual stream, attention output, MLP output, and attention heads.

\subsection{Gemma 3}
\label{sec:gemma}
\begin{figure*}
\centering

  \begin{subfigure}{\linewidth}
    \centering
    \includegraphics[width=\linewidth]{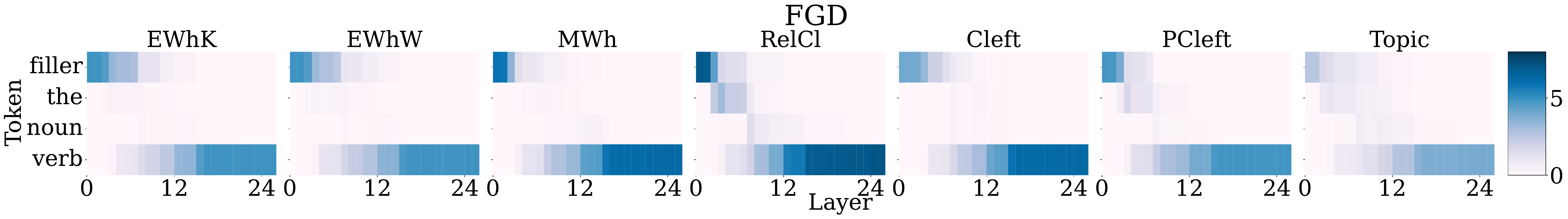}
    \caption{Residual stream}
  \end{subfigure}
  \begin{subfigure}{\linewidth}
    \centering
    \includegraphics[width=\linewidth]{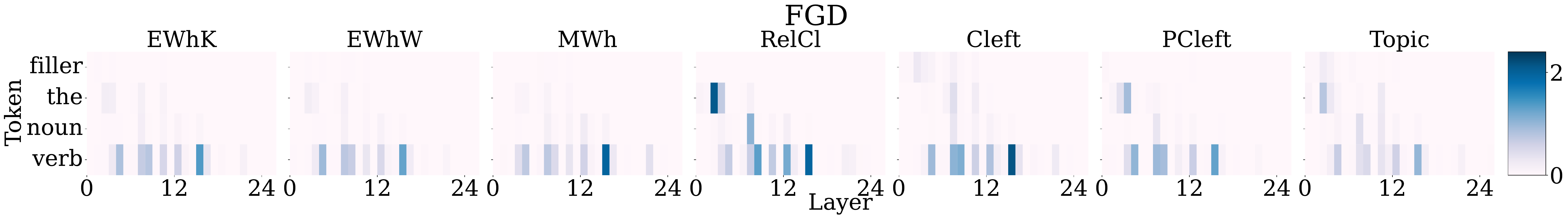}
    \caption{Attention}
  \end{subfigure}
  \begin{subfigure}{\linewidth}
    \centering
    \includegraphics[width=\linewidth]{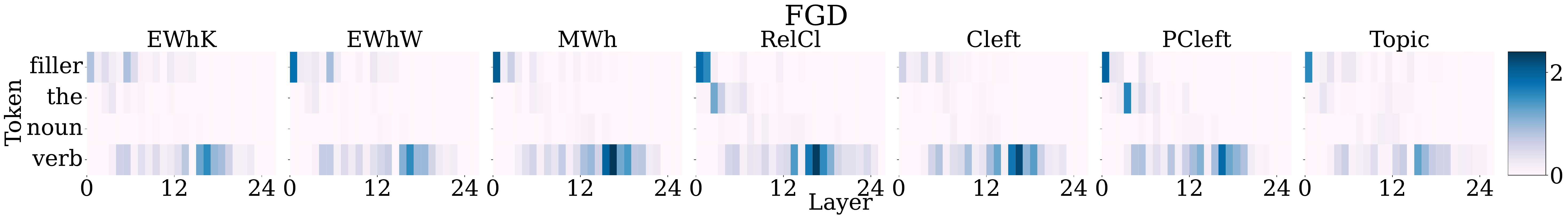}
    \caption{MLP}
  \end{subfigure}
  \begin{subfigure}{\linewidth}
    \centering
    \includegraphics[width=\linewidth]{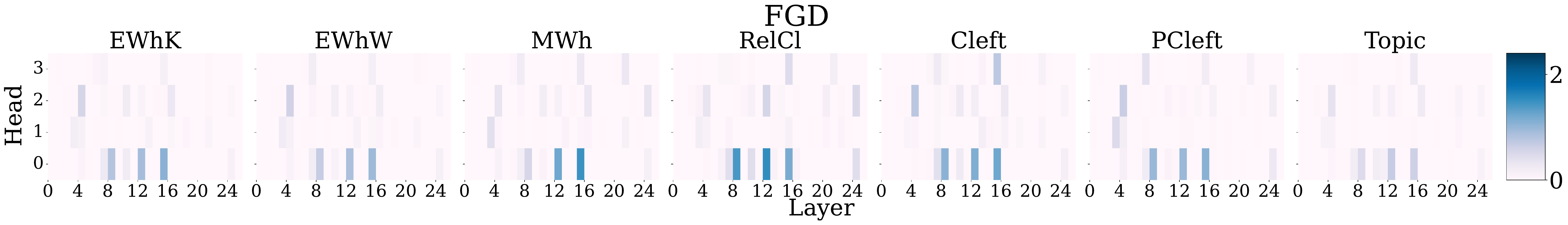}
    \caption{Attention head}
  \end{subfigure}

  \caption{\textsc{Odds} scores with activation patching in Gemma 3 1B in FGDs.}
  \label{fig:act_patch_fg_gemma3_1b}
\end{figure*}

\begin{figure*}
\centering

  \begin{subfigure}{\linewidth}
    \centering
    \includegraphics[width=\linewidth]{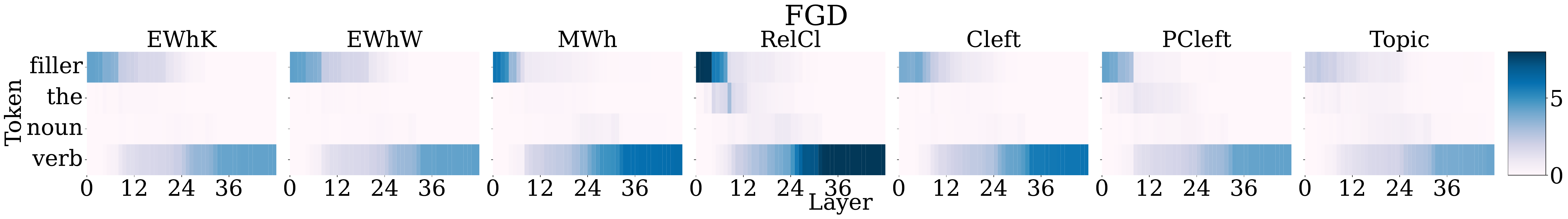}
    \caption{Residual stream}
  \end{subfigure}
  \begin{subfigure}{\linewidth}
    \centering
    \includegraphics[width=\linewidth]{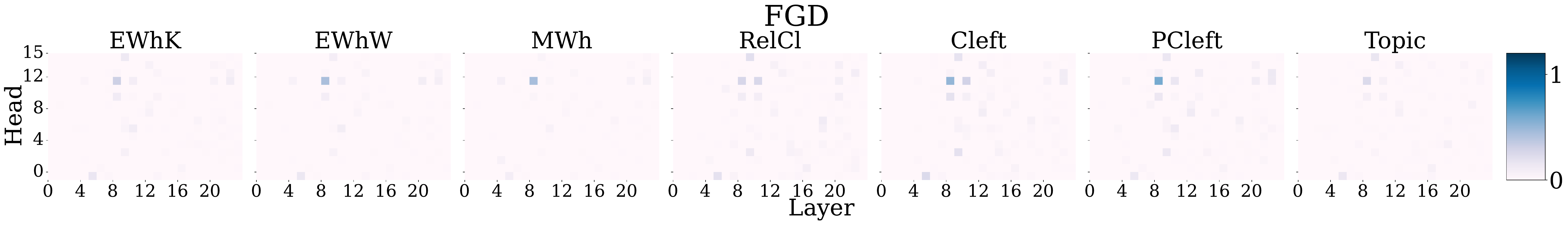}
    \caption{Attention head}
  \end{subfigure}

  \caption{\textsc{Odds} scores with activation patching in Gemma 3 12B in FGDs.
  The results of attention head are limited to layers 0-24 for clarity.}
  \label{fig:act_patch_fg_gemma3_12b}
\end{figure*}

We apply activation patching to Gemma 3 1B and 12B.
The results are shown in Figure~\ref{fig:act_patch_fg_gemma3_1b} and \ref{fig:act_patch_fg_gemma3_12b}.
Similar to Pythia models, a shared mechanism can be seen among FGD constructions in Gemma 3, although the scores were higher in \textsc{RelCl} than in other constructions.

\subsection{Comparison of Activation Patching and DAS in All Constructions in FGDs}
\label{sec:full_das}
\begin{figure*}
\centering

  \begin{subfigure}{0.315\linewidth}
    \centering
    \includegraphics[width=\linewidth]{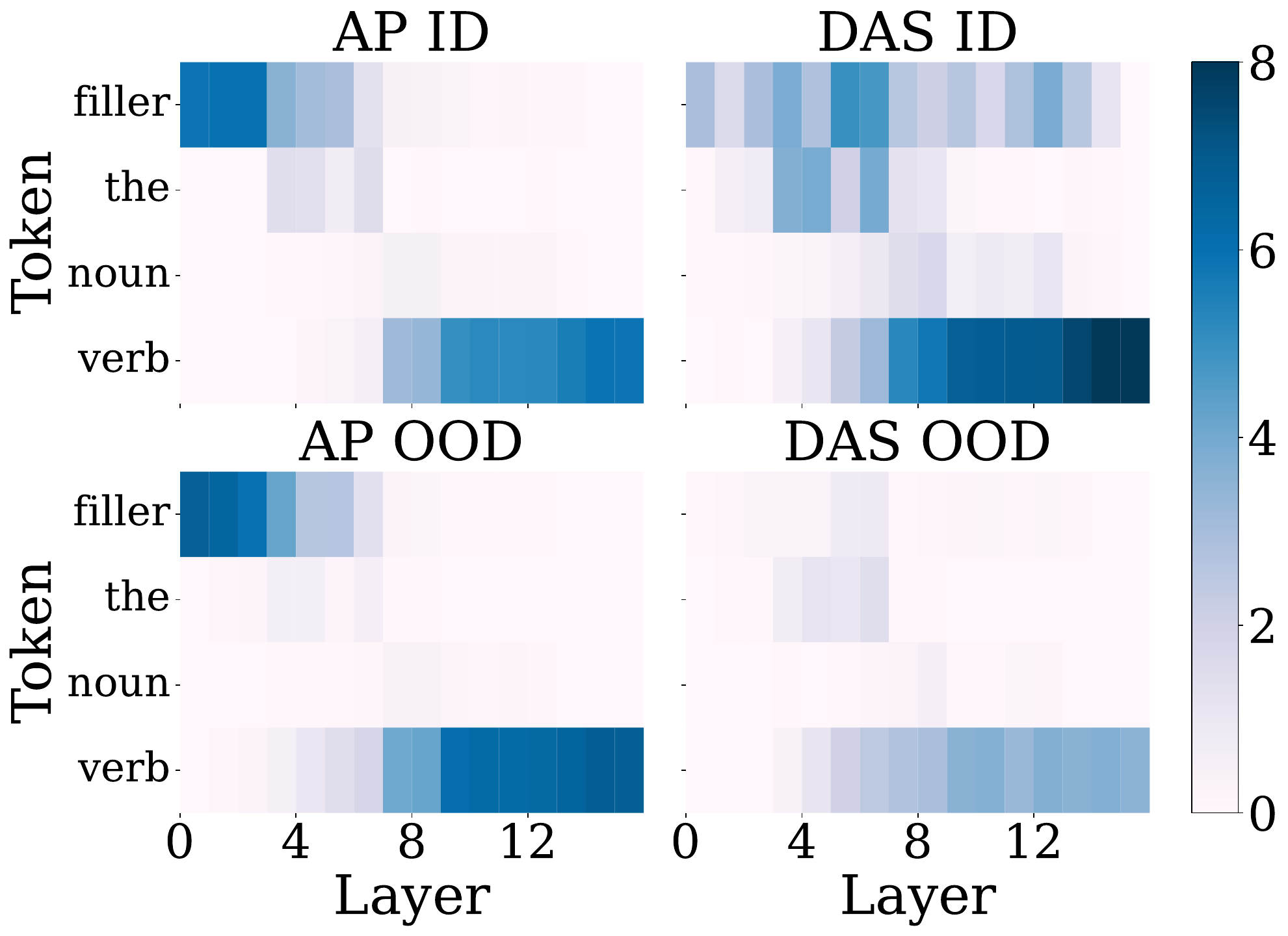}
    \caption{\textsc{EWhK}}
  \end{subfigure}
  \begin{subfigure}{0.315\linewidth}
    \centering
    \includegraphics[width=\linewidth]{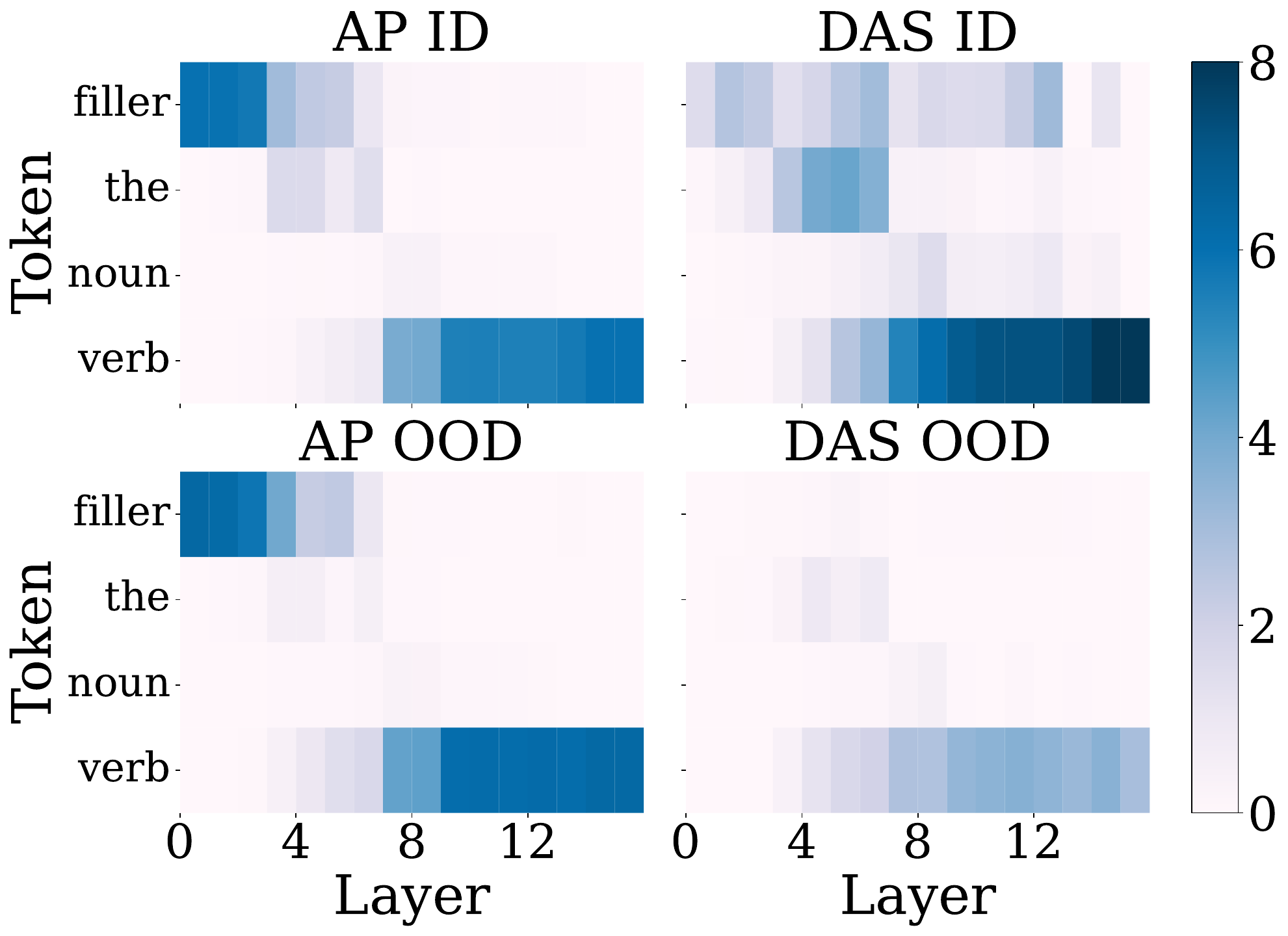}
    \caption{\textsc{EWhW}}
  \end{subfigure}
  \begin{subfigure}{0.315\linewidth}
    \centering
    \includegraphics[width=\linewidth]{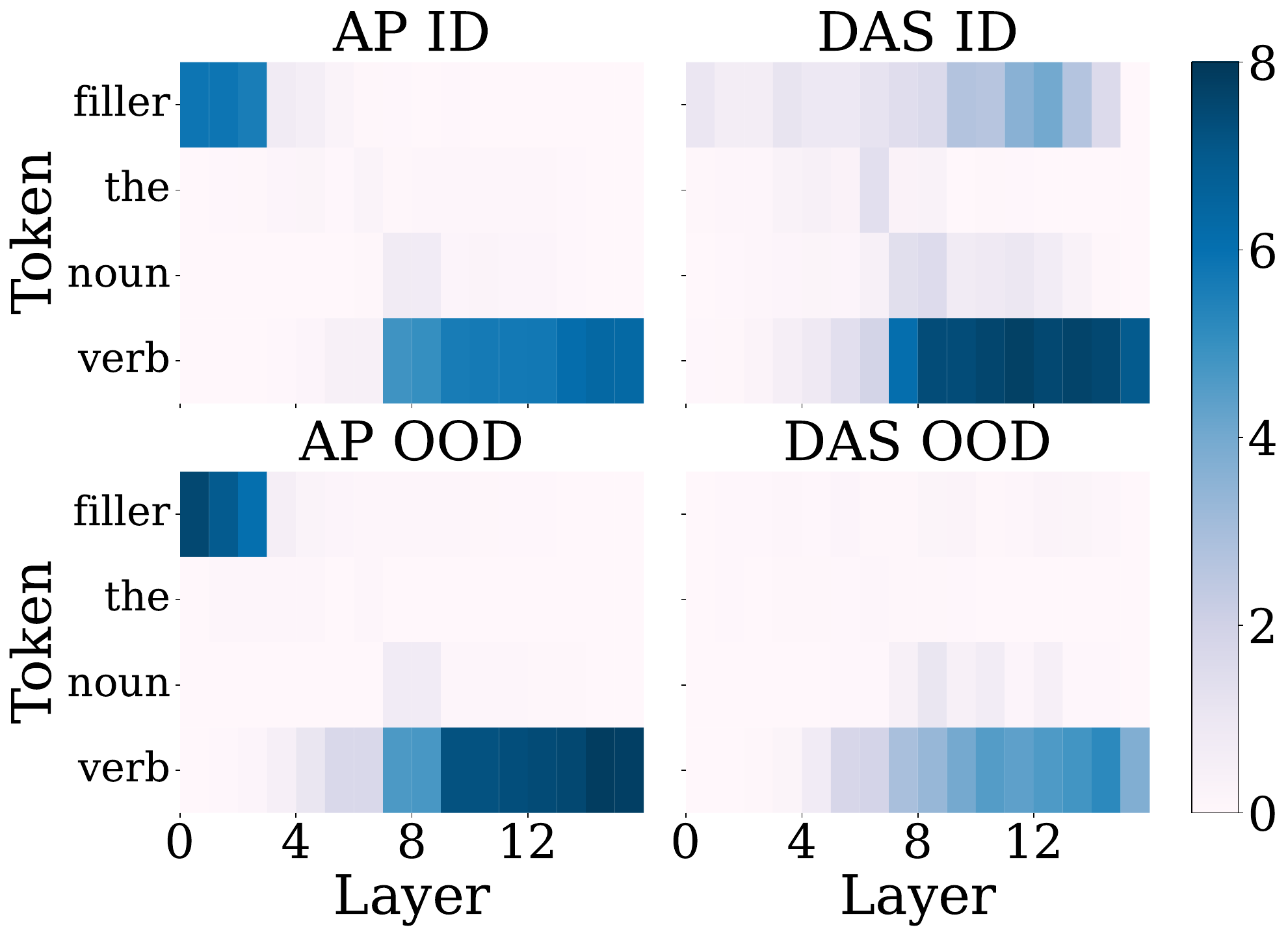}
    \caption{\textsc{MWh}}
  \end{subfigure}
  \begin{subfigure}{0.315\linewidth}
    \centering
    \includegraphics[width=\linewidth]{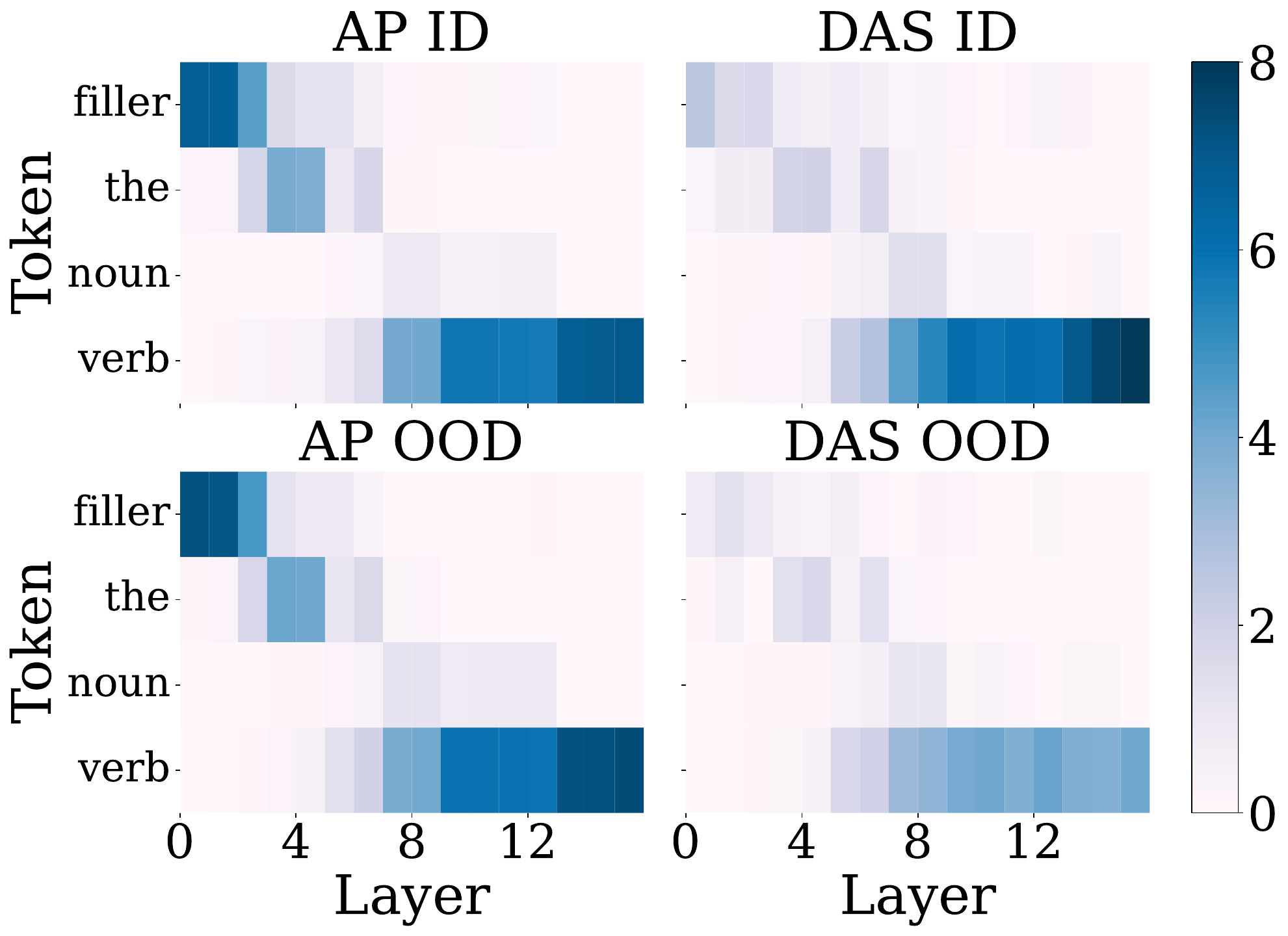}
    \caption{\textsc{RelCl}}
  \end{subfigure}
  \begin{subfigure}{0.315\linewidth}
    \centering
    \includegraphics[width=\linewidth]{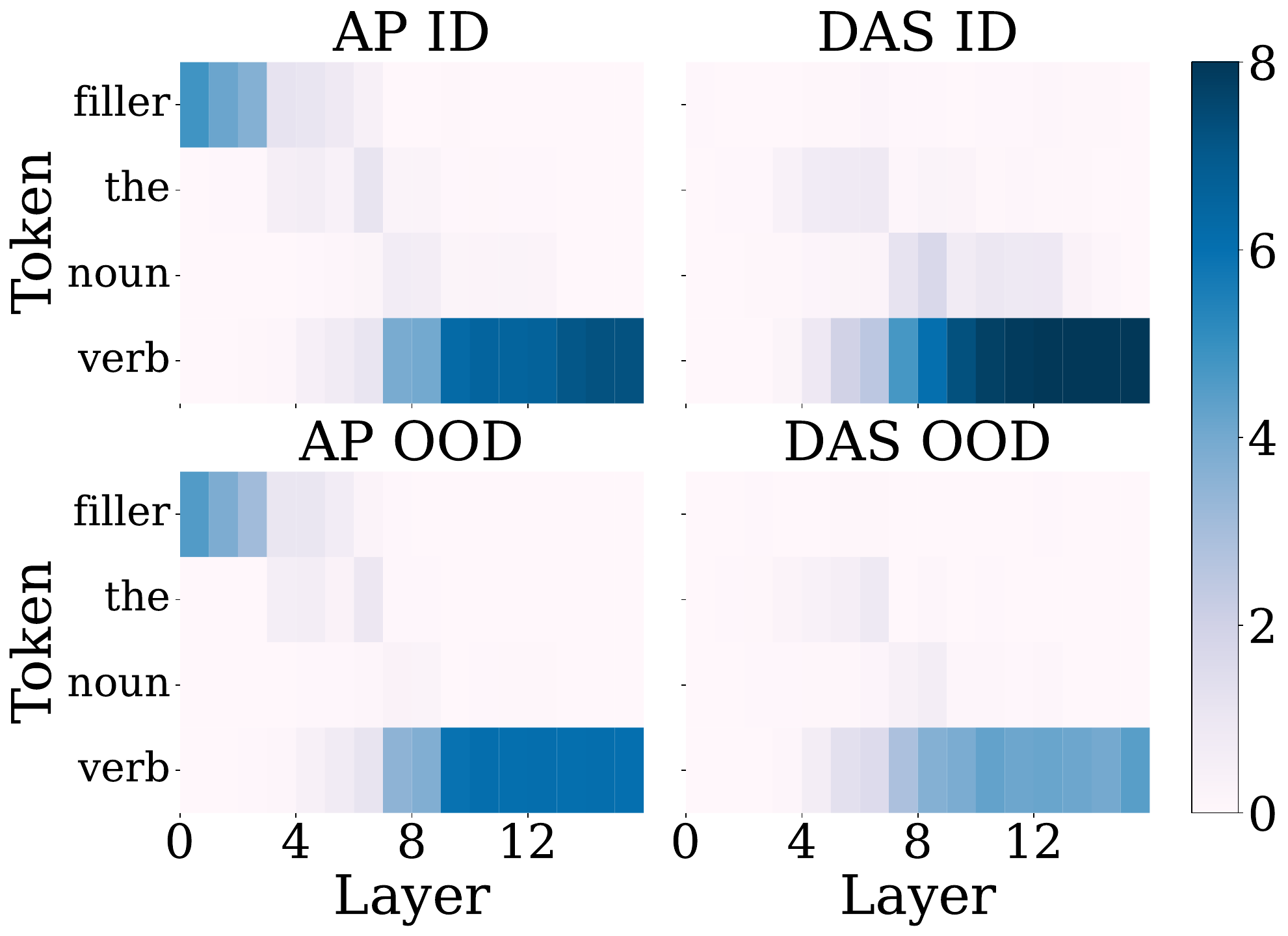}
    \caption{\textsc{Cleft}}
  \end{subfigure}
  \begin{subfigure}{0.315\linewidth}
    \centering
    \includegraphics[width=\linewidth]{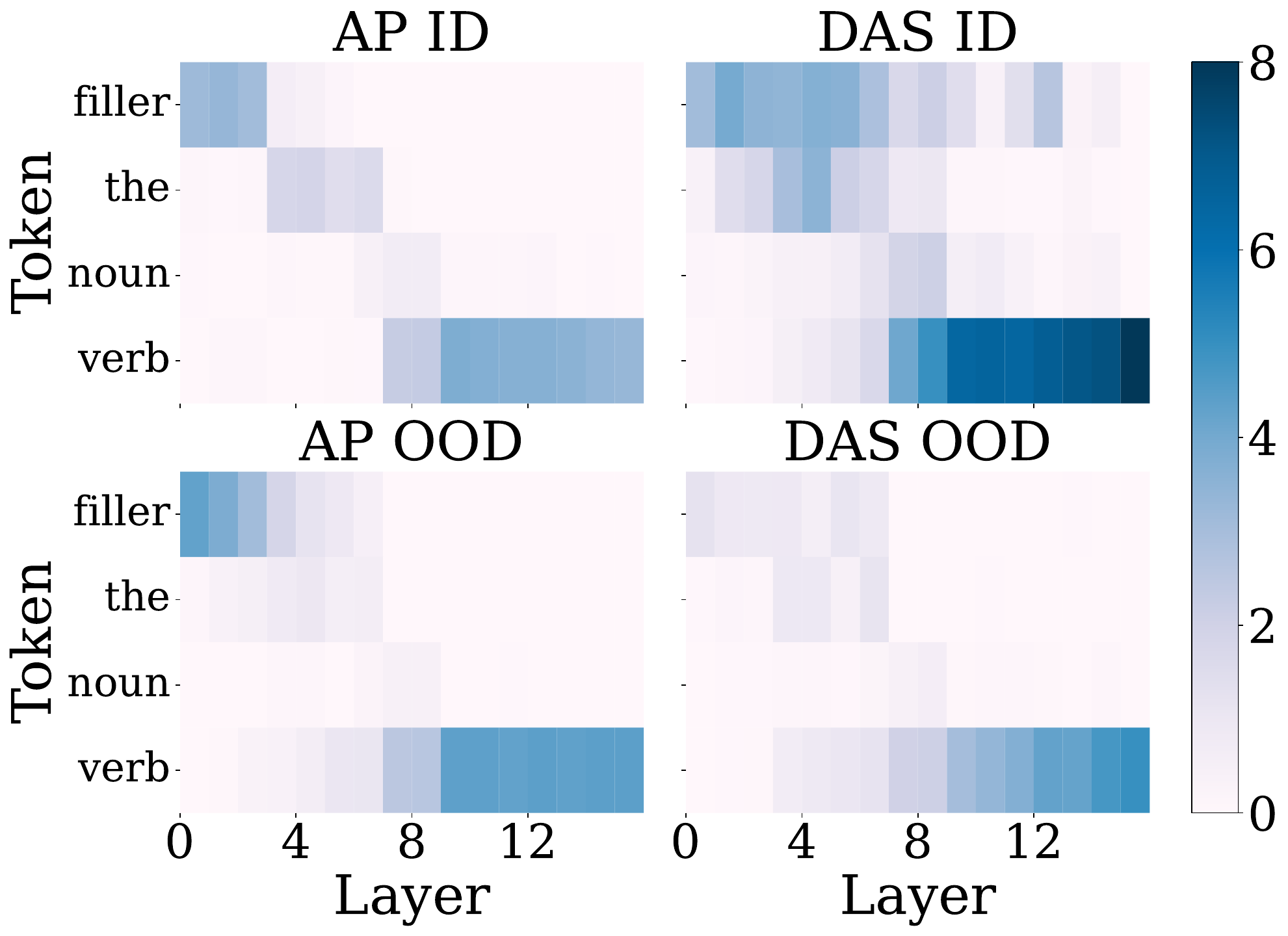}
    \caption{\textsc{PCleft}}
  \end{subfigure}
  \begin{subfigure}{0.315\linewidth}
    \centering
    \includegraphics[width=\linewidth]{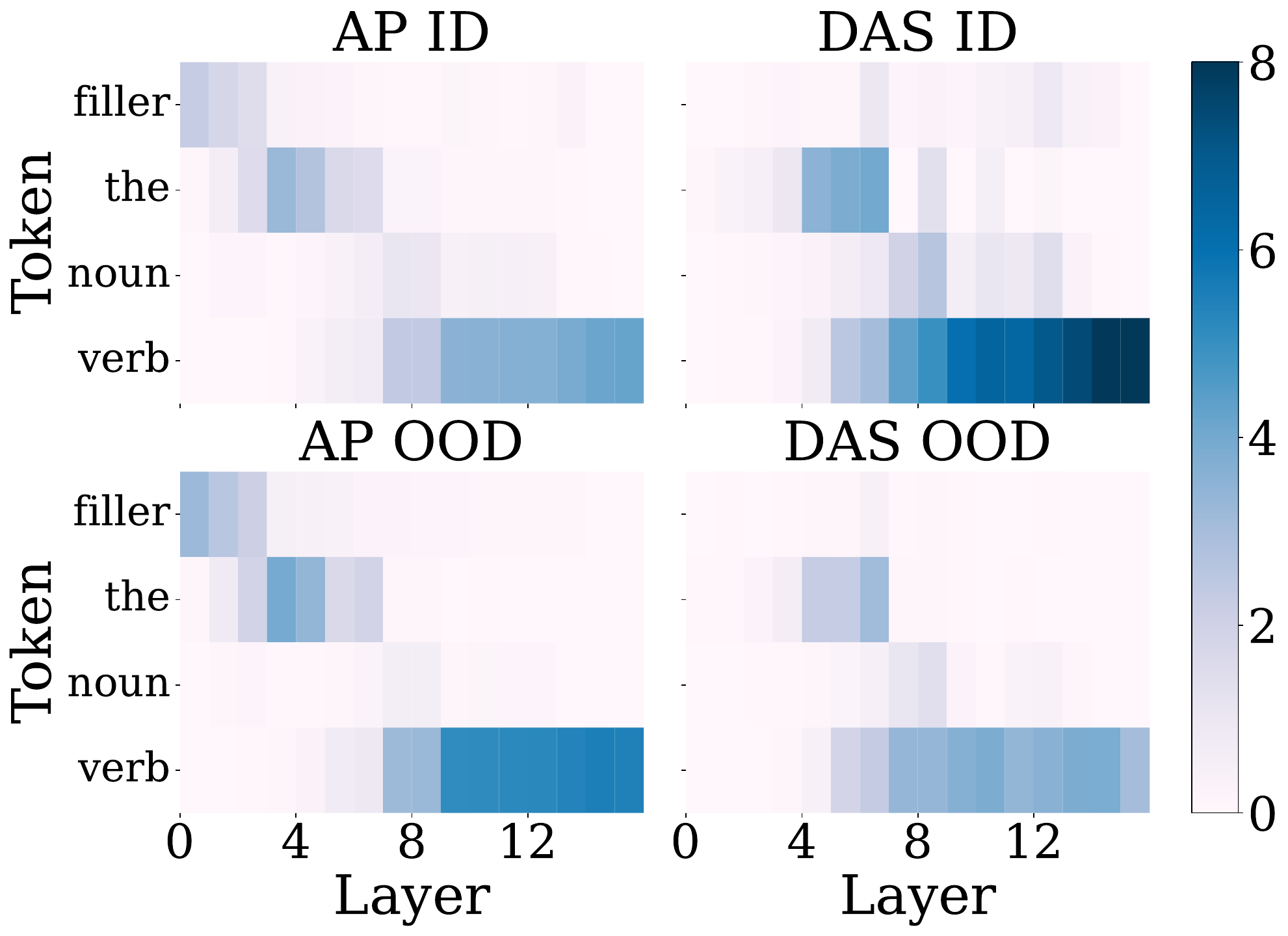}
    \caption{\textsc{Topic}}
  \end{subfigure}

  \caption{Comparison of \textsc{Odds} scores between activation patching (AP) and DAS of residual stream in Pythia 1B, evaluated on all the constructions in FGDs.}
  \label{fig:act_patch_fg_generalization_resid}
\end{figure*}

\begin{figure*}
\centering

  \begin{subfigure}{0.315\linewidth}
    \centering
    \includegraphics[width=\linewidth]{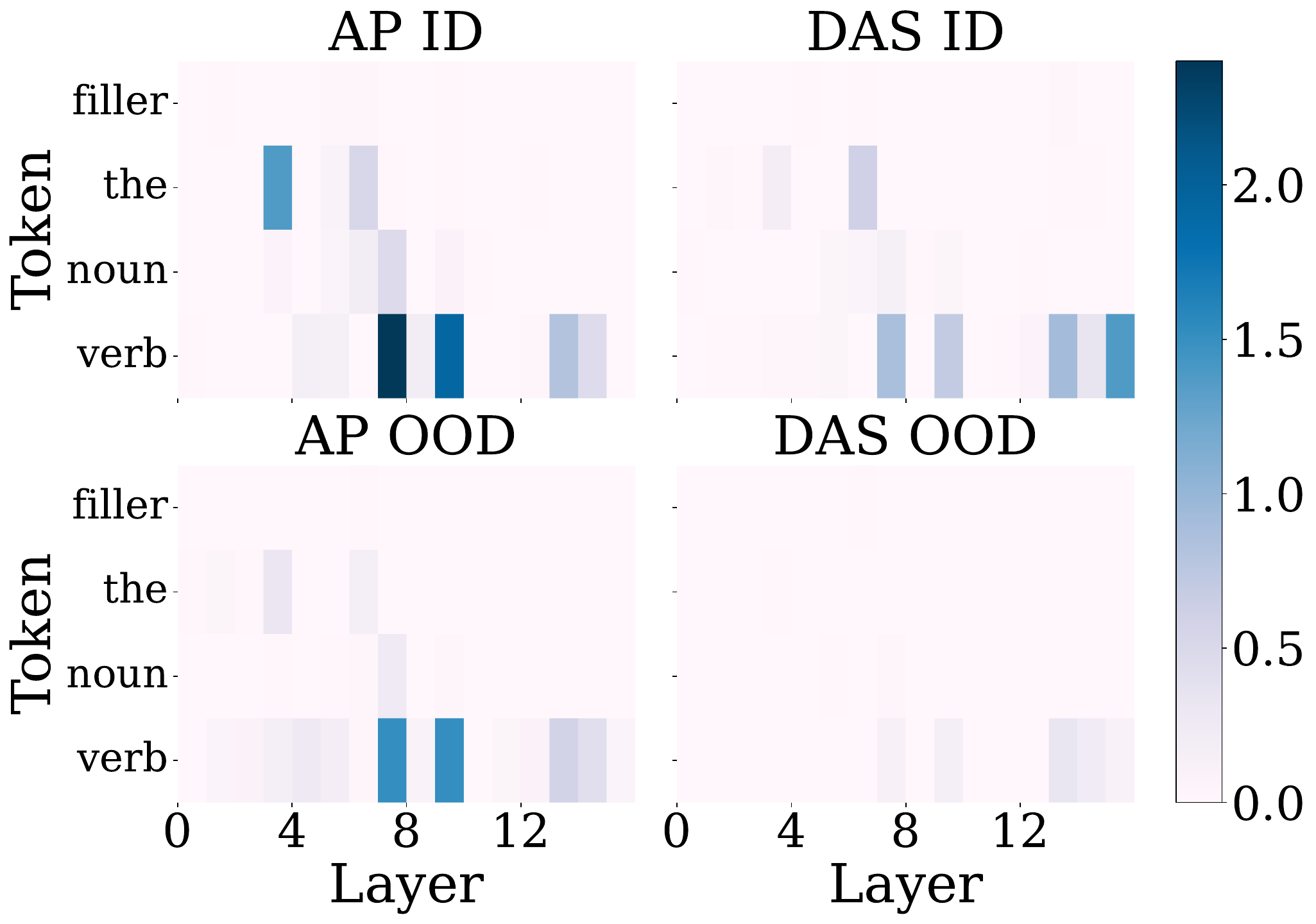}
    \caption{\textsc{EWhK}}
  \end{subfigure}
  \begin{subfigure}{0.315\linewidth}
    \centering
    \includegraphics[width=\linewidth]{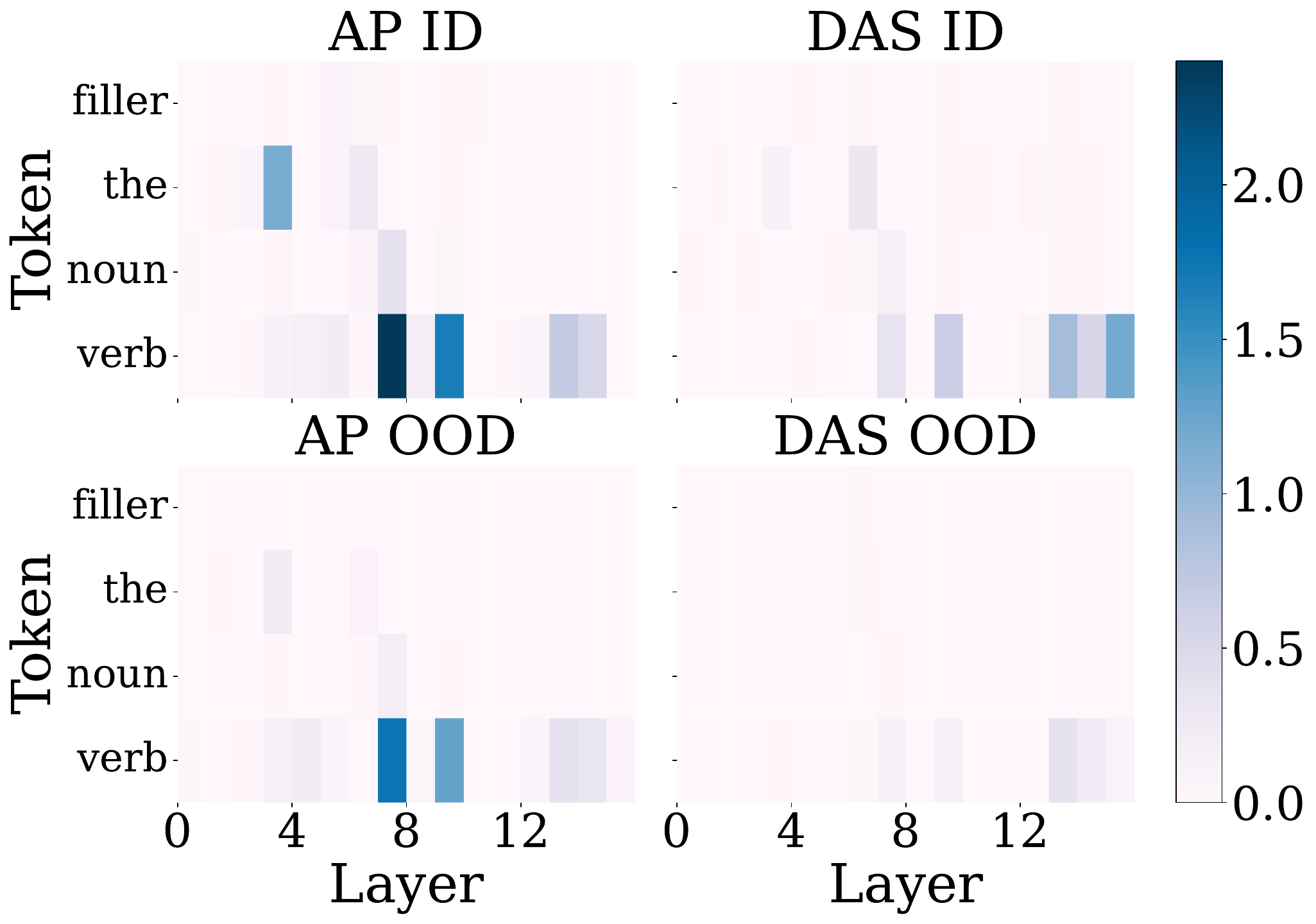}
    \caption{\textsc{EWhW}}
  \end{subfigure}
  \begin{subfigure}{0.315\linewidth}
    \centering
    \includegraphics[width=\linewidth]{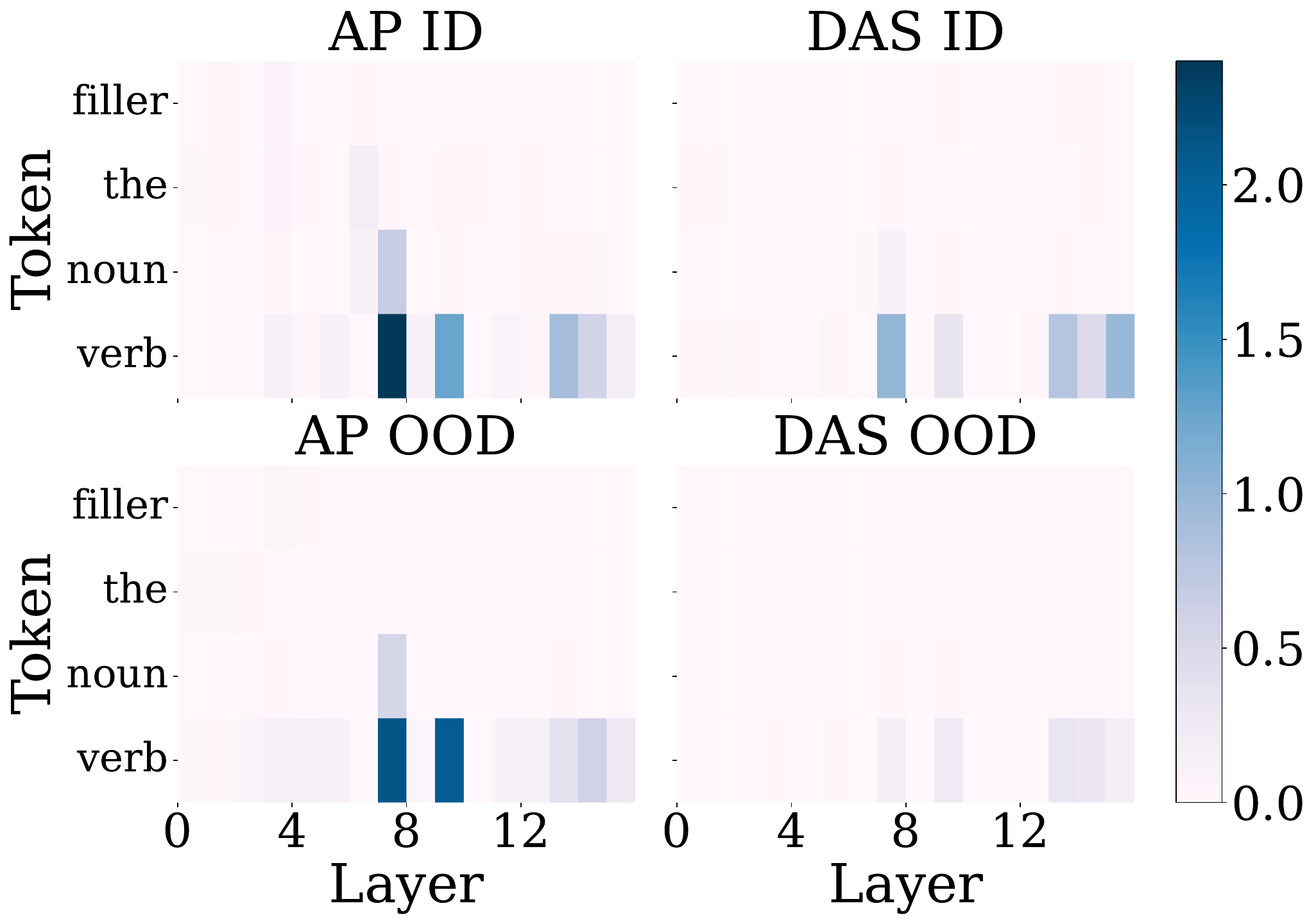}
    \caption{\textsc{MWh}}
  \end{subfigure}
  \begin{subfigure}{0.315\linewidth}
    \centering
    \includegraphics[width=\linewidth]{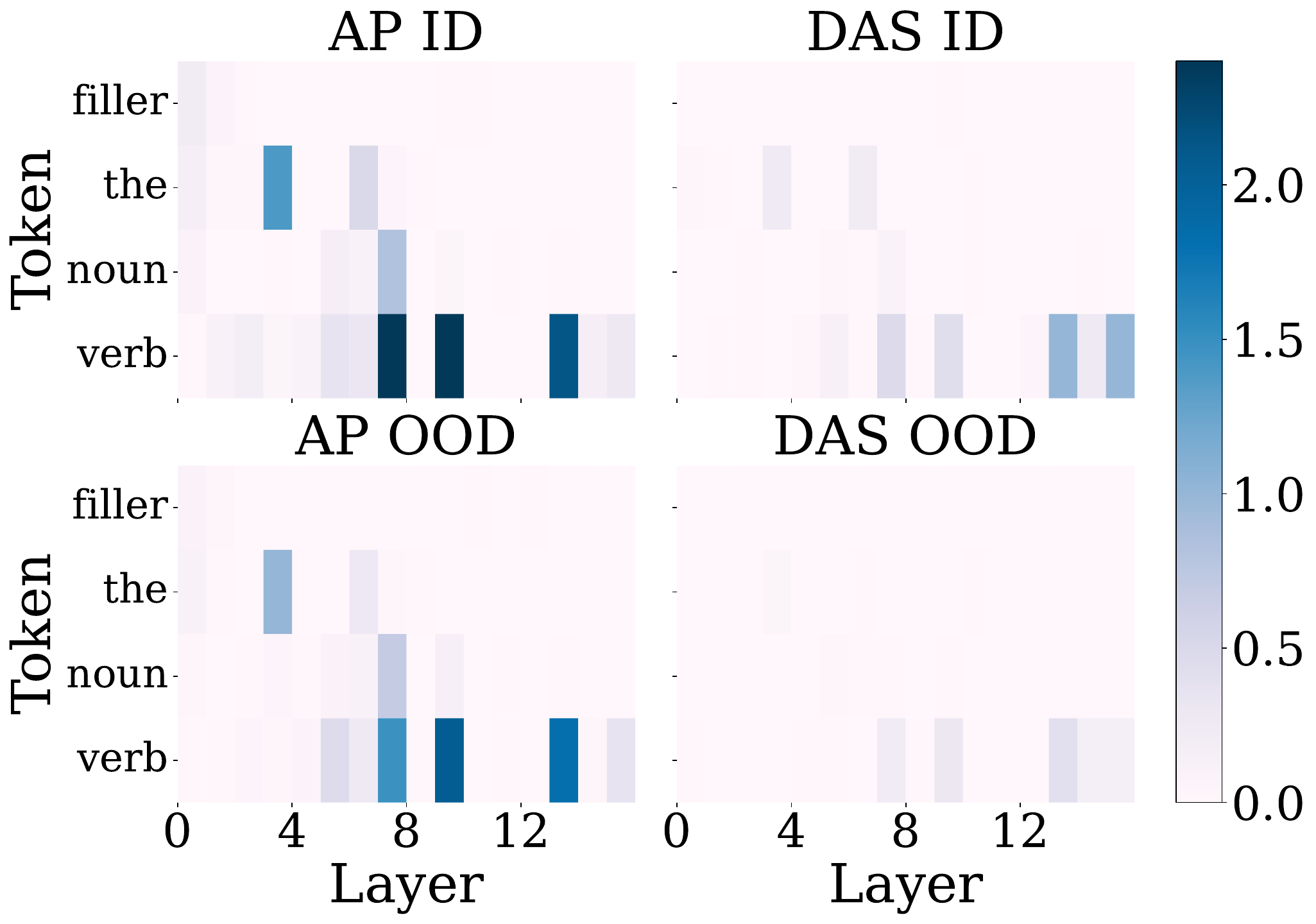}
    \caption{\textsc{RelCl}}
  \end{subfigure}
  \begin{subfigure}{0.315\linewidth}
    \centering
    \includegraphics[width=\linewidth]{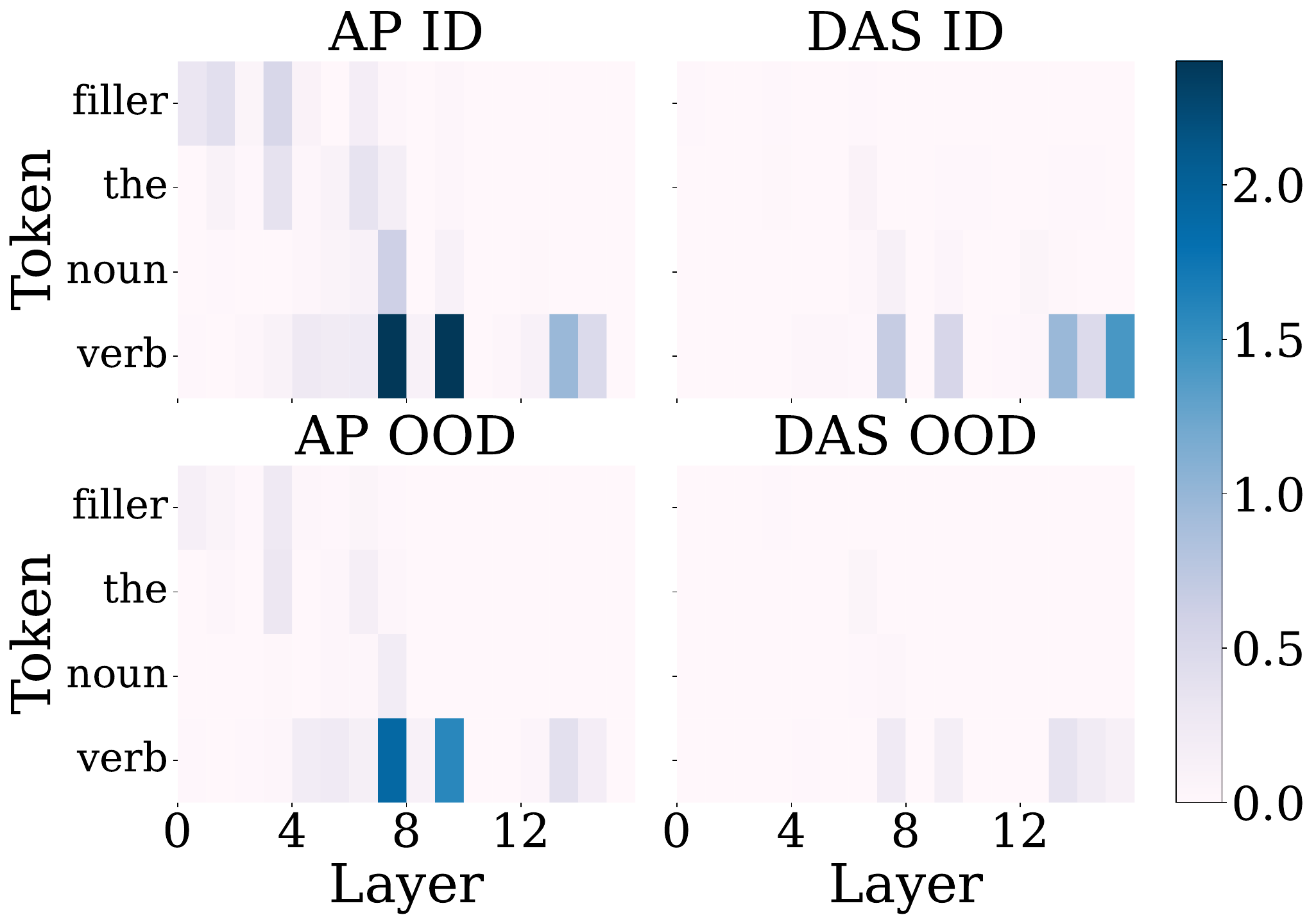}
    \caption{\textsc{Cleft}}
  \end{subfigure}
  \begin{subfigure}{0.315\linewidth}
    \centering
    \includegraphics[width=\linewidth]{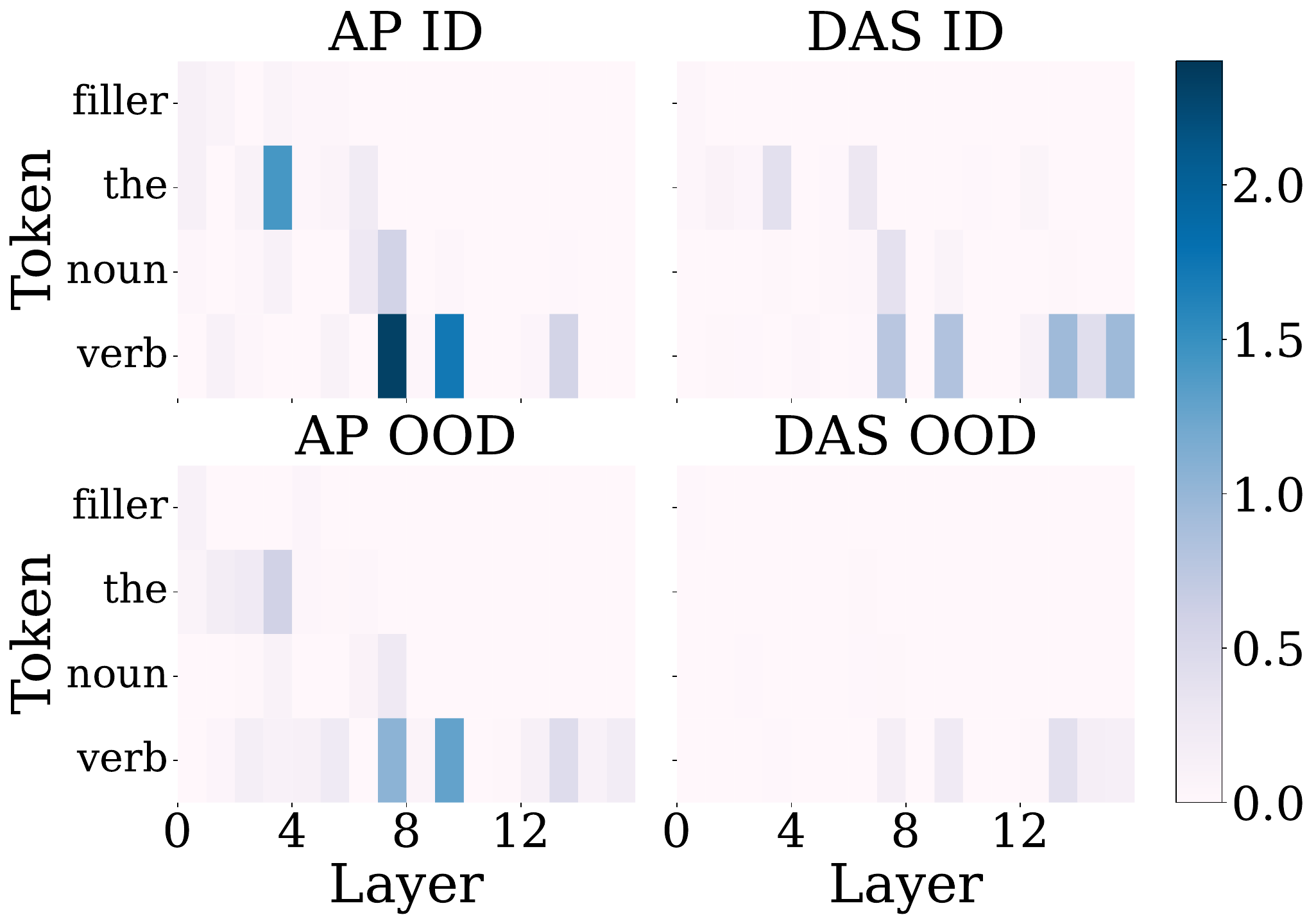}
    \caption{\textsc{PCleft}}
  \end{subfigure}
  \begin{subfigure}{0.315\linewidth}
    \centering
    \includegraphics[width=\linewidth]{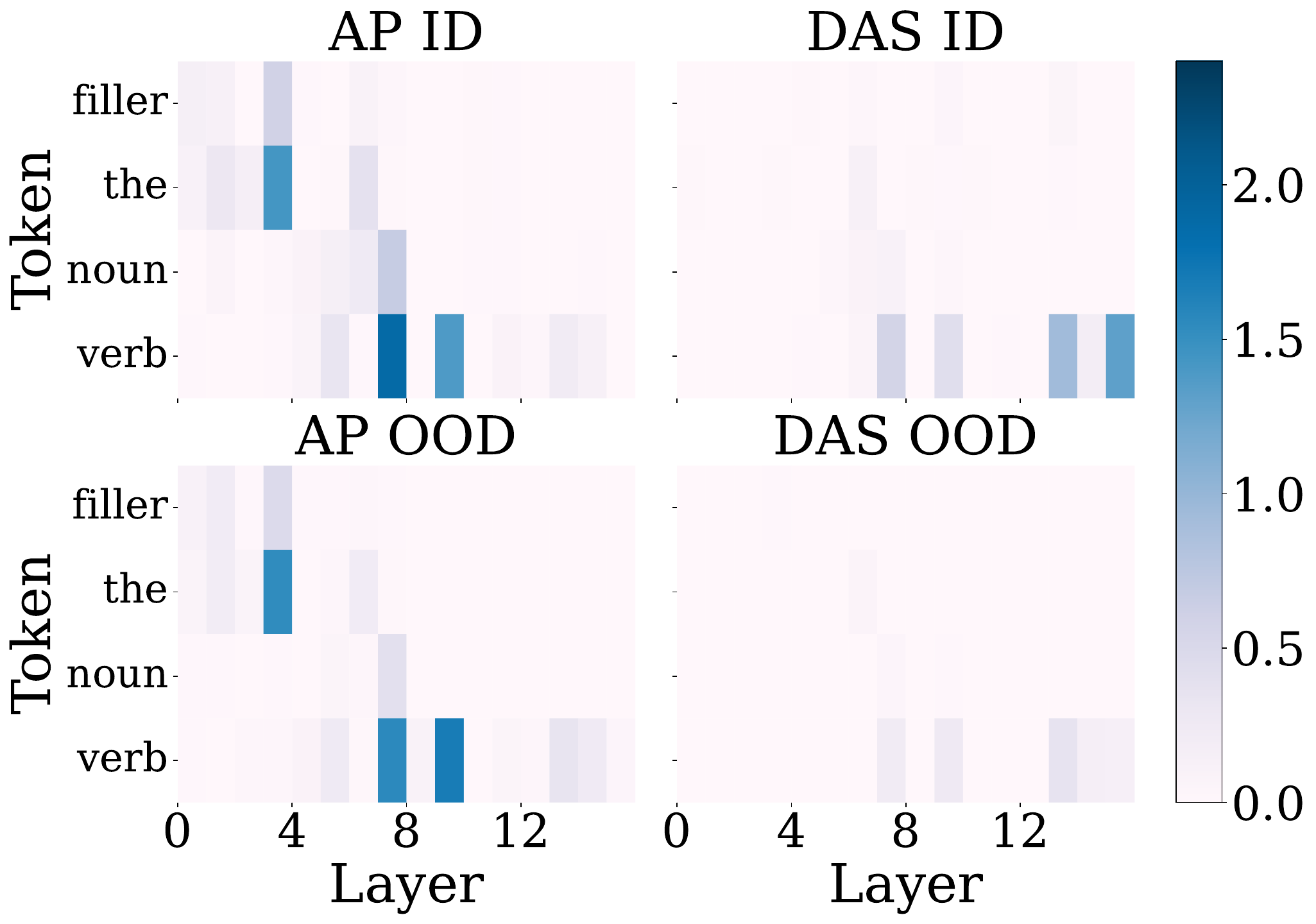}
    \caption{\textsc{Topic}}
  \end{subfigure}

  \caption{Comparison of \textsc{Odds} scores between activation patching (AP) and DAS of attention output in Pythia 1B, evaluated on all the constructions in FGDs.}
  \label{fig:act_patch_fg_generalization_attn}
\end{figure*}

\begin{figure*}
\centering

  \begin{subfigure}{0.315\linewidth}
    \centering
    \includegraphics[width=\linewidth]{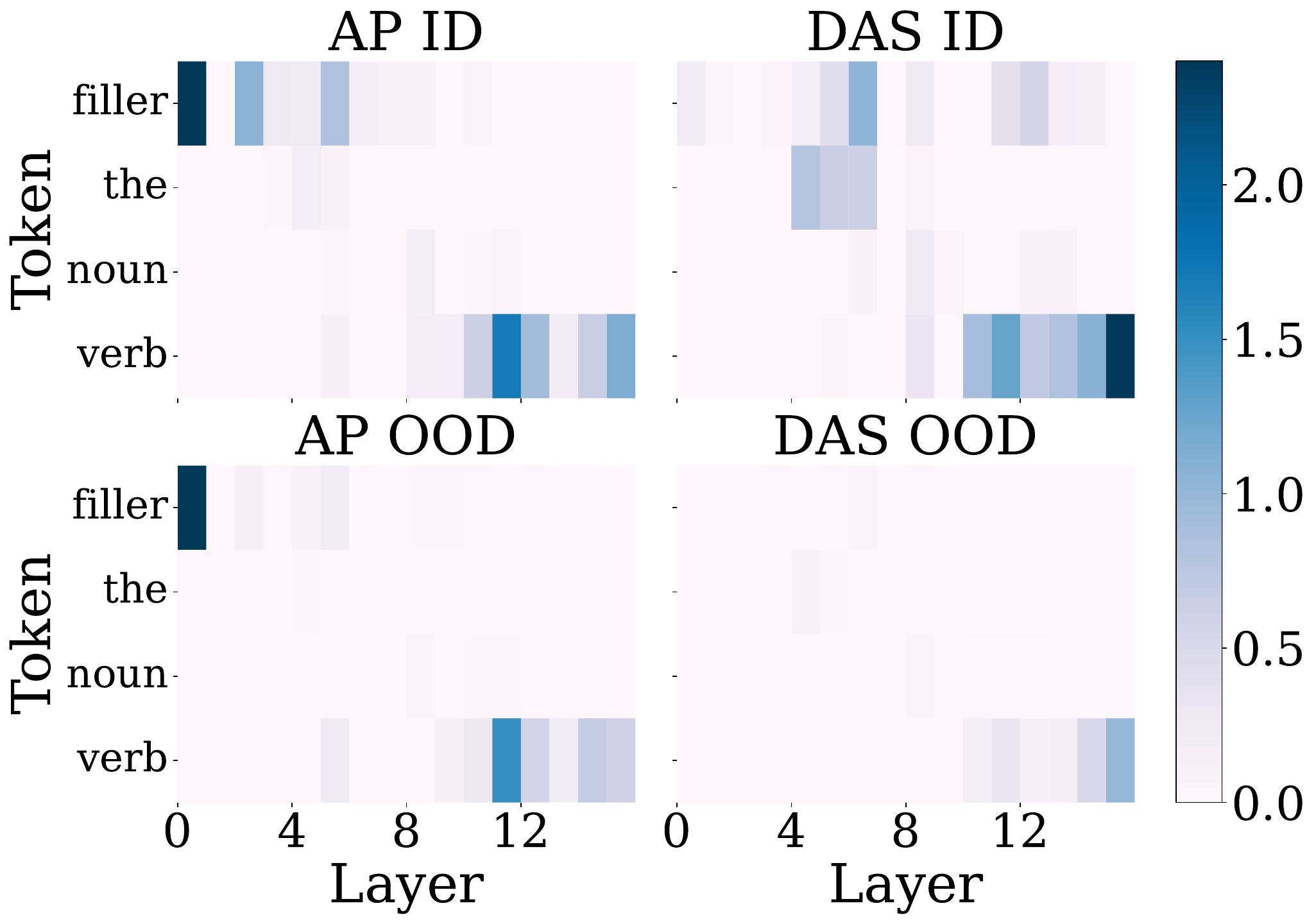}
    \caption{\textsc{EWhK}}
  \end{subfigure}
  \begin{subfigure}{0.315\linewidth}
    \centering
    \includegraphics[width=\linewidth]{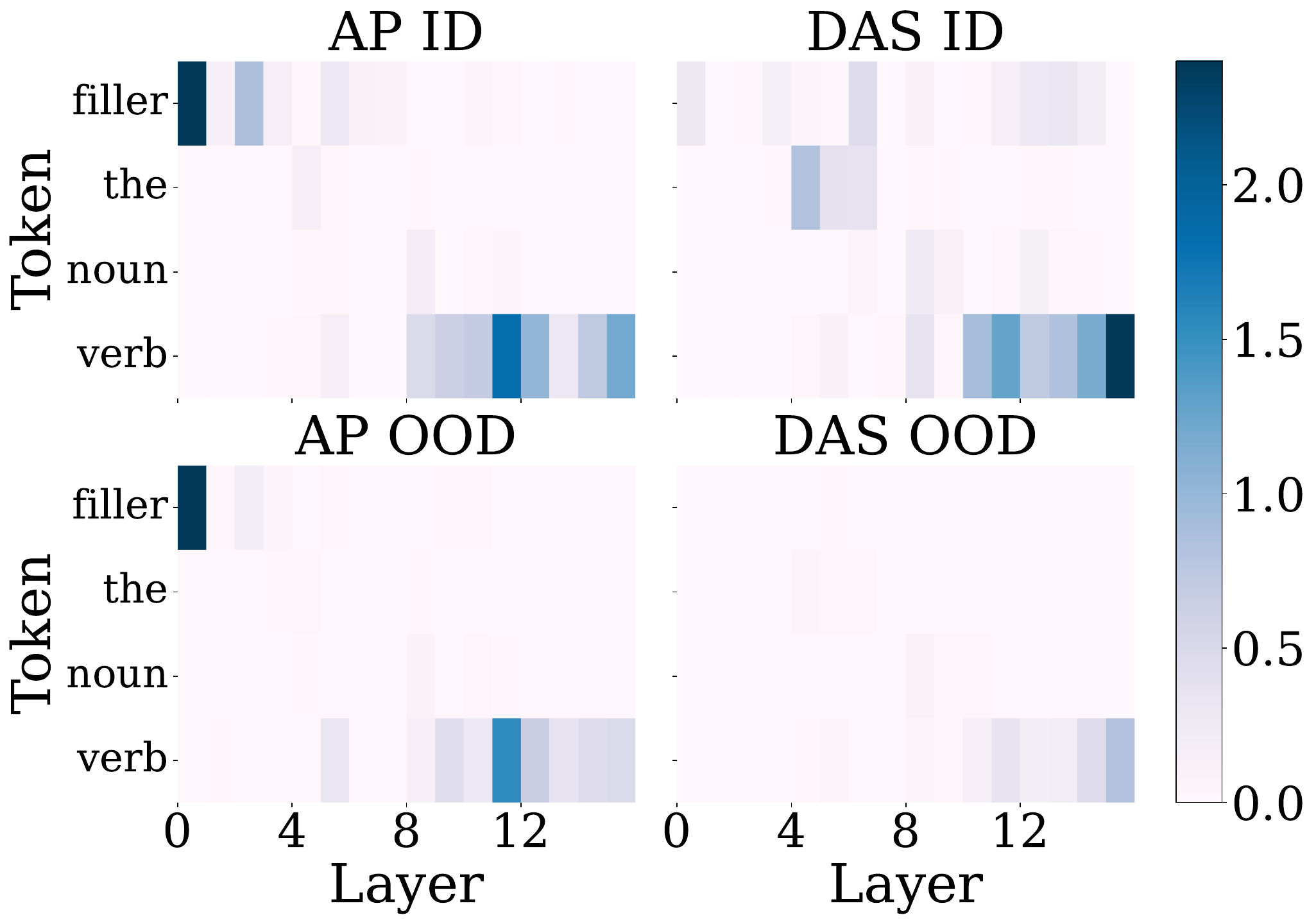}
    \caption{\textsc{EWhW}}
  \end{subfigure}
  \begin{subfigure}{0.315\linewidth}
    \centering
    \includegraphics[width=\linewidth]{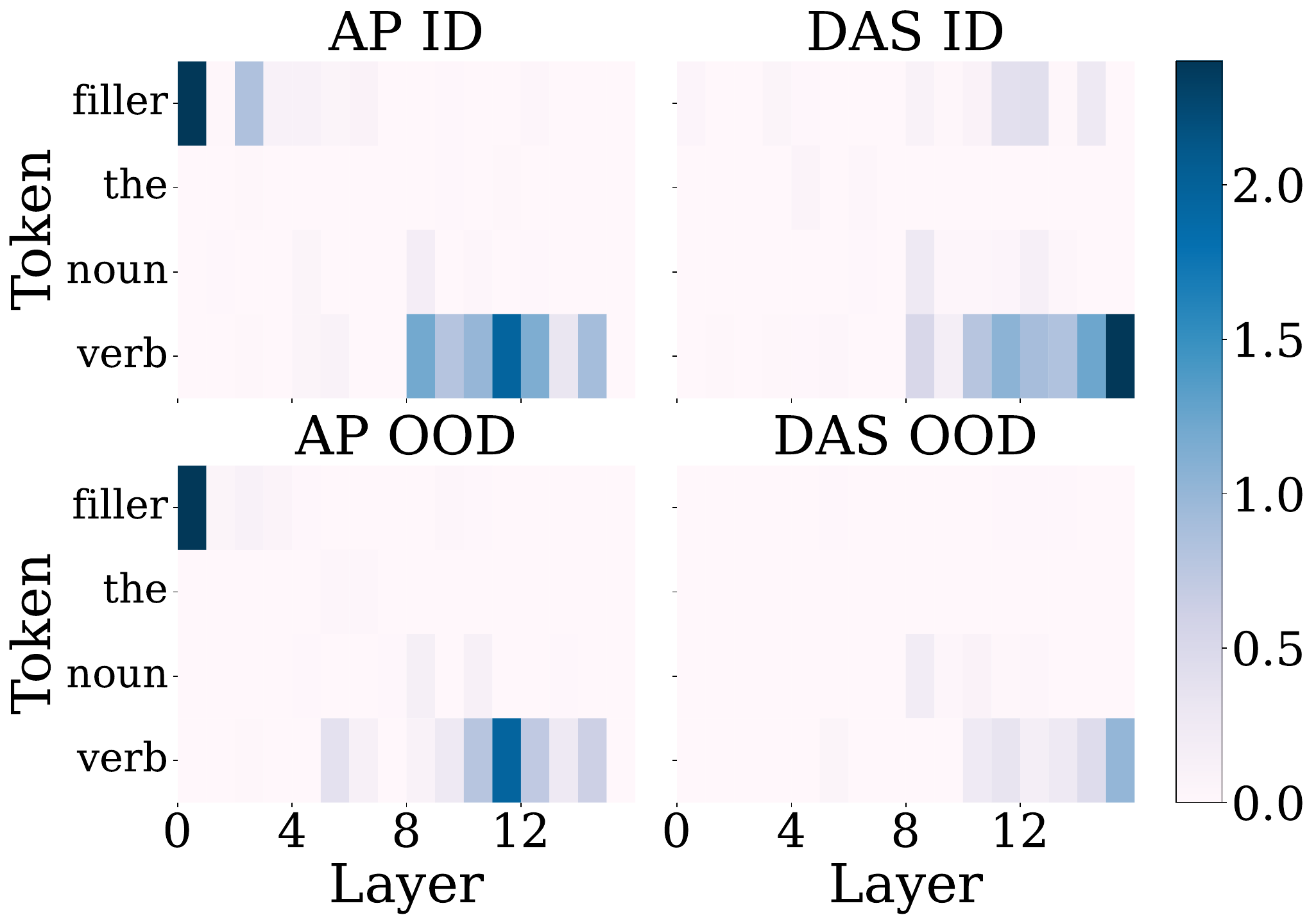}
    \caption{\textsc{MWh}}
  \end{subfigure}
  \begin{subfigure}{0.315\linewidth}
    \centering
    \includegraphics[width=\linewidth]{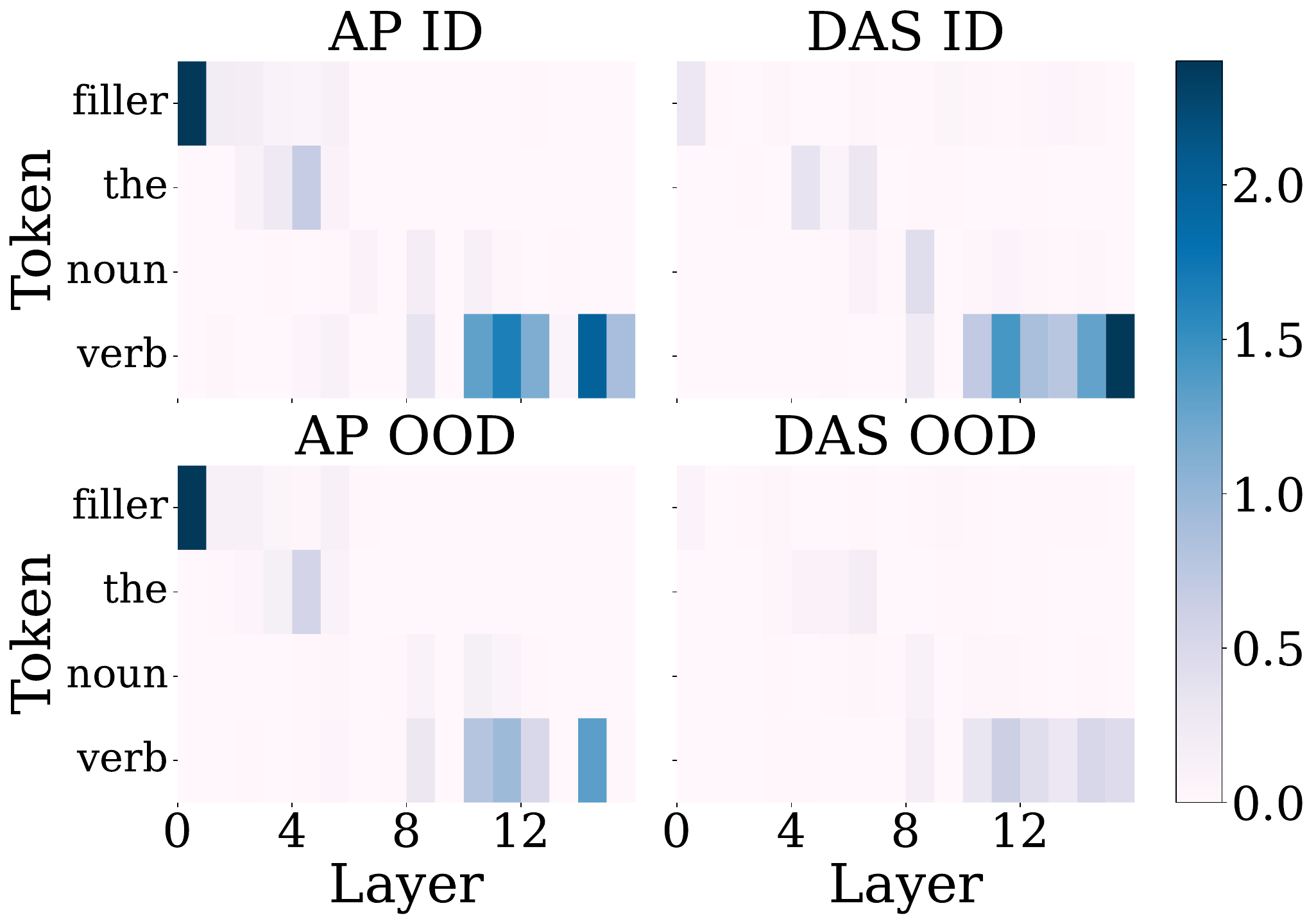}
    \caption{\textsc{RelCl}}
  \end{subfigure}
  \begin{subfigure}{0.315\linewidth}
    \centering
    \includegraphics[width=\linewidth]{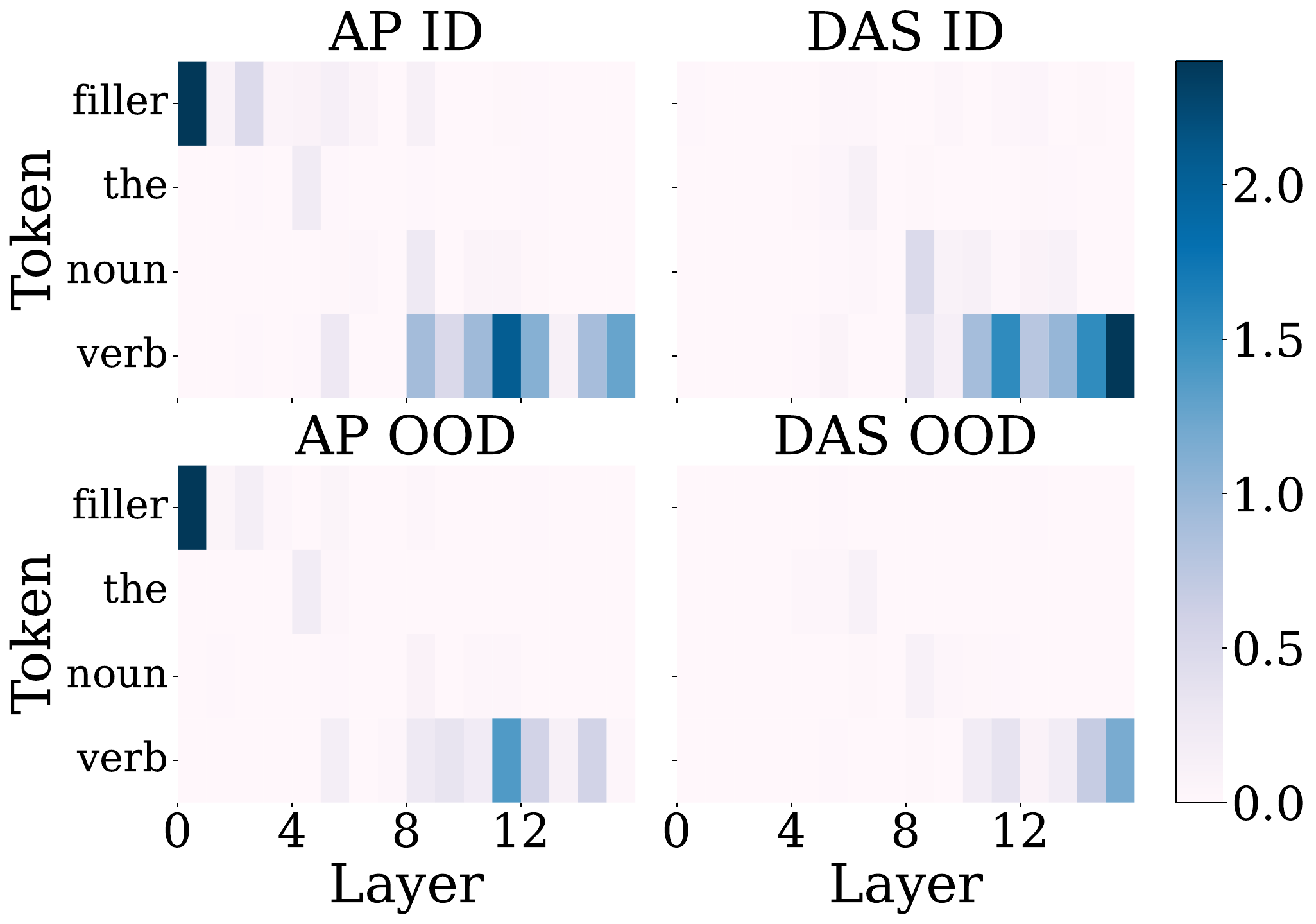}
    \caption{\textsc{Cleft}}
  \end{subfigure}
  \begin{subfigure}{0.315\linewidth}
    \centering
    \includegraphics[width=\linewidth]{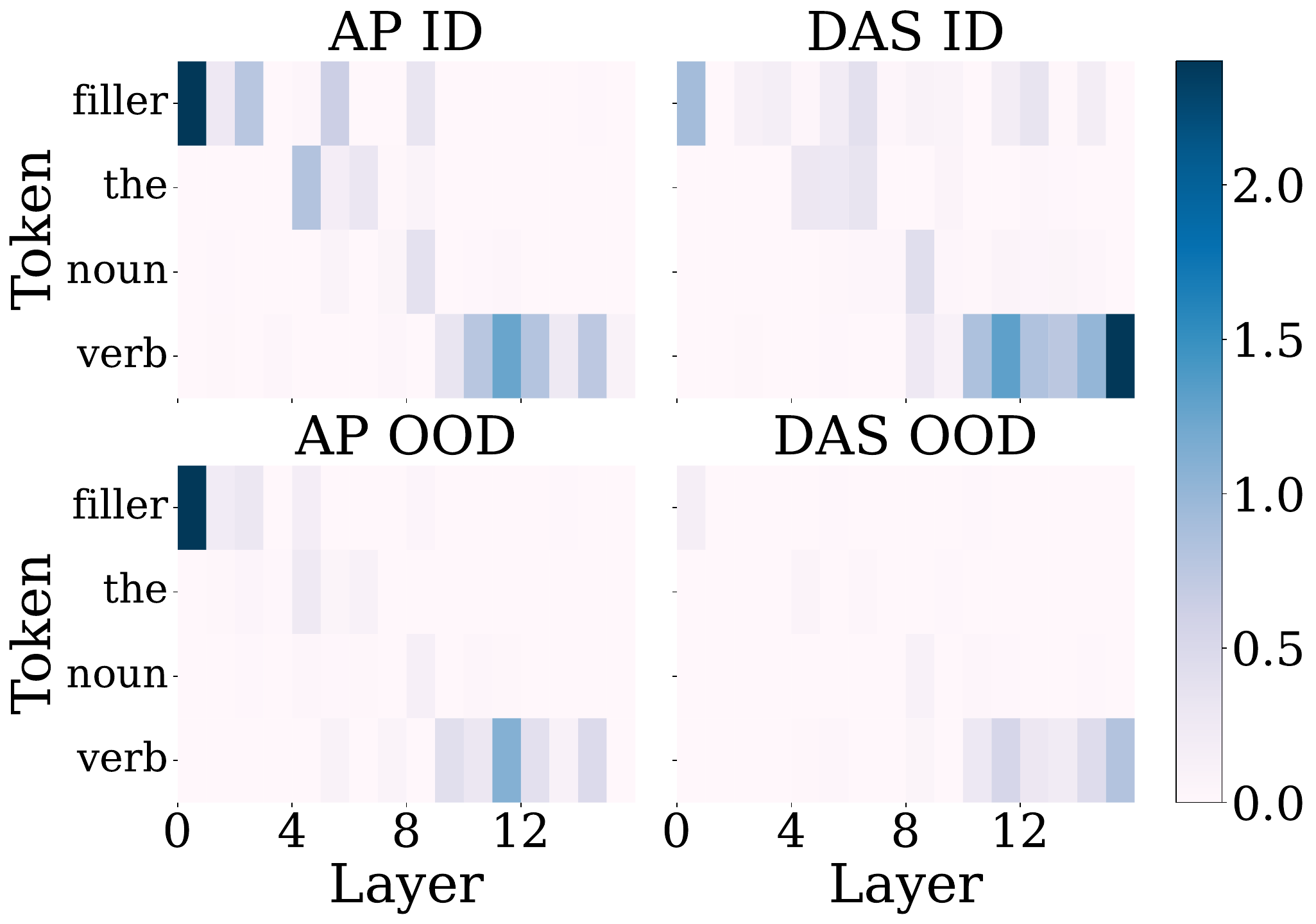}
    \caption{\textsc{PCleft}}
  \end{subfigure}
  \begin{subfigure}{0.315\linewidth}
    \centering
    \includegraphics[width=\linewidth]{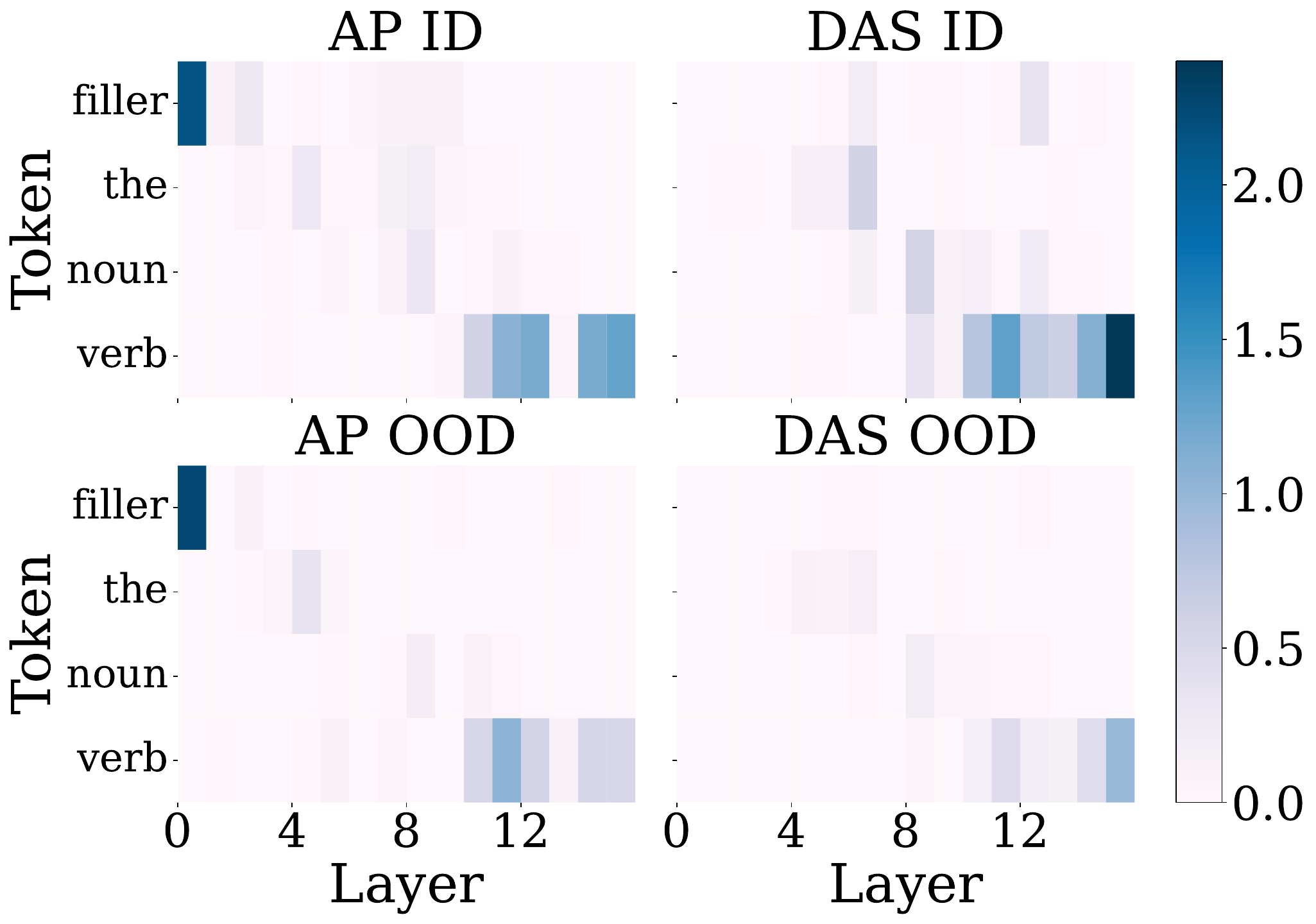}
    \caption{\textsc{Topic}}
  \end{subfigure}

  \caption{Comparison of \textsc{Odds} scores between activation patching (AP) and DAS of MLP output in Pythia 1B, evaluated on all the constructions in FGDs.}
  \label{fig:act_patch_fg_generalization_mlp}
\end{figure*}

\begin{figure*}
\centering

  \begin{subfigure}{0.315\linewidth}
    \centering
    \includegraphics[width=\linewidth]{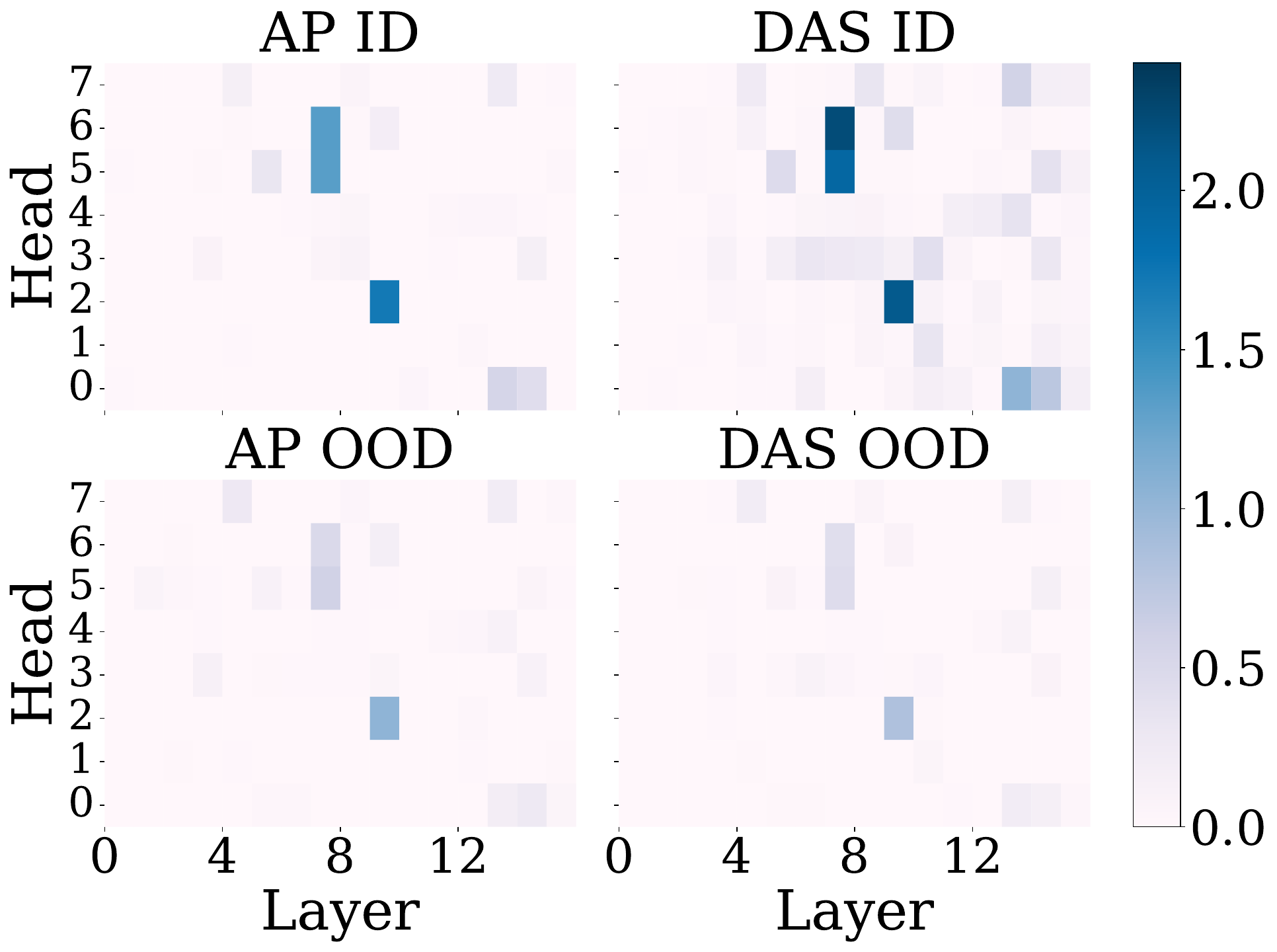}
    \caption{\textsc{EWhK}}
  \end{subfigure}
  \begin{subfigure}{0.315\linewidth}
    \centering
    \includegraphics[width=\linewidth]{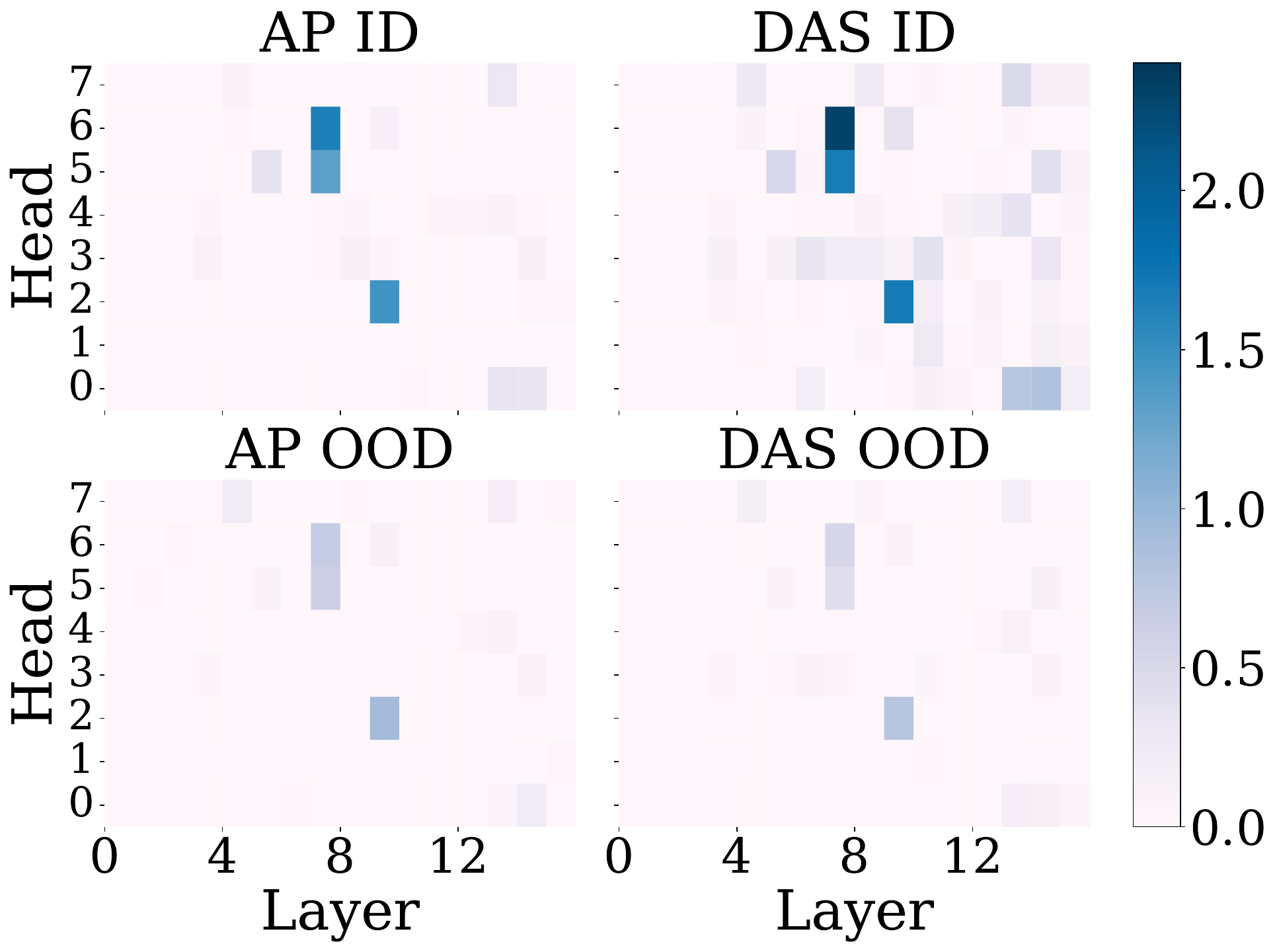}
    \caption{\textsc{EWhW}}
  \end{subfigure}
  \begin{subfigure}{0.315\linewidth}
    \centering
    \includegraphics[width=\linewidth]{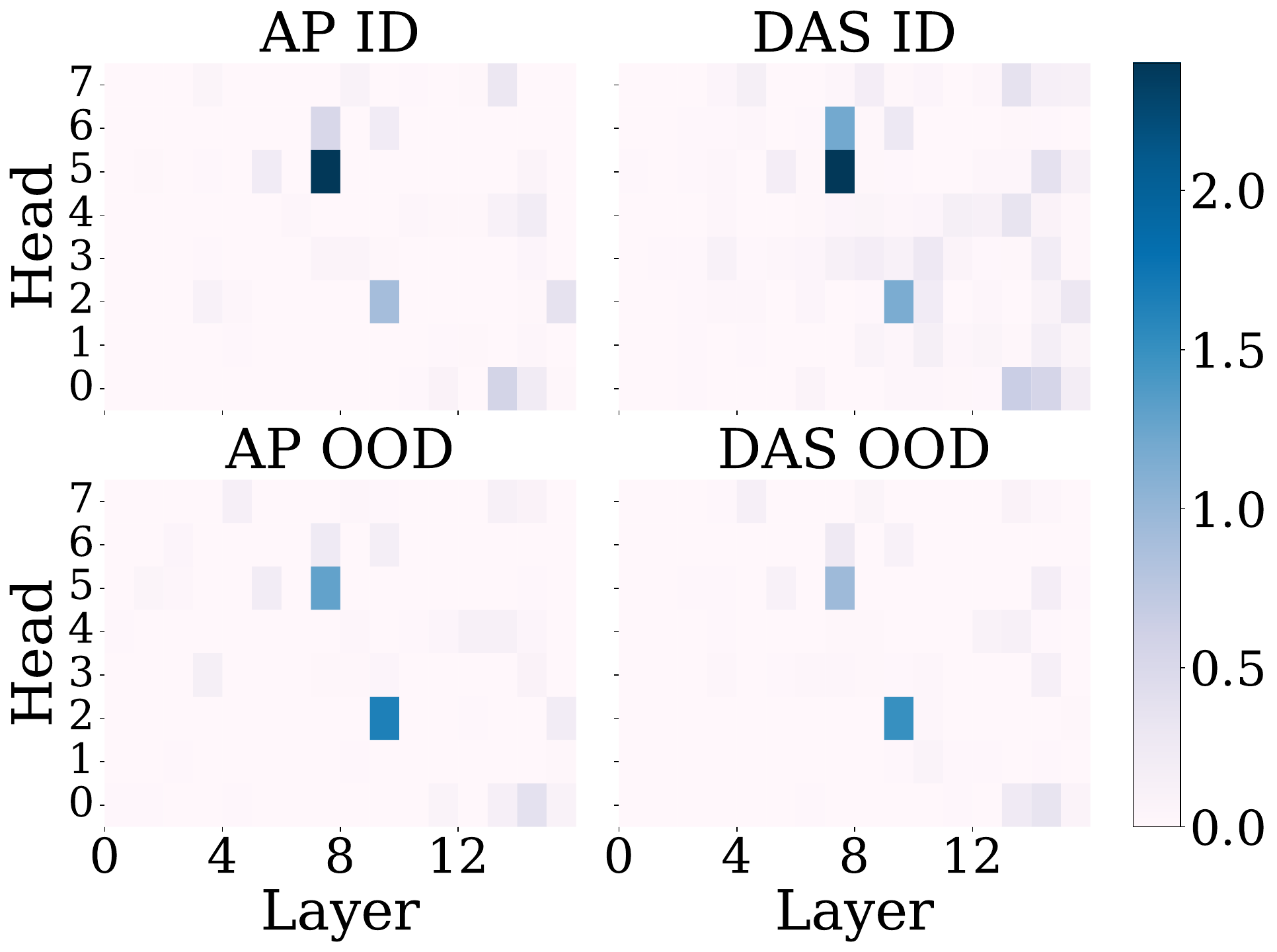}
    \caption{\textsc{MWh}}
  \end{subfigure}
  \begin{subfigure}{0.315\linewidth}
    \centering
    \includegraphics[width=\linewidth]{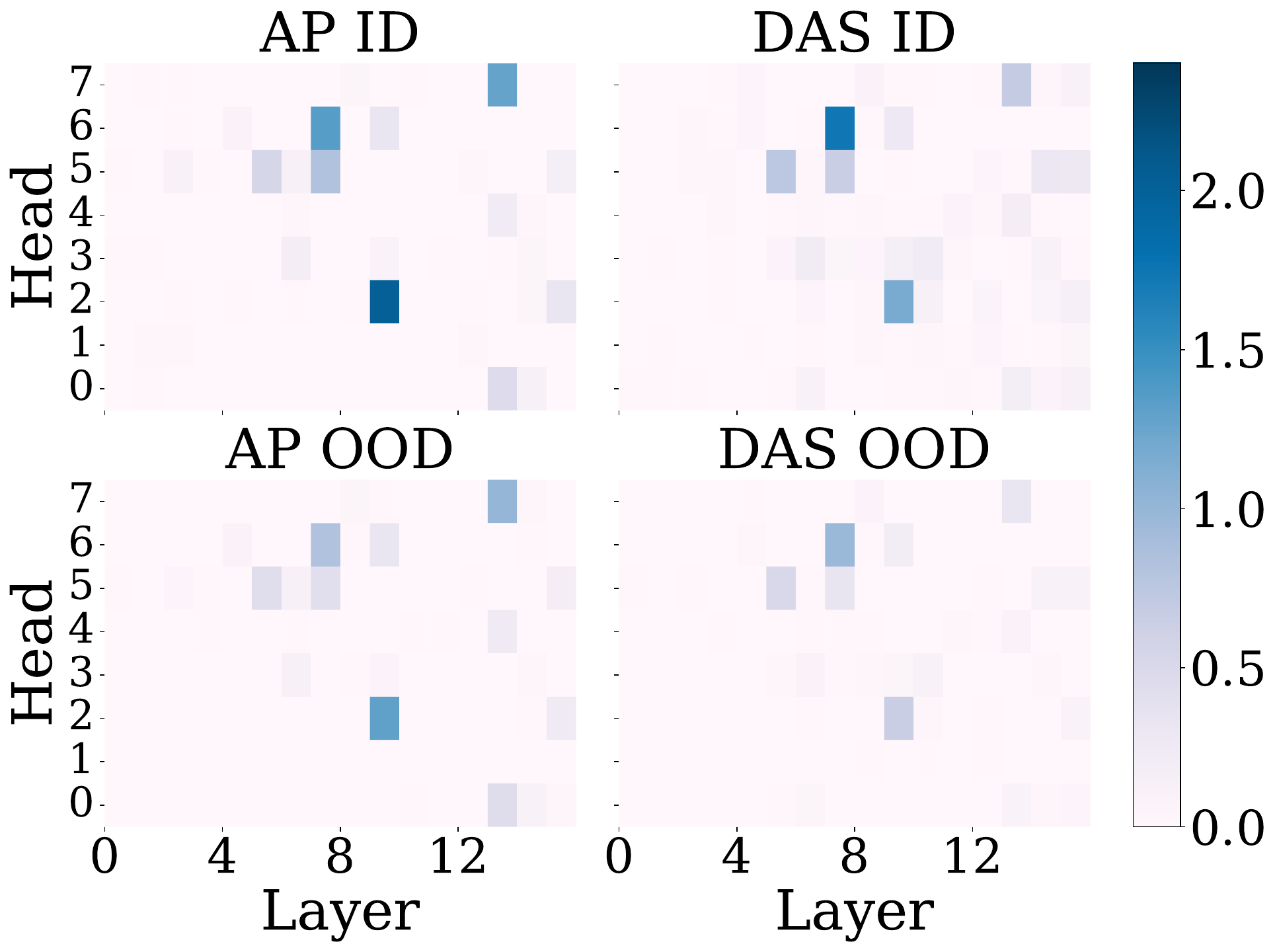}
    \caption{\textsc{RelCl}}
  \end{subfigure}
  \begin{subfigure}{0.315\linewidth}
    \centering
    \includegraphics[width=\linewidth]{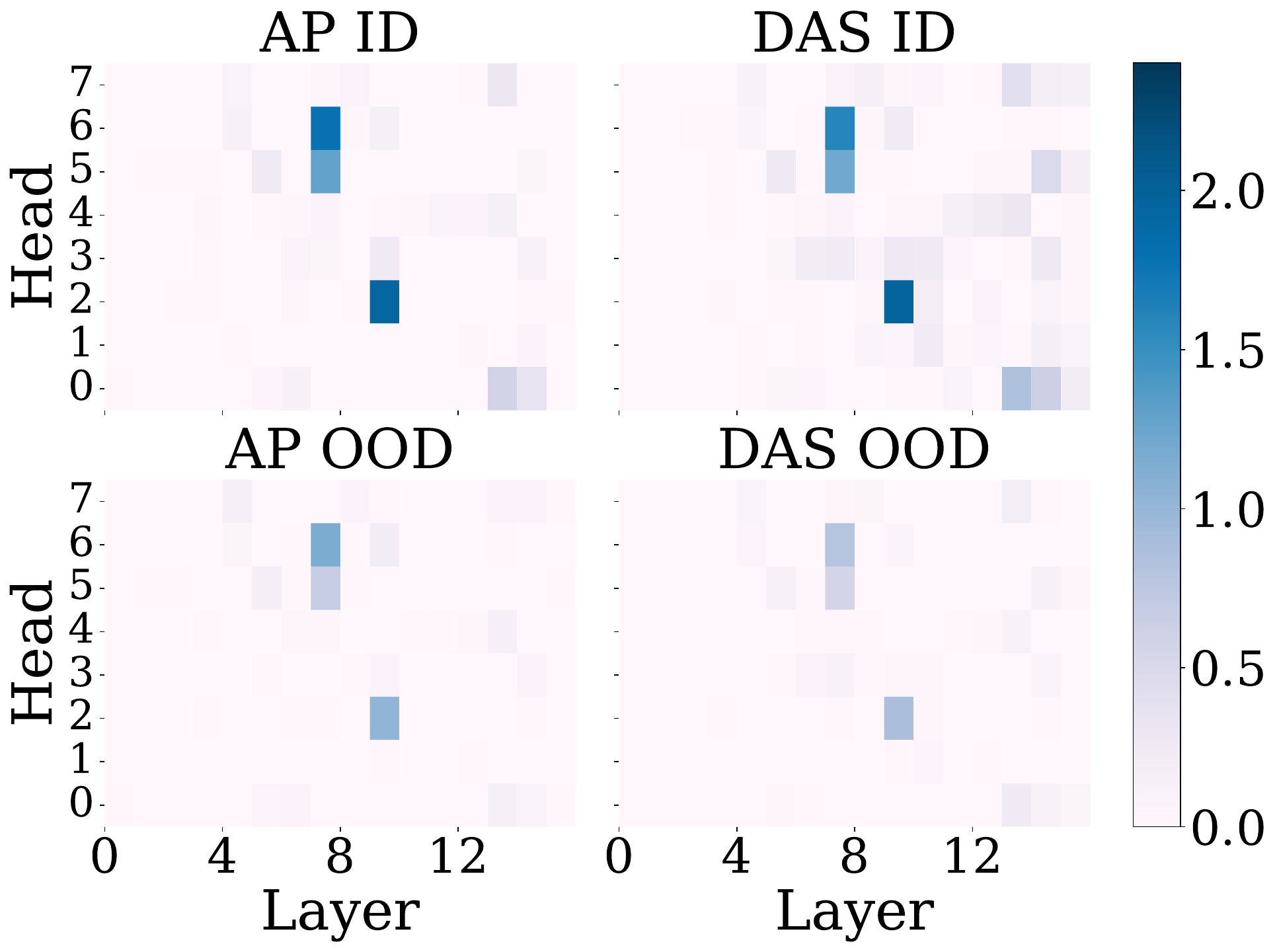}
    \caption{\textsc{Cleft}}
  \end{subfigure}
  \begin{subfigure}{0.315\linewidth}
    \centering
    \includegraphics[width=\linewidth]{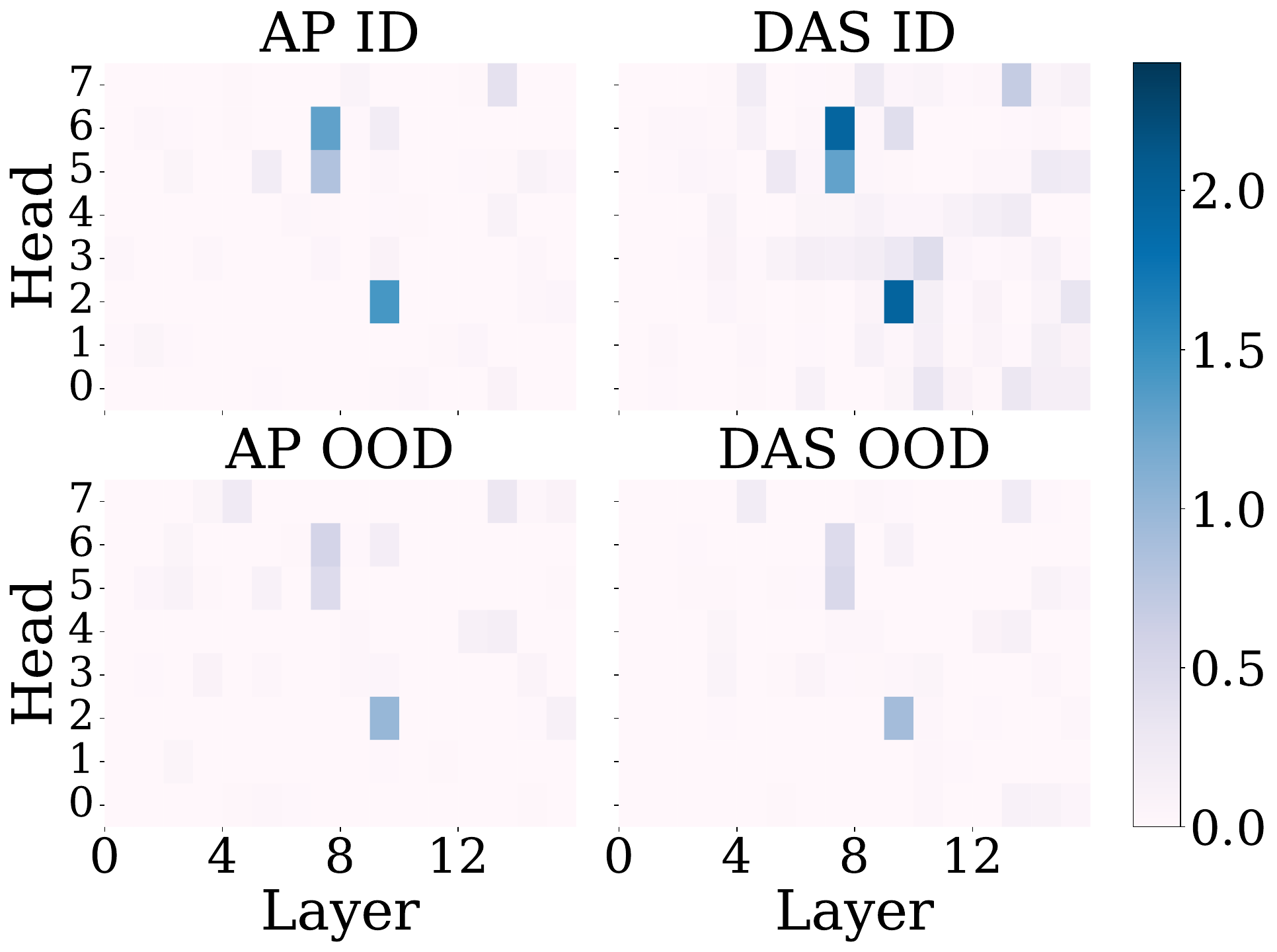}
    \caption{\textsc{PCleft}}
  \end{subfigure}
  \begin{subfigure}{0.315\linewidth}
    \centering
    \includegraphics[width=\linewidth]{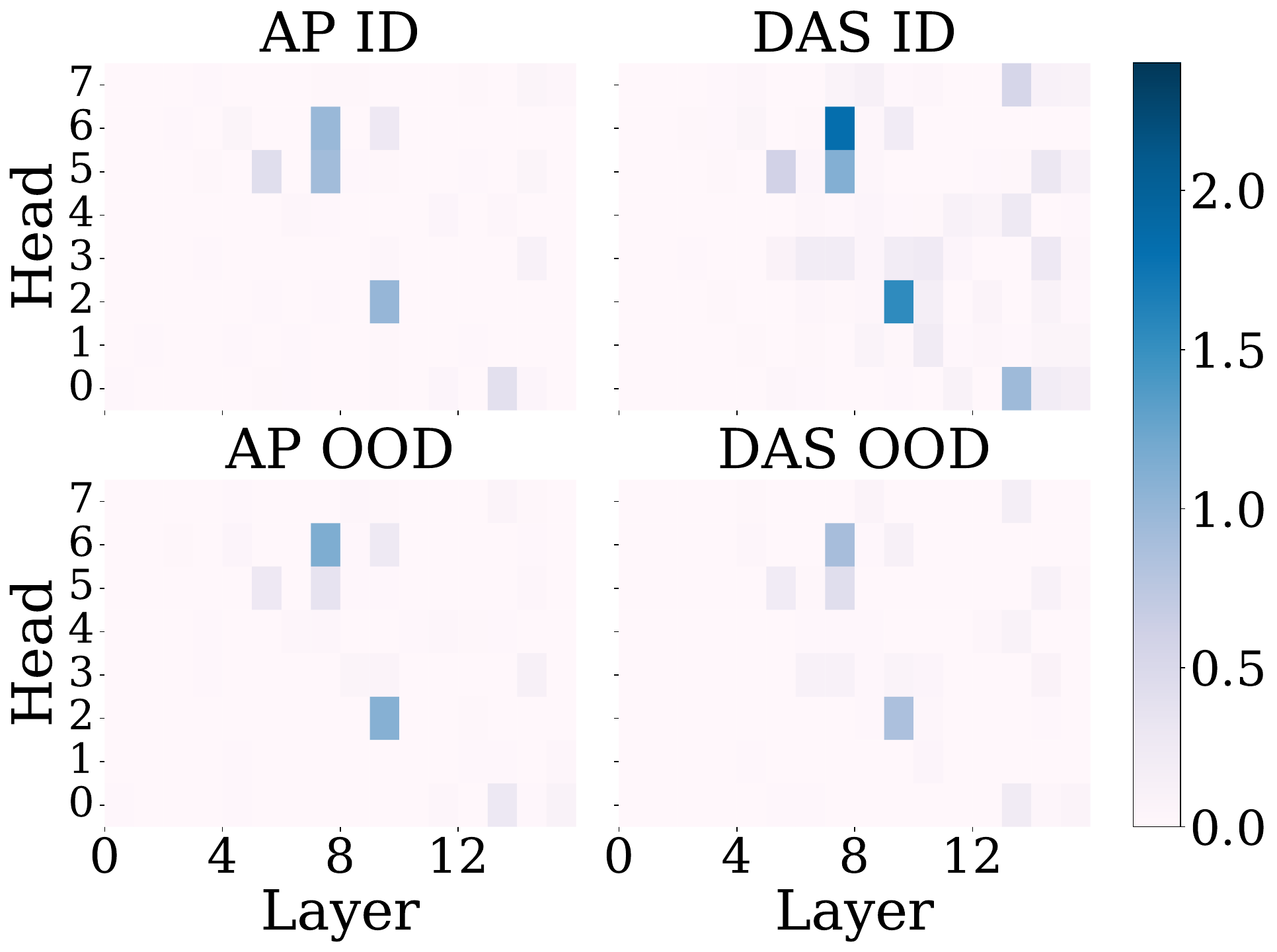}
    \caption{\textsc{Topic}}
  \end{subfigure}

  \caption{Comparison of \textsc{Odds} scores between activation patching (AP) and DAS of attention heads in Pythia 1B, evaluated on all the constructions in FGDs.}
  \label{fig:act_patch_fg_generalization_attn-head}
\end{figure*}

We present the results of comparisons between activation patching and DAS applied to Pythia 1B.
Figure~\ref{fig:act_patch_fg_generalization_resid} through \ref{fig:act_patch_fg_generalization_attn-head} show the results of residual stream, attention output, MLP output, and attention heads.

\subsection{Naturally Ocurring Data}
\label{subsec:natural}
\begin{figure}
\centering

  \begin{subfigure}{\linewidth}
    \centering
    \includegraphics[width=\linewidth]{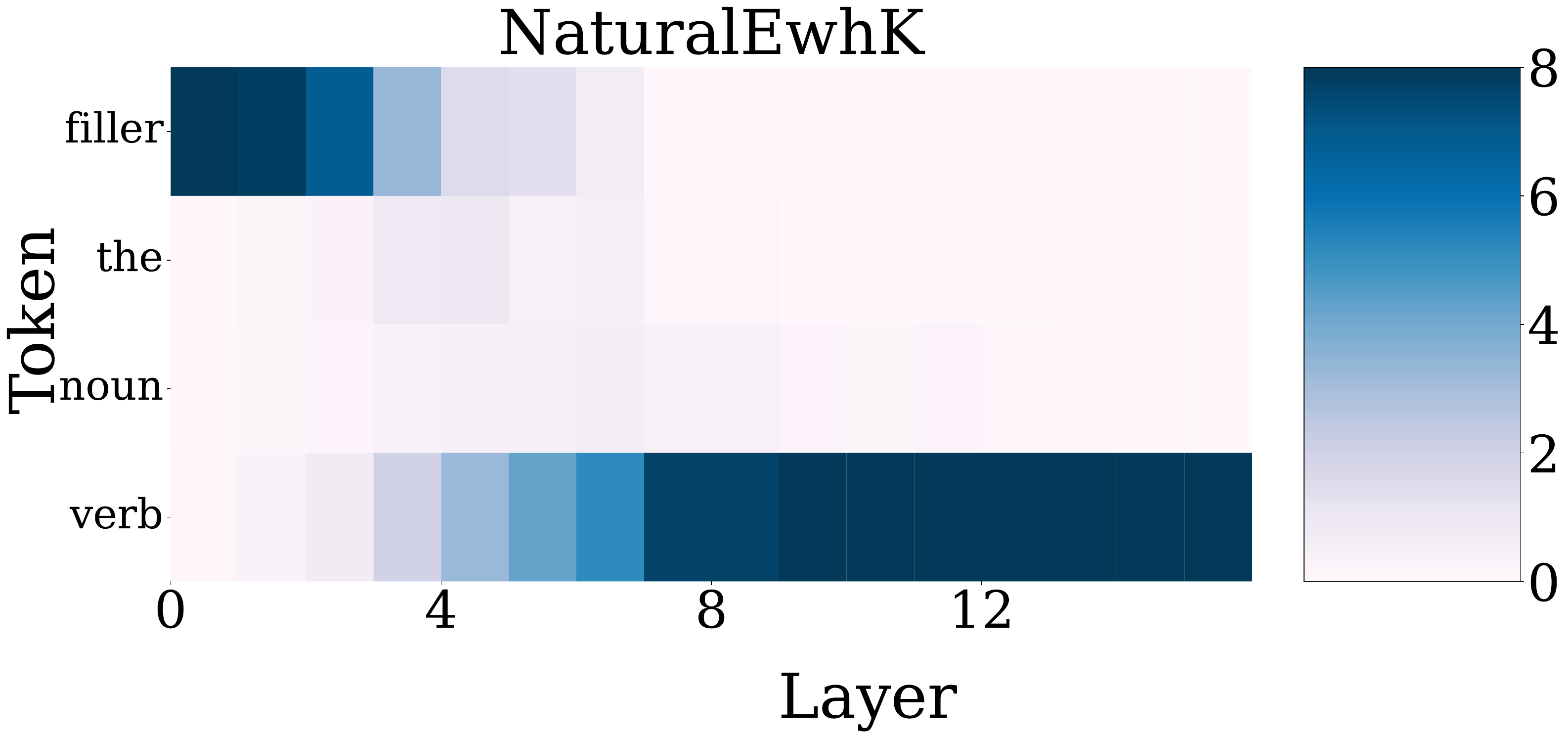}
    \caption{Residual Stream}
  \end{subfigure}
  \begin{subfigure}{\linewidth}
    \centering
    \includegraphics[width=\linewidth]{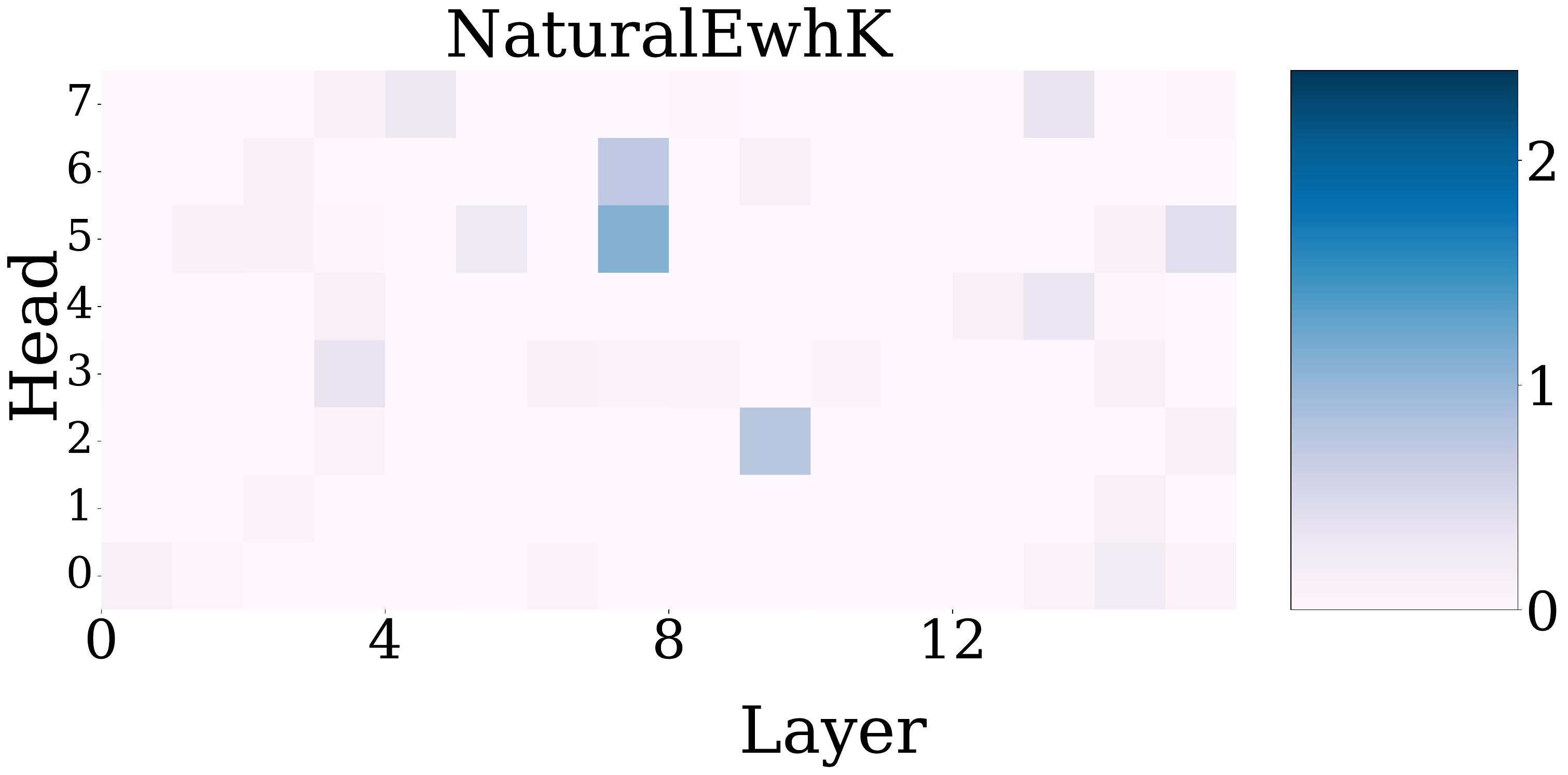}
    \caption{Attention head}
  \end{subfigure}
  \caption{\textsc{Odds} scores with activation patching in filler-gap dependencies in naturally occuring sentences.}
  \label{fig:act_patch_fg_natural}
\end{figure}

To investigate the mechanism of language models in processing natural sentences, we extract 20 naturally occurring sentences containing wh-embedding structures from the English-EWT Universal Dependencies dataset~\citep{de-marneffe-etal-2021-universal,nivre-etal-2020-universal,silveira-etal-2014-gold} and performed activation patching on Pythia-1B.

The results, shown in Figure~\ref{fig:act_patch_fg_natural}, demonstrate the exact same trends observed with our artificially constructed filler-gap structures, such as \textsc{EWhK} and \textsc{EWhW}.
Specifically, attention heads 7.5, 7.6, and 9.2 exhibited high scores, and the distribution across the residual stream remained identical.
These findings confirm that our analysis generalizes to naturally occurring sentences.

\clearpage

\section{Steering Results on SyntaxGym}
\label{sec:steering_syntaxgym}

\begin{figure*}
\centering

  \begin{subfigure}{0.47\linewidth}
    \centering
    \includegraphics[width=\linewidth]{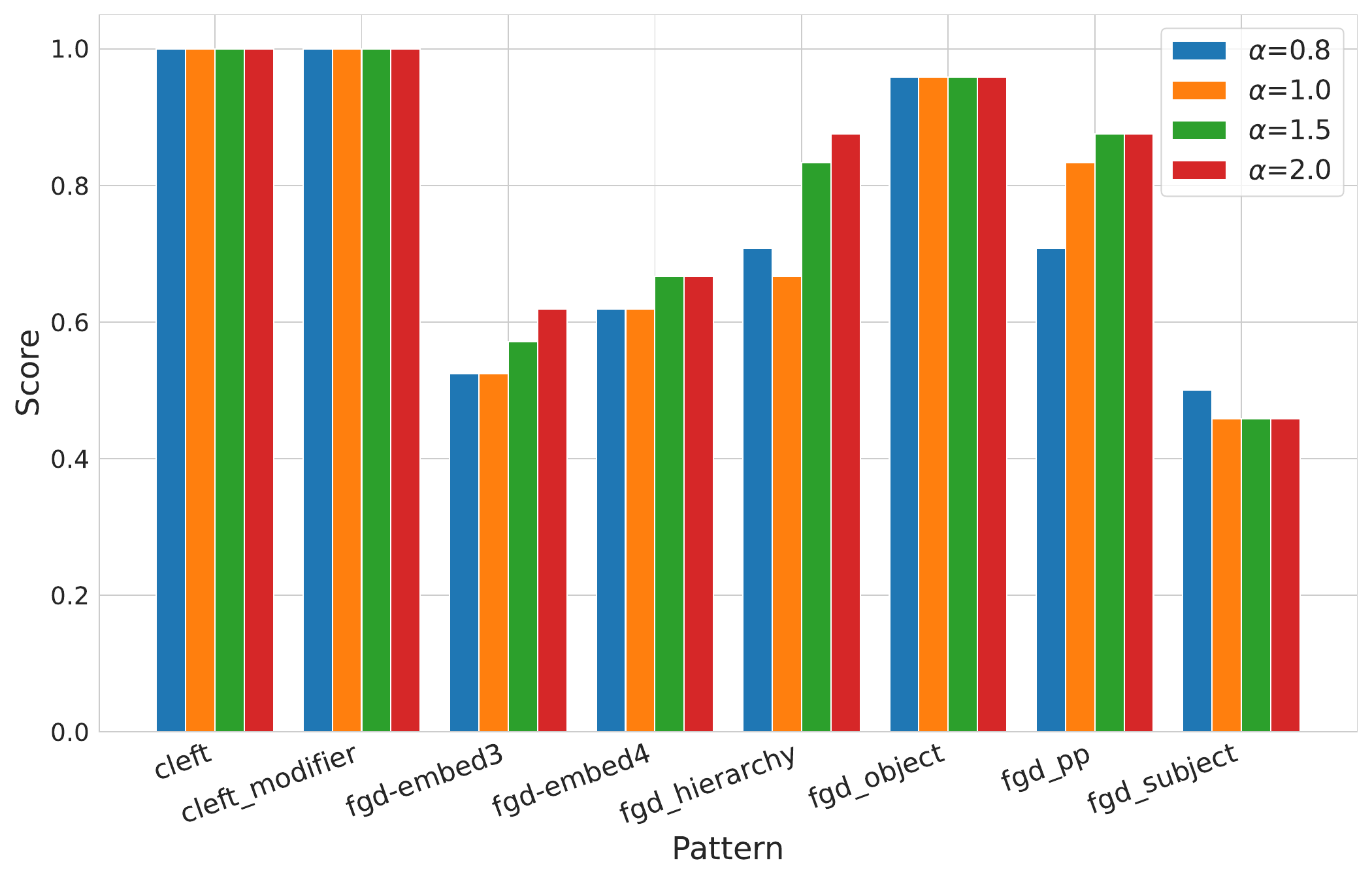}
    \caption{\textsc{Long-distance Dependencies} category}
     \label{fig:syntaxgym_fg}
  \end{subfigure}
  \begin{subfigure}{0.47\linewidth}
    \centering
    \includegraphics[width=\linewidth]{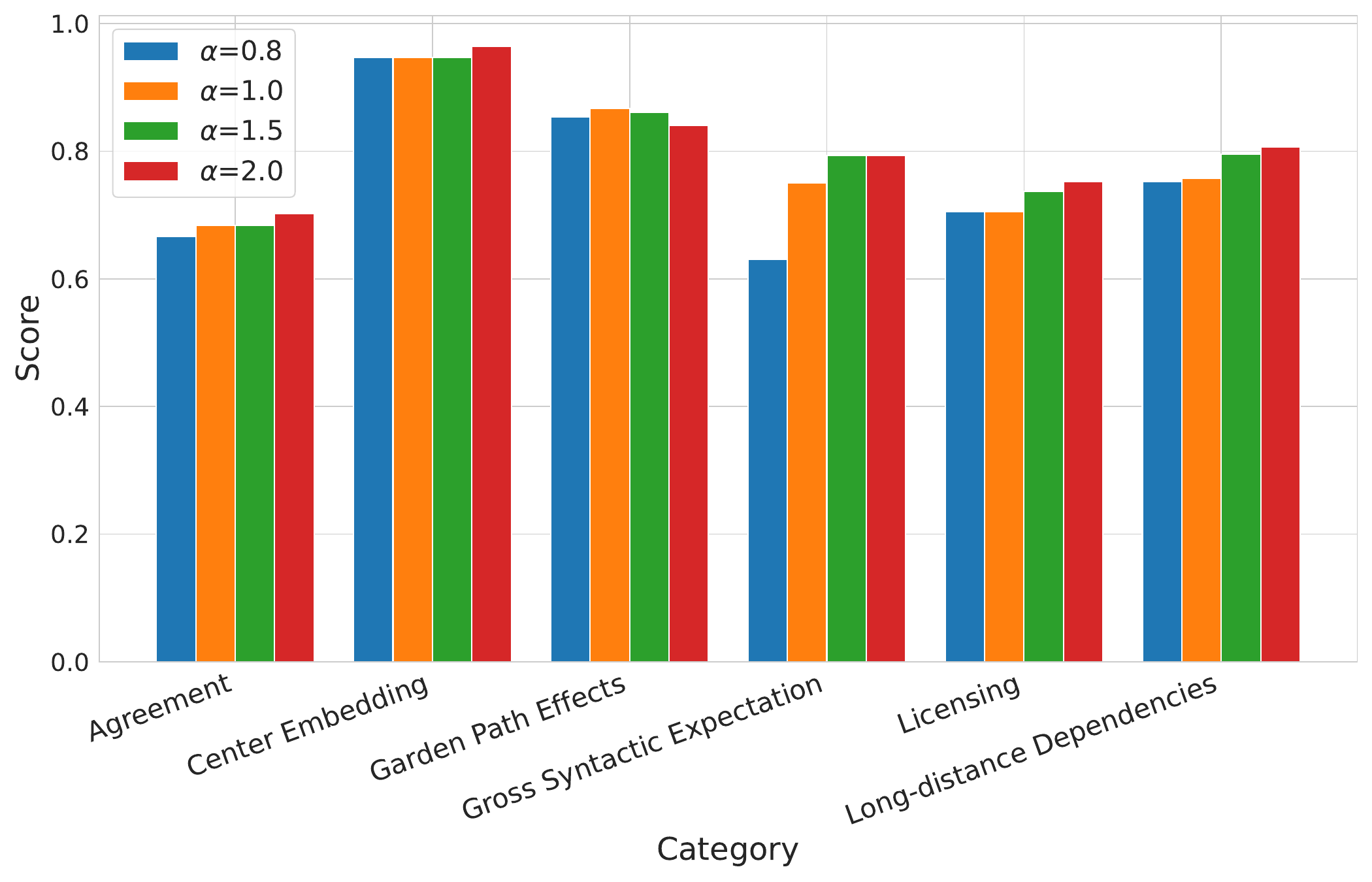}
    \caption{All categories}
    \label{fig:syntaxgym_all}
  \end{subfigure}

  \caption{Scores in the category involving filler-gap dependencies (left) and in all categories (right) in SyntaxGym after multiplying activations of several attention heads in Pythia 1B by $\alpha$.}
  \label{fig:steer_syntaxgym}
\end{figure*}

We extend the steering experiments described in Section~\ref{sec:steering} to SyntaxGym~\citep{hu-etal-2020-systematic}.
SyntaxGym differs from BLiMP in that it calculates the probability of only regions that differ among sentences in comparison, instead of using the probability of the entire sentence.
For example, in a subject-verb agreement pattern where the distractor is a prepositional phrase:
\begin{exe}
\ex The author next to the senators is${}_{V_\mathrm{sg}}$ good.
\ex[*]{The author next to the senators are${}_{V_\mathrm{pl}}$ good.}
\ex The authors next to the senator are${}_{V_\mathrm{pl}}$ good.
\ex[*]{The authors next to the senator is${}_{V_\mathrm{sg}}$ good.}
\end{exe}
The evaluation criteria are defined by the raw output probabilities $P(\cdot)$ at the verb position:
\begin{equation*}
P_1(V_{\mathrm{sg}})>P_2(V_{\mathrm{pl}}) \land P_3(V_{\mathrm{pl}})>P_4(V_{\mathrm{sg}})
\end{equation*}

However, we argue that certain SyntaxGym patterns for FGDs are unsuitable for evaluating the syntactic competence of causal language models.
Consider the subject extraction pattern:
\begin{exe}
\ex You remember that $\overbrace{\text{the businessman}}^{\alpha}$ showed the presentation to the visitors after lunch.
\ex[*]{You remember who $\overbrace{\text{the businessman}}^{\alpha}$ showed the presentation to the visitors after lunch.}
\ex You remember who $\overbrace{\text{showed}}^{\beta}$ the presentation to the visitors after lunch.
\ex[*]{You remember that $\overbrace{\text{showed}}^{\beta}$ the presentation to the visitors after lunch.}
\end{exe}
SyntaxGym use surprisals $S(\cdot)$ for evaluation:
\begin{equation*}
S_6(\alpha)>S_5(\alpha)\land S_8(\beta)>S_7(\beta)
\end{equation*}

Because causal language models process text incrementally, none of the strings above leading up to $\alpha$ or $\beta$ are ungrammatical, and their grammaticality depends on the following tokens.
Thus, comparing surprisals at these positions may not faithfully reflect the model's grammatical judgment.

Moreover, SyntaxGym has a considerably smaller sample size, around 20 to 40 per pattern, compared to the 1,000 samples per pattern in BLiMP.
This limits the sensitivity to capture subtle improvements in language models' syntactic capability.

Despite these caveats, we report results on SyntaxGym when the attention heads 7.5, 7.6, and 9.2 are manipulated.
The results are shown in Figure~\ref{fig:steer_syntaxgym}.
The overall trends were similar to those of BLiMP, as $\alpha>1$ led to improvement of the scores in many of the patterns and categories.
In the patterns where the scores did not improve, such as the ``fgd\_subject'' pattern, the results may have reflected the limitations of the evaluation criteria mentioned above rather than a failure of the manipulated mechanism.

\section{Steering Results on HANS}
\label{sec:hans}
\begin{table}[ht]
\centering
\setlength{\tabcolsep}{2pt}
\begin{tabular}{cc}
\toprule
$\alpha$ & Accuracy (\%)\\
\midrule
0.8 & 51.6 \\
1.0 & 52.3 \\
1.2 & 53.5 \\
1.5 & 54.9 \\
\bottomrule
\end{tabular}

\caption{Accuracy in HANS after multiplying activations of several attention heads in Gemma 3 1B IT by $\alpha$.
}
\label{tab:hans}
\end{table}

Our findings suggest potential performance gains in downstream tasks that rely on syntactic dependency processing.
To investigate this, we conducted additional steering experiments using HANS~\citep{mccoy-etal-2019-right}, an NLI benchmark.
We multiply the activation values of the attention heads that contribute to the processing of filler-gap dependencies by a factor of $\alpha$ and observe the resulting changes in NLI performance.
We follow \citet{madaan-etal-2025-lost} for prompt and few-shot examples for the evaluation with HANS, and we use Gemma 3 1B IT as the model.

The results of the performance are shown in Table~\ref{tab:hans}.
The results show that amplifying the activation values of identified attention heads leads to a better performance in NLI, and the findings in the paper are applicable to application tasks.

\section{Details of Computational Experiments}
We used nnsight~\citep{fiottokaufman2024nnsightndifdemocratizingaccess} and pyvene~\citep{wu-etal-2024-pyvene} for the implementation of the experiments.
We ran each experiment one time.
We used A100 GPUs for around 300 hours for the experiments.

\end{document}